\pdfoutput=1

\documentclass[11pt]{article}

\usepackage[final]{acl}

\usepackage{times}
\usepackage{latexsym}

\usepackage[T1]{fontenc}

\usepackage[utf8]{inputenc}

\usepackage{microtype}

\usepackage{inconsolata}

\usepackage{graphicx}

\usepackage{layouts}

\usepackage{hyperref}
\usepackage{url}

\usepackage{amsthm,amsmath,amssymb,amsfonts}
\usepackage{wrapfig}
\usepackage{caption}
\usepackage{subcaption}
\usepackage{booktabs}       %
\usepackage{xcolor,colortbl}
\usepackage[most]{tcolorbox}
\usepackage{multirow}
\usepackage{arydshln}
\usepackage{stackengine}  %

\usepackage{soul}

\usepackage{xspace}

\newcommand{\eg}[0]{\textit{e.g.},\xspace}
\newcommand{\ie}[0]{\textit{i.e.},\xspace}
\newcommand{\llm}[0]{LLM\xspace}
\newcommand{\llms}[0]{LLMs\xspace}

\newcommand{\olmo}[0]{\texttt{OLMo (7B)}\xspace}
\newcommand{\gemma}[0]{\texttt{Gemma (2B)}\xspace}
\newcommand{\llama}[0]{\texttt{LLama3 (70B)}\xspace}
\newcommand{\llamasmall}[0]{\texttt{LLama3 (8B)}\xspace}
\newcommand{\gemini}[0]{\texttt{Gemini}\xspace}
\newcommand{\mixtralmoelg}[0]{\texttt{Mixtral 8x22B}\xspace}
\newcommand{\mixtralmoe}[0]{\texttt{Mixtral 8x7B}\xspace}
\newcommand{\chatgpt}[0]{\texttt{ChatGPT}\xspace}
\newcommand{\gptf}[0]{\texttt{GPT-4}\xspace}
\newcommand{\gptfo}[0]{\texttt{GPT-4o}\xspace}

\title{Perceptions of Linguistic Uncertainty by Language Models and Humans}

\author{
  \textbf{Catarina Belem\textsuperscript{*,1}},
  \textbf{Markelle Kelly\textsuperscript{*,1}},
  \textbf{Mark Steyvers\textsuperscript{1,2}},
  \textbf{Sameer Singh\textsuperscript{1}},
  \textbf{Padhraic Smyth\textsuperscript{1}},
\\
\\
  \textsuperscript{1}Department of Computer Science, University of California Irvine
\\  \textsuperscript{2}Department of Cognitive Sciences, University of California Irvine
}

\begin{document}
\maketitle
\begin{abstract}
\textit{Uncertainty expressions} such as ``probably'' or ``highly unlikely'' are pervasive in human language.
While prior work has established that there is population-level agreement in terms of how humans quantitatively interpret these expressions, there has been little inquiry into the abilities of language models in the same context.
In this paper, we investigate how language models map linguistic expressions of uncertainty to numerical responses. 
Our approach assesses whether language models can employ theory of mind in this setting: 
understanding the uncertainty of another agent about a particular statement, independently of the model's own certainty about that statement. 
We find that 7 out of 10 models are able to map uncertainty expressions to probabilistic responses in a human-like manner. 
However, we observe systematically different behavior depending on whether a statement is actually true or false. 
This sensitivity indicates that language models are substantially more susceptible to bias based on their prior knowledge (as compared to humans). 
These findings raise important questions and have broad implications for human-AI and AI-AI communication.

\end{abstract}

\section{Introduction}
\label{sec:introduction}
The expression of uncertainty is ubiquitous in human communication --- in relaying predictions (``it is likely to rain tomorrow''), conveying imperfect knowledge (``I think I have a copy in my desk''), and describing unknown information (``the artifact could be more than 500 years old''). 
\footnotetext[0\def\thefootnote{}]{$^*$Authors contributed equally to this work. Correspondence: \href{mailto:cbelem@uci.edu}{cbelem@uci.edu}.}
Expressing uncertainty is particularly critical in fields such as medicine, law, and politics, where statements including \textit{uncertainty expressions} (\eg ``likely,'' ``doubtful'') are frequently used to support medical, judicial, and political decisions~\citep{Karelitz2004}.
Domain experts use these expressions to communicate uncertainty across a variety of situations, such as the likelihood of side-effects of a medical treatment~\citep{medical-Sawant2018, patt2005communicating}, the chances of a not-guilty verdict in legal cases~\citep{Fore_2019}, the probability of environmental events resulting from climate change~\citep{patt2005communicating, ho2015improving}, or the likelihood of emergence of military conflicts~\citep{military-Duke2023}.
Prior work has found that, in general, humans are well-attuned to the use of such uncertainty expressions, exhibiting population-level agreement in mapping these expressions to corresponding probabilities~\citep{wallsten1986measuring,Willems2019VariabilityIT,fagen-ulmschneider2019}.

\begin{figure}[tb]
    \centering
     \includegraphics[width=0.9\columnwidth]{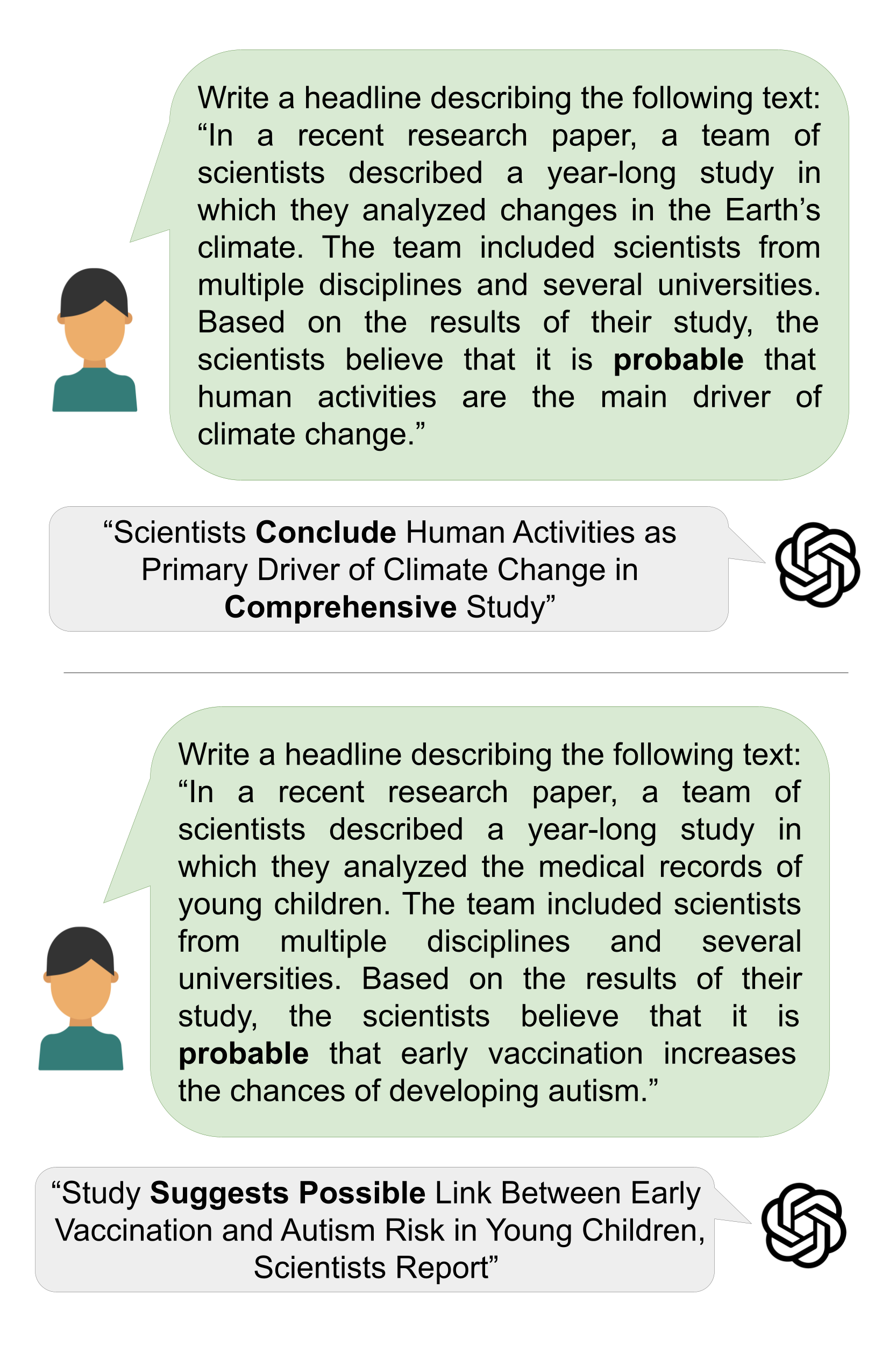}
    \caption{\textbf{Two interactions with \chatgpt (June 2024) concerning the generation of a headline for a short passage}. Both passages are structured identically and qualified with the word ``probable,'' but the first is about climate change and the second about the link between vaccines and autism. For the first passage, \chatgpt generates a confident-sounding headline, using the words  ``conclude'' and ``comprehensive.'' The second headline is weaker, with words like ``suggests'' and ``possible.''}
    \label{fig:gpt}
    \vspace{-1.1em}
\end{figure}

However, the topic of how large language models (\llms) interpret linguistic uncertainty has received relatively little attention. 
In particular, given text where a speaker expresses uncertainty about a particular statement, this paper investigates whether \llms can interpret the uncertainty not as a function of the model's internal beliefs, but by objectively assessing the speaker's uncertainty about the statement. 
Consider the motivating example in Figure~\ref{fig:gpt}: when writing a headline for a statement qualified by the word ``probable,'' \chatgpt expresses substantially different uncertainty depending on its prior belief about the statement.\footnote{When prompted about its belief about these statements, \chatgpt agrees with the first and disagrees with the second; see Figure \ref{fig:appdx:gpt_belief} in the Appendix~\ref{app:additional-motivation}.}
In this example, \chatgpt is conflating the speaker's uncertainty with its own uncertainty about the statement---in effect,  a failure of ``theory of mind.'' 

In this work, we investigate the abilities of \llms to provide quantitative interpretations of uncertainty expressions, focusing in particular on how the prior knowledge of an \llm affects this ability.
To this end, we propose to evaluate models' capabilities as a function of their ability to map text containing uncertainty expressions to numerical responses. %
We analyze the performance of both humans and 10 popular \llms on this task, enabling direct comparison between humans and models.\footnote{The code and data can be found at \url{https://github.com/UCIDataLab/llm-uncertainty-perceptions}.} 
We find that larger, newer models like \gptf and \llama can consistently map uncertainty expressions to numerical responses that align with human population-level perceptions. 
However, we also show that the responses \llms generate are susceptible to bias based on their prior knowledge---with much greater susceptibility than that of humans. 

The models' sensitivity to their prior knowledge has concerning implications %
for the use of \llms in tasks in which they must process and generate text containing uncertainty expressions, \eg summarizing scientific reports or writing news articles~\citep{shao2024assistingwritingwikipedialikearticles,laban2024summaryhaystackchallengelongcontext}.
When an \llm's ability to quantify uncertainty can be ``poisoned'' by its beliefs, its downstream performance is dependent on its parametric or pretraining knowledge (which can be obsolete or wrong \cite{liangadvances2022, longpre2023pretrainer}), rather than on critical contextual information \citep{longpre2021entity}. 
Further, this means that the biases of a model (including the many well-documented potentially harmful biases of \llms, \eg \citet{wan2023kelly, kotek2023gender,salewski2024context,scherrer2024evaluating,motoki2024more}) can subtly manifest in how it interprets and generates uncertainty language and, as a consequence, have broader implications for human-AI and AI-AI interactions. 

\section{Related Work}
\label{sec:background}

\paragraph{Human Perceptions of Uncertainty Expressions.}
In fields like medicine, finance, law, and politics, where it is impossible to make predictions with complete certainty,  decisions are often informed by subjective probabilities~\citep{Karelitz2004,Dhami-Wallsten-2005,Fore_2019}. 
Subjective probabilities can be communicated quantitatively, \eg through numerical probabilities, odds, percentages, intervals, or qualitatively, through the use of uncertainty expressions or epistemological markers (\eg ``I believe'', ``According to'')~\citep{Dhami2022}. 
Although they are less precise than numerical values~\citep{WALLSTEN1986571,Brun1988,Budescu_Por_Broomell_Smithson_2014}, humans generally prefer to use linguistic expressions to communicate uncertainty~\citep{Erev-Cohen-1990,Wallsten-et-al-1993}. 

Interested in the efficacy of how humans communicate  uncertainty linguistically, researchers have examined how individuals map uncertainty expressions into numerical values across different fields and expertise levels (\citet{windschitl1996measuring,Karelitz2004,wallsten2008intel,wallsten1986measuring,Fore_2019}; \textit{inter alia}). 
Although there can be considerable variation in responses at the individual level, these studies have   revealed consistent and systematic patterns relating uncertainty expressions and numerical responses at the population level \citep{wallsten2008intel,Willems2019VariabilityIT,fagen-ulmschneider2019}. 

\paragraph{Uncertainty Quantification in \llms.} 
The need for more reliable \llms has prompted researchers to investigate new methods for communicating the internal uncertainty of \llms.
Proposed methods can be differentiated in terms of the information used to estimate the model's uncertainty: 
from token-level information~\citep{jiang-etal-2021-know,kuhn2023semantic,duan2024shifting}, to dissimilarities across multiple samples~\citep{si-etal-2022-examining,chen2023quantifying,xiong2024llms,hou2024decomposing,lin2024generating,aichberger2024semantically}, to training external classifiers using the inputs and/or \llms' representations~\citep{jiang-etal-2021-know,mielke-etal-2022-reducing,Shrivastava2023LlamasKW}, or even directly eliciting confidence estimates from \llms as output tokens~\citep{lin2022teaching,tian-etal-2023-just}. 
Furthermore, several works have analyzed the impact of \llm-articulated uncertainty in human-AI interaction, finding that participants adjust their perception of \llms' correctness when shown \llm outputs that include uncertainty expressions~\citep{Zhou2024RelyingOT,kim-im-not-sure-faact2024,steyvers2024calibration}. 
With the goal of calibrating human-AI interaction, ~\citet{chaudhry2024finetuninglanguagemodelsemit} propose fine-tuning \llms to convey uncertainty expressions that faithfully reflect their intrinsic uncertainty.
While these works investigate how we can gauge \llms' intrinsic uncertainty and how humans react to various uncertainty expressions in text, there has been far less work on the questions we focus on in this paper, \ie how \llms interpret linguistic uncertainty and how closely these interpretations match those of humans.

\paragraph{\llm Perceptions of Uncertainty Expressions.} 
A small body of recent work has begun to investigate the relationships between uncertainty expressions and model behavior. 
Recently, \citet{yona2024largelanguagemodelsfaithfully} found low correlations between \llms' intrinsic uncertainty and the use of uncertainty expressions.
\citet{sileo-moens-2023-probing} investigate whether \llms are able to discriminate between two uncertainty expressions and reason in terms of compositions of expressions. However, the paper focuses on \llms' binary rankings of expressions, as opposed to numerical interpretations, and does not compare \llm and human behavior. 
Most directly related to our work is that of \citet{Maloney2024-coordinationgame} which compares numerical probability estimates from \texttt{GPT-4} and humans using a small set of ``context'' prompts, and the work of \citet{tang2024evaluationestimativeuncertaintylarge} who compare the numeric-textual mapping of uncertainty expressions across four different contexts in both Chinese and English.  
Our paper goes significantly beyond this work by assessing a broad range of \llms using a more diverse and natural set of contexts. We additionally conduct human experiments, assessing the performance of humans on the same task to facilitate a direct comparison between humans and \llms. 
Further, our approach is designed to target ``theory of mind''---the task requires humans and \llms to quantify what an uncertainty expression reflects about the speaker's belief, rather than what the expression means to the human or \llm. Finally, our work is the first that we are aware of to investigate how \llms can be biased by their prior knowledge in mapping uncertainty expressions to numerical responses.

\section{Baseline Human Study}
\label{sec:initalexp}
\begin{figure}[tb]
    \centering
    \includegraphics[width=\linewidth]{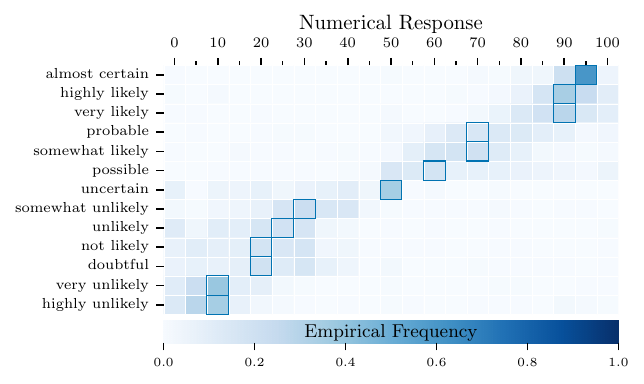}
    \caption{\textbf{Human empirical distributions of numerical responses per uncertainty expression in the non-verifiable setting}. Highlighted blue boxes represent the mode value for each expression. Overall, population-level perceptions increase monotonically with the use of more confident uncertainty expressions.}
    \label{fig:hist-nv:humans}
    \vspace{-1em}
\end{figure}

As a baseline for how people map uncertainty expressions to numerical probabilities, we first conducted an experiment in which 94 humans were shown uncertainty expressions and asked to provide corresponding numerical responses. 
We focused on a set of 14 uncertainty expressions (\eg ``almost certain,'' ``unlikely''---the full list is provided in Appendix \ref{app:ExperimentDetails} and is also shown on the y-axis in Figure \ref{fig:hist-nv:humans}), drawn from \citet{wallsten1986measuring} and \citet{wallsten2008intel}.
In this initial experiment, our goal is to assess how people perceive these uncertainty expressions ``in the wild,'' putting them in the context of plausible real-world statements. 
In addition, we  use types of statements that attempt to minimize the potential for people to conflate their own beliefs about these statements with their assessment of the confidence of the person making the statement. 

To this end, we constructed a set of statements $(u, s, e)$ which include uncertainty expressions $u \in \mathcal{U}$ used by speakers $s \in \mathcal{S}$ to convey their degree of certainty about the truthfulness or falsehood of a statement or event $e \in \mathcal{E}$. 
By presenting statements as being made by a specific speaker $s$,\footnote{Names are selected arbitrarily from a pre-defined pool of names. Participants see each name once throughout the experiment. For more details, see Appendix \ref{app:ExperimentDetails}.} we are asking participants to use theory of mind to estimate how likely it is that the speaker believes that the statement is true. 
We then query participants about the speaker's degree of certainty, clearly distinguishing this notion from the participant's own beliefs. 
For instance, given the statement ``Sonia believes it is unlikely it will rain today,'' we can ask  participants to quantify with a numerical response how likely \textit{Sonia} thinks it is that it will rain today, distinct from  the participants' own  beliefs about how likely it is to rain. 
We use the term {\it numerical response} throughout the paper to refer to the participant's numerical assessment of the likelihood that speaker $s$ believes statement $e$ to be true when using uncertainty expression $u$. 
We instruct participants that responses should be numbers between 0 and 100, where 0 implies that speaker $s$ believes there is a $0\%$ chance that statement $e$ is true while 100 implies the speaker believes there is a $100\%$ chance that the statement is true.

In our baseline experiment, we use \textit{non-verifiable} statements to separate the meaning of the uncertainty expressions from uncertainty about the statements themselves. 
Non-verifiable statements are statements that are not sufficiently grounded with specific contextual information to allow an external observer to be confident in either the truth or falsity of the statement. 
For example, in the context of a prompt such as ``Maria believes it is likely that [statement],'' we consider statements such as {\it her boss has two pets} or {\it her flight will land around 6pm} non-verifiable---as there is not enough context for an observer to be able to assess the likelihood that the statement is true. 
In contrast, \textit{verifiable} statements (which we discuss further in Section \ref{ssec:datasetcreation}) can be verified as correct or incorrect in a context-free sense (e.g., {\it the capital city of Peru is Lima}); humans and \llms will often have strong prior beliefs about the likelihood that such statements are true.

\begin{figure}[tb]
    \centering
    \includegraphics[width=\columnwidth]{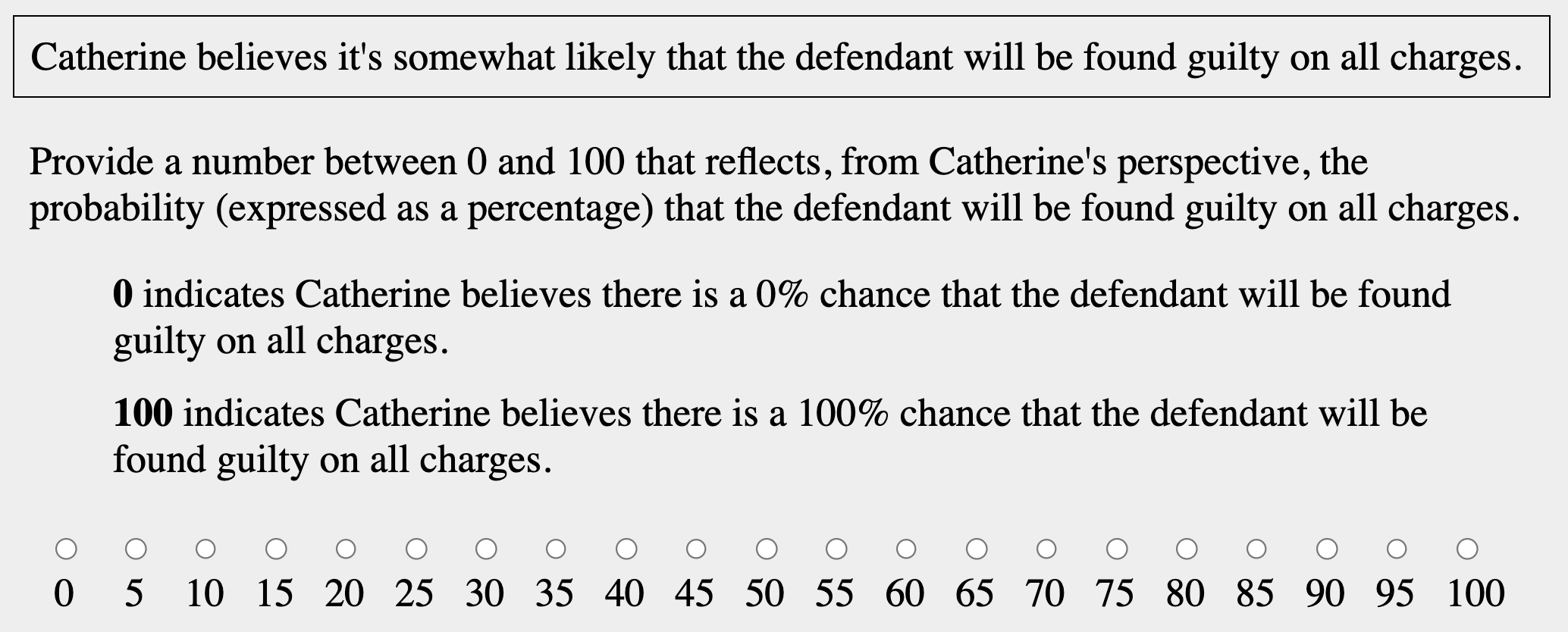}
    \caption{\textbf{Example of a non-verifiable statement provided to participants in the baseline experiment}. Each example uses a unique name and statement. Participants see one question at a time.}
    \label{fig:nv_example_human}
\end{figure}

We manually constructed a set of 60 non-verifiable statements and systematically combined these with 14 uncertainty expressions. 
We randomly selected speaker names, generating sentences describing the belief of a hypothetical speaker in the form: ``[Speaker] believes it is [uncertainty expression] that [statement].'' 
For each sentence, participants were asked to quantify the speaker's belief about the statement. In particular, they were asked what the probability is {\it from the speaker's perspective} that the statement is correct. 
Participants then provided their response quantized to numerical bins $0, 5, 10, \ldots, 95, 100$ (see example in Figure \ref{fig:nv_example_human}). 
Each of the 94 participants provided annotations for 28 randomly chosen statements (and speaker names). As a result, every uncertainty expression was annotated twice by each participant.

The outcome of this experiment\footnote{Additional details about the experiments can be found in Appendix \ref{app:ExperimentDetails}.} is an empirical distribution over the $2 \times 94$ numerical responses that participants associated with each uncertainty expression. 
For example, Figure \ref{fig:histograms-detail} reflects the histogram of responses assigned to the phrases ``very likely'' and ``very unlikely''; results for all 14 uncertainty expressions can be summarized in a heat-map as shown in Figure \ref{fig:hist-nv:humans}.  
We note that our results are in agreement with prior work on human perceptions of uncertainty expressions~\citep{wallsten2008intel,wallsten1986measuring,Willems2019VariabilityIT}, including  consistent ordering in aggregate population patterns, in terms of the mode of the empirical distributions.

\section{Methodology}
\label{sec:methodology}

Our full set of experiments includes experiments with both humans and \llms as participants and with both non-verifiable (NV) and verifiable (V) statements. 
The baseline experiment, as described in Section \ref{sec:initalexp}, consists of human participants and non-verifiable statements (denoted \texttt{human+NV}). 
In this section, we describe extensions to include verifiable statements and \llm participants, resulting in three additional sets of experiments: \texttt{human+V}, \texttt{LLM+V}, \texttt{LLM+NV}. 
To help us draw comparisons between the three additional settings and the baseline experiment, we conclude this section with the description of two metrics.

\subsection{Verifiable Statements}
\label{ssec:datasetcreation}

In addition to the non-verifiable statements described in Section \ref{sec:initalexp}, our dataset also includes \textit{verifiable} statements, for the purpose of assessing the effects of prior knowledge on quantifying linguistic uncertainty.
To increase the chances that both \llms and humans are familiar with the verifiable statements, we focus on concise, general-knowledge statements based on widely recognized facts (\eg geography, history of art, science).
Specifically, we create 60 verifiable statements based on a multiple-choice question-answering trivia dataset from The Question Company.\footnote{ \url{https://www.thequestionco.com/}}
Starting with 30 of the dataset's ``easy'' questions and corresponding multiple-choice options, we write \textit{true} statements that use the correct answer and \textit{false} statements using one of the incorrect answers. 
Examples of verifiable statements and additional details about the dataset are included in Appendix \ref{app:dataset-selection}. 
We focus on results for both humans and \llms with these 60 verifiable statements in the main paper but include a validation analysis using 400 additional statements in Section~\ref{ssec:ablation:generalization-study}. 
These additional statements are extracted via a similar procedure from  AI2-ARC~\citep{Clark2018ThinkYH}, a grade-school level, multiple-choice science question answering dataset used to evaluate state-of-the-art \llms' reasoning capabilities~\citep{open-llm-leaderboard,jiang2023mistral,openai2024gpt4}. 

\subsection{Numerical responses from \llms}
\label{ssec:methods-llms}

To obtain uncertainty estimates from \llms, we create prompts similar to the queries provided to humans (see Appendix \ref{app:ExperimentDetails}). 
Our goal is to estimate an empirical distribution, per uncertainty expression $u$, over each \llm's generated responses, in a manner similar to how empirical distributions for humans are generated (\eg see Figure~\ref{fig:hist-nv:humans}). 
In the results in this paper we focus on greedy decoding, where we select the numerical response that has the highest probability in the next-token probability distribution generated by the \llm conditioned on the prompt, \ie decoding with \texttt{temperature=0}. 
Because this sampling approach requires no knowledge about the weights or next-token probabilities, it is applicable to any model, including those behind black-box APIs, such as \gemini~\citep{geminiteam2024gemini} and \gptf~\citep{openai2024gpt4}. 
While focusing primarily on this greedy sampling approach allow us to efficiently compare the modal behavior of different \llm model families in equal terms (regardless of the available information), it does not provide insight into distributional behavior. Thus, in Section~\ref{sec:results} we include results using probabilistic decoding with \texttt{temperature=1} to evaluate the sensitivity of our conclusions to decoding method.
Additional information on the extraction methodologies used can be found in Appendix \ref{app:sec:extract-llm-numerical-response}.

\subsection{Metrics}
\label{ssec:metrics}

We treat the empirical distribution obtained for the non-verifiable statements with human participants (described in Section \ref{sec:initalexp}) as our \textit{reference distribution} for evaluation purposes, since it reflects human perceptions of uncertainty expressions in a setting that is designed to be free of prior information or biases about the corresponding statements.
For every uncertainty expression $u \in \mathcal{U}$, we define a \textit{reference conditional probability distribution} $P(k|u)$, $k=0,5,10,\ldots,95,100$, where $P(k|u)$ is the empirical distribution from the baseline experiment. 
Given a response from any agent, human or \llm, in the context of a particular uncertainty expression $u$ we measure the quality of the response using the reference distribution $P(k|u)$.

The primary quality metric that we propose is \textbf{Proportional Agreement (PA)}. 
PA can be defined as follows: if an agent's response matches bin $k$ for uncertainty expression $u$, then the PA value for that response is defined as $P(k|u)$, where $P$ is the reference (population) distribution defined above. 
Intuitively, for an expression $u$, this PA score $P(k|u)$ represents the probability that the agent's response $k$ agrees with that of a randomly selected individual, and is upper bounded for any expression by $\arg \max_k P(k|u)$, \ie by the mode of the $P(k|u)$ values. 
The higher the PA value, the better the quality of the response in terms of agreement with the aggregate human population (as reflected by $P(k|u)$).
To compute a single score for a particular \llm or individual human, we average the PA score over multiple responses and over the 14 uncertainty expressions.\footnote{Note that the PA metric is similar to the log-probability metric widely used to score probabilistic models in machine learning. 
However, it is not a likelihood in the sense that a likelihood corresponds to measuring the probability mass a model assigns to an observed outcome. 
Thus, in this non-likelihood context it is appropriate to average the PA scores directly (rather than taking products of probabilities as would be done under an IID likelihood assumption).}

One drawback of using the PA metric is that it penalizes deviations from the reference distribution's mode. 
As a result, it may fail to capture the nuanced distributional differences that emerge when conveying uncertainty about events with varying severity or base rate~\citep{wallsten1986measuring,weber1990contextual,Willems2019VariabilityIT}.
However, in our experiments, we do not expect to observe systematic distributional differences, since the uncertainty expressions are used in neutral contexts that concern unknown people and extraneous events/facts. 
Nonetheless, we include an alternative to the PA metric that compares histograms of responses, \eg based on multiple responses from agents for a particular uncertainty expression $u$. In particular, we provide numerical results for histogram comparisons (using the Wasserstein distance between histograms) in Appendix \ref{app:additional-results}. %

As an additional measure of alignment between the reference distribution and the agent's distribution, we also compute the \textbf{Mean Absolute Error (MAE)}, for each uncertainty expression $u$, defined as the absolute difference between (i) the mean of the responses across statements involving $u$ for an agent, and (ii) the mean of the reference distribution for $u$, $P(k|u)$. 
We then average across the 14 expressions $u$ to get a single score per agent.

\section{Results}
\label{sec:results}
This section examines the ability of several well-known \llms to interpret uncertainty expressions.
We begin by assessing models' abilities to produce numerical responses that resemble human-like trends (\eg higher numerical responses assigned to higher-certainty expressions and vice-versa).
We then study the effect of prior knowledge in the perception of uncertainty of both humans and models.
We conclude with an assessment of the generalizability of our findings.

\subsection{How well do \llms perceive uncertainty?}
\label{ssec:rq1}

\begin{figure}[tb]
    \centering 
    \includegraphics[width=\linewidth]{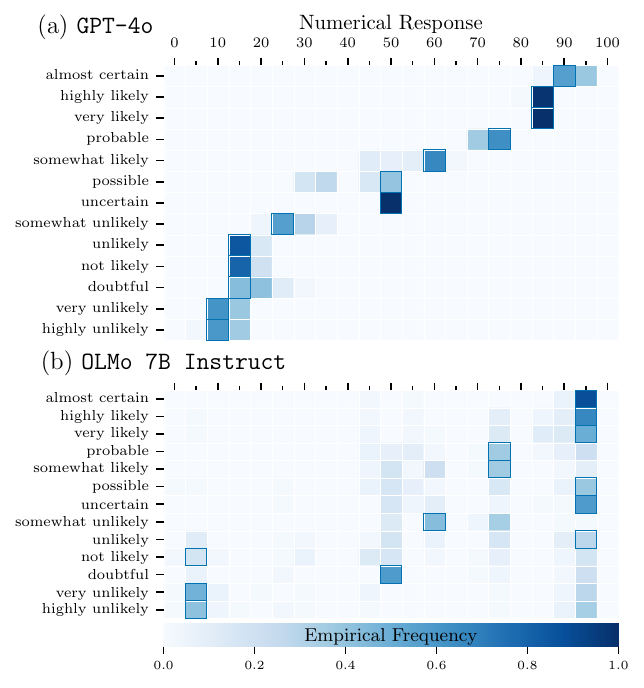}
    \caption{\textbf{Model empirical distributions of numerical responses per uncertainty expression in the non-verifiable setting (\texttt{LLM+NV})}. Highlighted boxes represent the mode value for each expression. Even though we found no evidence of explicit instruction tuning datasets focusing on uncertainty estimation tasks, these results suggest that \gptfo generally manifests human-like behavior, whereas \olmo does not.}
    \label{fig:histograms-nv-subset}
    \vspace{-1.3em}
\end{figure}

As established in prior work \citep{WALLSTEN1986571,fagen-ulmschneider2019,Willems2019VariabilityIT} and in our baseline experiment (Section \ref{sec:initalexp}), humans show population-level agreement in mapping uncertainty expressions to numerical responses. 
In this section, we assess whether \llms possess a similar ability to ascribe numerical responses to uncertainty expressions. 
To this end, we prompt \llms to provide responses for the same non-verifiable (NV) statements as in the baseline experiment.
Figure \ref{fig:histograms-nv-subset} shows the expression-wise histograms for these responses for two \llms: \gptfo and \olmo, each of which can be directly compared to the histogram for humans in Figure \ref{fig:hist-nv:humans}. 

Visually, we observe that \gptfo matches the human distributions well, with smaller variance per distribution, while \olmo is less aligned. 
In Figure \ref{fig:hist-nv} in Appendix \ref{app:additional-results}, we show that most \llms map uncertainty expressions to numerical responses in a manner consistent with human behavior, with higher values for expressions that are perceived by humans as higher-certainty (\eg ``almost certain,'' ``highly likely'') and lower values for lower-certainty expressions (\eg ``very unlikely''). 
Only two of the \llms evaluated, \olmo and \gemma, fail to reproduce this ``increasing'' pattern across expressions. 
One clear difference between humans and \llms is that the conditional distributions of \llms have lower entropy (or variance) relative to the human distributions, with the \llm distributions tending to be much more concentrated\footnote{Appendix \ref{app:ssec:add-results:interquartile-per-expr} shows evidence that greedy decoding results in lower variability in estimated distributions compared to humans. Additionally, while probabilistic decoding (\texttt{temperature=1}) generally increases variability for most models, this effect is not observed for \gptf.} on a small number of responses compared to the variance in responses from a population of humans.

These observations are reflected more precisely by the PA scores in Tables \ref{tab:agreement-nv} and \ref{tab:app:agreement-nv:with-confidence-intervals}.
We observe that larger and newer \llms (in particular, \gptf, \llama, and \gemini) perform especially well on this task under the PA metric, being at 85\% or above in terms of matching the modal scores that a human population assigns to each uncertainty expression.
In fact, 7 out of the 10 \llms evaluated are significantly better matched to population modal responses than are individual humans on average\footnote{The average performance of individual humans is represented by the Human Individual row in Table \ref{tab:agreement-nv}.}. 
This aligns with the high-level findings of \citet{Maloney2024-coordinationgame}, in particular, that the difference between the numerical responses of \texttt{GPT-4} and humans were similar to (or smaller than) inter-human differences. 
In the context of our experiments, these high scores reflect that \llms tend to be more consistent than individual humans in terms of agreement with aggregate human responses. 

The MAE scores in Table \ref{tab:agreement-nv} (lower is better) are highly anti-correlated with the PA scores and tell a similar story in terms of which models perform better. 
To provide a sense of scale, the MAE numbers are lower-bounded by 0 and upper-bounded by 25 (the expected MAE for random responses).
\begin{table}
\small
\centering
\caption{\textbf{Human-LLM agreement for non-verifiable statements}. 
Average Proportional Agreement (PA), PA as a fraction of the \textit{Human Mode} results (\% PA), and absolute error between mean responses (MAE). 
\emph{Human Mode} represents the mode of the human NV distribution, whereas \emph{Human Individual} represents the PA score of individual human responses relative to the population.
}
\label{tab:agreement-nv}
\begin{tabular}{l|cc|c}
\toprule
                & \textbf{PA} & \textbf{\% PA} & \textbf{MAE} \\
\midrule
Human Mode      & 27.6 & --- & ---  \\
Human Individual& $17.6$ & 63.8 & $8.91$ \\
\hdashline
\addlinespace
\chatgpt        & $19.7$     & 71.4      & $6.80$ \\
\gptf           & $24.4$     & 88.4      & $4.64$\\
\gptfo          & $18.9$     & 68.5      & $5.58$\\
\gemini         &  $25.4$    &  {92.0}   & $4.09$\\
\llamasmall     &  $17.8$    &  {64.5}   & ${11.99}$\\
\llama          &  $23.6$    &  {85.5}   & ${5.56}$\\
\mixtralmoe     &  $21.8$    &  {79.0}   & ${5.88}$\\
\mixtralmoelg   &  $21.8$    &  {79.0}   & ${7.20}$\\
\olmo           &  $11.1$    &  {40.2}   & ${21.41}$\\
\gemma          &  $8.1$       &  {29.3}   & ${20.17}$\\
\bottomrule
\end{tabular}
\end{table}

\subsection{Does knowledge affect uncertainty perceptions of \llms?}
\label{ssec:rq3}

In this section, we assess the extent to which \llms, and humans, are biased by their prior knowledge or beliefs in mapping uncertainty expressions to numerical responses. 
To investigate this question we collect responses from humans and \llms on our verifiable (V) dataset, which includes both true and false common-knowledge statements (based on correct or incorrect answers, respectively, to multiple-choice questions). 
We find that average PA scores for both humans and \llms are systematically lower for verifiable statements compared to the non-verifiable responses (Tables \ref{tab:agreement-v} and \ref{tab:app:agreement-v:with-confidence-intervals}). 
This suggests that prior knowledge about a statement\footnote{We validempirically ated our use of true correctness as a proxy for the LLMs’ beliefs in Appendix \ref{app:ssec:llm-belief-vs-statement-correctness}.} makes it more difficult to quantify the beliefs of someone else about that statement. 
While humans show a small drop in their PA score, this reduction in PA is particularly pronounced for \llms: all 10 \llms demonstrated a significant reduction in PA, averaging a 4.3 point drop in score (across all models), compared to a 0.9 point drop for humans.
\begin{table}[tb]
\small
\centering
\caption{\textbf{Human-LLM agreement for verifiable statements}. Average Proportional Agreement (PA), absolute error between mean responses (MAE), and the difference between these scores and those from the non-verifiable statements (Table \ref{tab:agreement-nv}) ($\Delta$ PA and $\Delta$ MAE, respectively). Again, \emph{Human Mode} represents the mode of the human NV distribution, whereas \emph{Human Individual} represents the average behavior across individual humans on the verifiable setting.
}
\label{tab:agreement-v}
\begin{tabular}{l|cc|cc}
\toprule
& \textbf{PA} & \textbf{$\Delta$ PA} & \textbf{MAE}& \textbf{$\Delta$ MAE} \\
\midrule
Human Mode &  {27.6} & --- & --- & ---\\
Human Individual & $16.7$ & -0.9 & $9.35$ & 0.44\\
\hdashline
\addlinespace
\chatgpt        & ${15.3}$ & -4.4        & ${8.57}$ & 1.77\\
\gptf           & ${22.1}$ & -2.3        & ${3.84}$ & -0.80 \\
\gptfo          & ${15.2}$ & -3.7        & ${7.05}$ & 1.47\\
\gemini         & ${21.3}$ & -4.1        & ${7.23}$ & 3.14 \\
\llamasmall     & ${10.1}$ & -7.7         & ${16.59}$ & 4.60 \\
\llama          & ${18.9}$ & {-4.7}      & ${13.73}$ & 8.17 \\
\mixtralmoe     & ${15.2}$  & -6.6       & ${12.23}$ & 6.35\\
\mixtralmoelg   & ${18.6}$  & -3.2       & ${9.78}$ & 2.58\\
\olmo           & ${7.6}$  & -3.5          & ${33.66}$ & 12.25\\
\gemma          & ${5.3}$  & -2.8          & ${25.07}$ & 4.9\\
\bottomrule
\end{tabular}
\vspace{-1.2em}
\end{table}

\begin{figure}[tb]
    \centering
    \includegraphics[width=\linewidth]{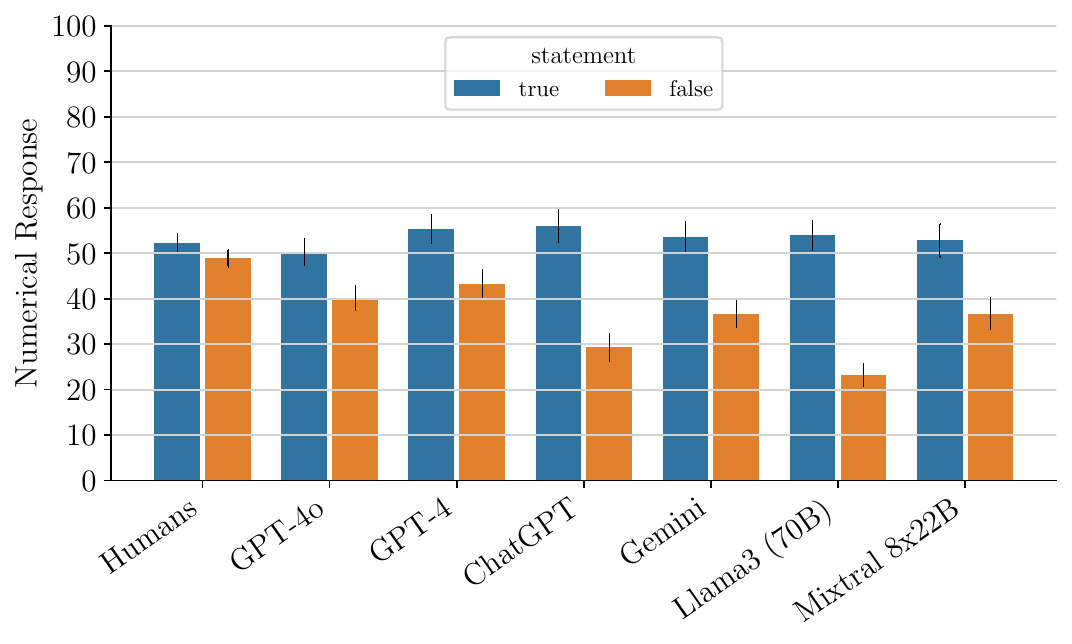}
    \caption{\textbf{Mean numerical response for the verifiable statements discriminated by truthfulness of statements}. The mean numerical responses produced by \llms when evaluated in the context of true statements is significantly larger than when evaluated with the false statements. This difference is much larger in magnitude than the difference shown by a human population.}
    \label{fig:bias-from-nv-to-v}
    \vspace{-1.em}
\end{figure}

To investigate these differences in more detail, we consider the mean response values produced by the 6 models exhibiting the highest PA score. 
These values differ systematically depending on whether the statement is true or false: across the 6 \llms in Figure \ref{fig:bias-from-nv-to-v}, the mean response is 7.0 points lower for false than true statements. 
This indicates that the models assign higher response values to the same uncertainty expression when they believe the associated statement is true than when they believe it to be false (providing a quantitative validation of the \chatgpt example in Section~\ref{sec:introduction}). 

\begin{figure}[tb]
    \small
    \centering
    \begin{subfigure}[b]{\linewidth}
        \centering
        \includegraphics[width=0.9\linewidth]{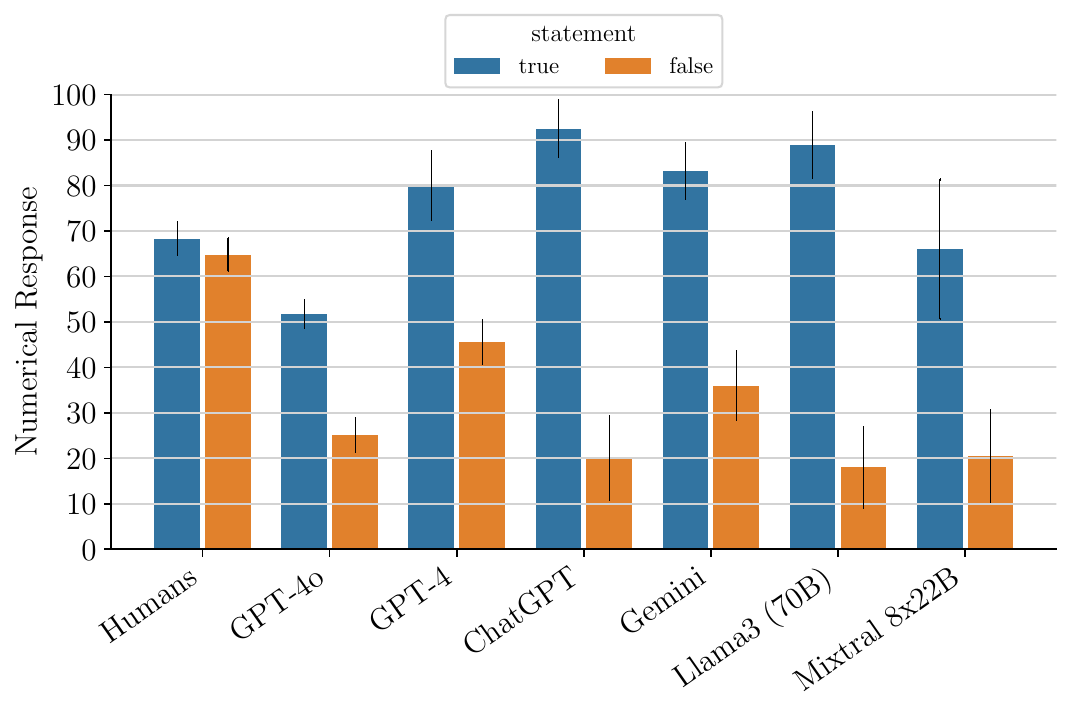}
        \caption{``possible''}
        \label{sfig:bias:possible}
    \end{subfigure}
    \vspace{-0.2em}
    \begin{subfigure}[b]{\linewidth}
        \centering
        \includegraphics[width=0.9\linewidth]{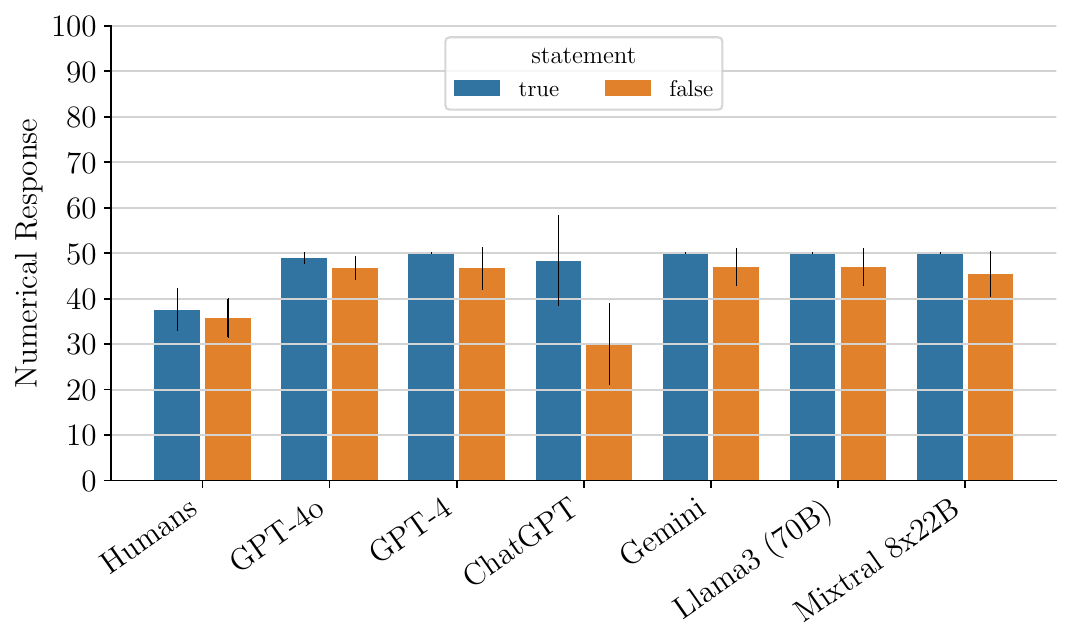}
        \caption{``uncertain''}
        \label{sfig:bias:uncertain}
    \end{subfigure}
    \vspace{-0.2em}
    \caption{\textbf{Mean response for verifiable statements discriminated by truthfulness of statements for three different uncertainty expressions}. We observe that the differences in mean numerical responses differ by uncertainty expression. Despite changes in magnitude, \chatgpt systematically exhibits the largest gap.}
    \vspace{-2em}
\label{fig:truevfalse_statement}
\end{figure}

Results for a subset of specific uncertainty expressions are shown in Figure \ref{fig:truevfalse_statement}. 
We observe that the prior-knowledge bias is remarkably different depending on the uncertainty expression: the difference between true and false statements is much higher (49.5 percentage points) for the expression ``possible'' than for the expression ``uncertain'' (for this expression most models are close to the average 5.7 percentage point difference). 

Overall, we find that all of the evaluated \llms demonstrate significant biases based on their prior knowledge, well beyond those of humans. 
Our results indicate that when an \llm believes a statement is false, it tends to perceive the speaker's certainty as low, regardless of the actual uncertainty expressed by the speaker (and vice versa), lacking ``theory of mind.''
Our findings in general are also consistent with recent results in the context of \llm reading comprehension, where the efficacy of \llm comprehension is sensitive to the degree of prior knowledge it has about the text it is analyzing~\cite{basmov2024llms}.

\subsection{How generalizable are our findings?}
\label{ssec:ablation:generalization-study}

In the previous sections, our analyses are conducted on a manually curated set of 120 statements, comprised of 60 NV statements and 60 V statements. 
To further validate our findings concerning \llms' prior knowledge biases, we re-assess the impact of knowledge in \llms' perceptual capabilities by obtaining their responses for 400 additional verifiable statements. 
Similarly to the original study, Figure \ref{fig:generalization-results-bias-shift-average} shows that, on average, all models except \gemma exhibit significant perceptual differences between true and false statements---between 5.87 (\olmo) and 17.26 (\llama) percentage points. 
While the magnitude of the difference is different across the two datasets (potentially due to semantic differences between the two QA datasets used to curate the verifiable statements), the directionality of the results with this larger dataset nonetheless  corroborates our knowledge bias finding by showing that these perceptual differences persist in a different context. %
See Appendix \ref{app:sec:generalization-results} for a more detailed description of the experimental setup and additional results.

\subsection{How does decoding impact our findings?}
\label{ssec:ablation:fractional-analysis}
The previous analyses employ greedy decoding (i.e., \texttt{temperature=0}) when  obtaining numerical responses from the \llms. 
In this section, we investigate the impact of the decoding technique in the models' abilities to perceive linguistic uncertainty, by considering richer probability information (\ie \texttt{temperature=1}) when obtaining the response.\footnote{This analysis requires full next-token probability information from an \llm, which is prohibitively expensive to obtain empirically through sampling as it would require a large sample size (per $(u, s, e)$) to obtain accurately. 
As a result, we limit our analysis to OpenAI models for which the top 20 next-token probabilities are available. 
See Appendix \ref{app:sec:fractional-analysis} for additional discussion on this topic.}

Table \ref{tab:fractional-avg-prop-agreement} summarizes the change in agreement between \llm and human responses between the verifiable and non-verifiable settings (in terms of change in PA and MAE) when using probabilistic decoding. 
Validating the results reported in Section \ref{ssec:rq3} with greedy decoding, we observe a clear difference in PA score between non-verifiable and verifiable statements when using probabilistic decoding. 
Further, comparing responses across true and false statements, we observe large mean response differences of 11.4, 11.7, and 27.9 percentage points for \gptfo, \gptf, and \chatgpt, respectively (see Figure \ref{fig:decoding:bias-knowledge} and Figure \ref{fig:fractional:mean-rated-prob} in the Appendix for a breakdown across expressions). 
Although \gptfo mean responses are considerably lower than in the greedy decoding setting (with a 20 percentage points drop), the gap between true and false statements persists. 
Ultimately, this analysis confirms the robustness of our previous findings to the decoding strategy.

\begin{table}
\small
\centering
\caption{\textbf{Differences in average proportional agreement and mean responses from non-verifiable to verifiable settings when considering probabilistic decoding (\texttt{temperature=1})}. Even with a different decoding, we observe the same decrease in \llm perceptions when comparing non-verifiable with verifiable settings.}
\label{tab:fractional-avg-prop-agreement}
\begin{tabular}{lcc}
\toprule
& \textbf{$\Delta$ PA} & \textbf{$\Delta$ MAE} \\
\midrule
\chatgpt &  {-3.6} &  {0.4} \\
\gptf &  {-3.0} &  {1.9} \\
\gptfo &  {-4.2} &  {-0.6} \\
\bottomrule
\end{tabular}
\vspace{-1em}
\end{table}

\begin{figure}[tb]
    \centering
    \includegraphics[width=\linewidth]{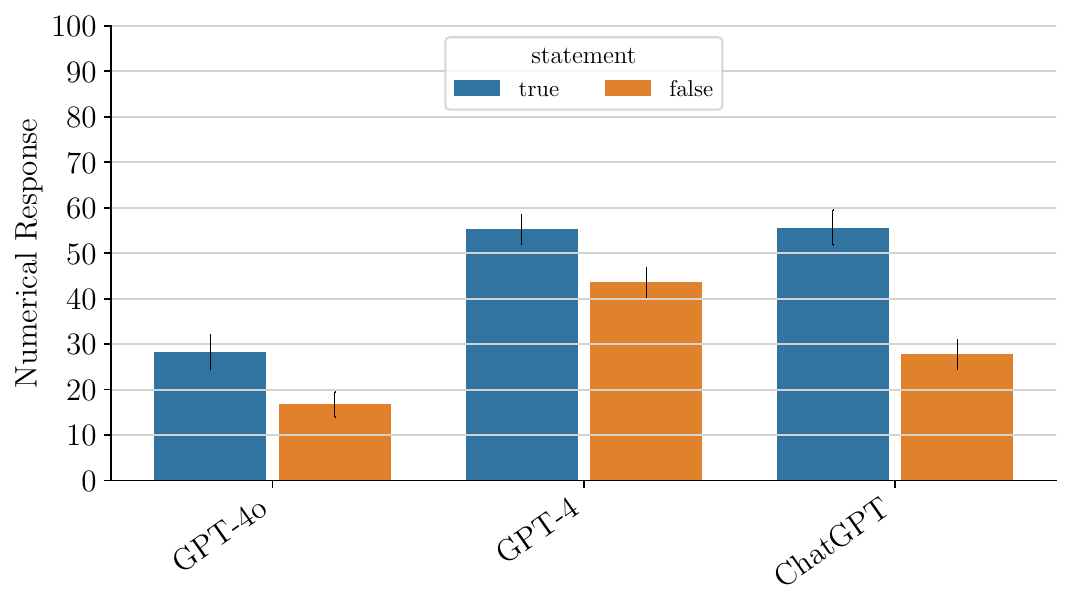}
    \caption{\textbf{Mean response on the verifiable statements discriminated by truthfulness of statements when decoding probabilistically \texttt{temperature=1}}. Despite being to a less extent, the prior-knowledge bias remains larger than that exhibited by human population.}
    \label{fig:decoding:bias-knowledge}
    \vspace{-1.5em}
\end{figure}

\section{Discussion}
\label{sec:discussion}

\paragraph{Theory of mind.} A growing body of work aims to assess the \textit{theory of mind} capabilities of \llms in different contexts (\eg \citep{street2024llms, verma2024theory, sap2022neural, zhou2023far}). 
The task of mapping uncertainty expressions to numerical responses, from the perspective of some speaker, is one component of a general theory of mind ability. 
Our results indicate that \llms have room for improvement in this area, in particular, that they are prone to confusing their own belief about a statement with the belief of someone else.

\paragraph{Connection to human behavior simulation using \llms.}
Our experiments reveal that, despite agreeing with population-level perceptions of linguistic uncertainty, models do not capture the full diversity of human behavior. 
While this lack of variability depends on the decoding algorithm used to extract the numerical response, we note that, for popular models like \gptf, the disparity persists regardless of the chosen algorithm.
Given the recent interest in using \llms to simulate human participants~\citep{aher-icml2023-simulability-using-llms,Gui2023TheCO,Dillion2023CanAL,Parketal2023,Namikoshi2024UsingLT}, our work raises important questions about whose opinions and behaviors are being simulated \citep{santurkar2023opinions,Motoki2023} and reveals a new dimension in which human and model diversity can differ.

\section{Conclusions}
\label{sec:conclusion}
We evaluate the abilities of \llms to interpret uncertainty in language and evaluate numerous models in this context.
Our results show that many \llms can competently map uncertainty expressions to numerical responses in a way that aligns with population-level human perceptions, although the responses they choose can be much less diverse than those by humans. 
Additionally, we find that \llms are more susceptible to conflating their own uncertainty about a statement with the statement speaker's uncertainty, resulting in output that is biased by the \llm's belief about the statement. 
By highlighting systematic inconsistencies related to the perceptions of linguistic uncertainty in the presence of knowledge, we shed light into overlooked model behaviors that are critical for understanding human-AI communication and downstream \llm performance.

\section*{Limitations}
\label{sec:limitations}

\noindent\emph{Biases in Interpreting Uncertainty Expressions:} Prior work has raised several concerns about the consistency of humans' interpretations of uncertainty expressions, demonstrating that they are subject to a number of biases and nuances. For example, people may conflate the speaker's confidence with the speaker's estimated uncertainty \citep{fleiner2024towards}, statements worded in terms of confidence (``I am almost certain'') or likelihood (``I believe it is almost certain'') are interpreted as primarily communicating different types of uncertainty (epistemic and aleatoric, respectively) \citep{ulkumen2016two}, and these statements are directional, emphasizing either the occurrence or non-occurrence of an event \citep{teigen2023dimensions}. Further, these interpretations are context-dependent, affected by factors including the individual's perception of the speaker and the severity of the event in question \citep{budescu1985consistency, COLLINS201867, bonnefon2006tactful, Juanchich-sirota2013, Brun1988, weber1990contextual}. Thus, in interpreting uncertainty expressions, high variability can occur both across \citep{finance-Zhang2023, COLLINS201867} and within \citep{clarke1992ratings, van2019communicating} individuals. In this paper we do not explore these dimensions of the interpretation of uncertainty expressions, which could limit the generality of our conclusions.

\noindent\emph{US Centric View:} In this paper we focus on a small set of uncertainty expressions in English and our baseline is drawn from participants located in the United States. Investigating the role that cultural and language differences play in communicating uncertainty is important future work that will help better characterize the downstream abilities of \llms for all users. The recent work by \citet{tang2024evaluationestimativeuncertaintylarge} represents a first step in this direction, but it does not account for the cultural context that would inform the meaning of a speaker~\citep{huang-yang-2023-culturally}. In particular, the meaning of ``probable'' and ``very likely'' may differ significantly between English and Chinese. Such cultural difference may suffice to explain the observed differences in perceptions measured between US population and \texttt{GPT-4} when prompted with Chinese contexts. 

\noindent\emph{Lack of Explanation:} Our results highlight the \llms' abilities to interpret uncertainty phrases in a way that agrees with population-level human distribution in the non-verifiable and to less extent in the verifiable setting. 
It is not clear why models have acquired this general ability, given that there are not similarly framed tasks in available instruction-tuning and human feedback datasets \citep{wang2022supernaturalinstructions,bai2022training}.
We hope that future work will explore the origins of this behavior.

\section*{Acknowledgments}

The authors would like to thank all the reviewers, the members from the UCI-NLP, DataLab, and MADLAB at UC Irvine for the provided feedback and insightful discussions regarding this project.
This research was supported in part by the National Science Foundation under awards RI-1900644 and CCRI-1925741, by CCRI and CAREER under the awards CNS-1925741 and IIS-2046873, and in part by the Hasso Plattner Institute (HPI) Research Center in Machine Learning and Data Science at the University of California, Irvine.
The views expressed in this paper are those of the authors and do not reflect the policy of the funding agencies.

\bibliography{literature-cogsci,literature-nlp,literature-models}

\appendix
\section{Motivating example}
\label{app:additional-motivation}

\begin{figure*}[tb]
    \centering
    \includegraphics[width=0.5\textwidth]{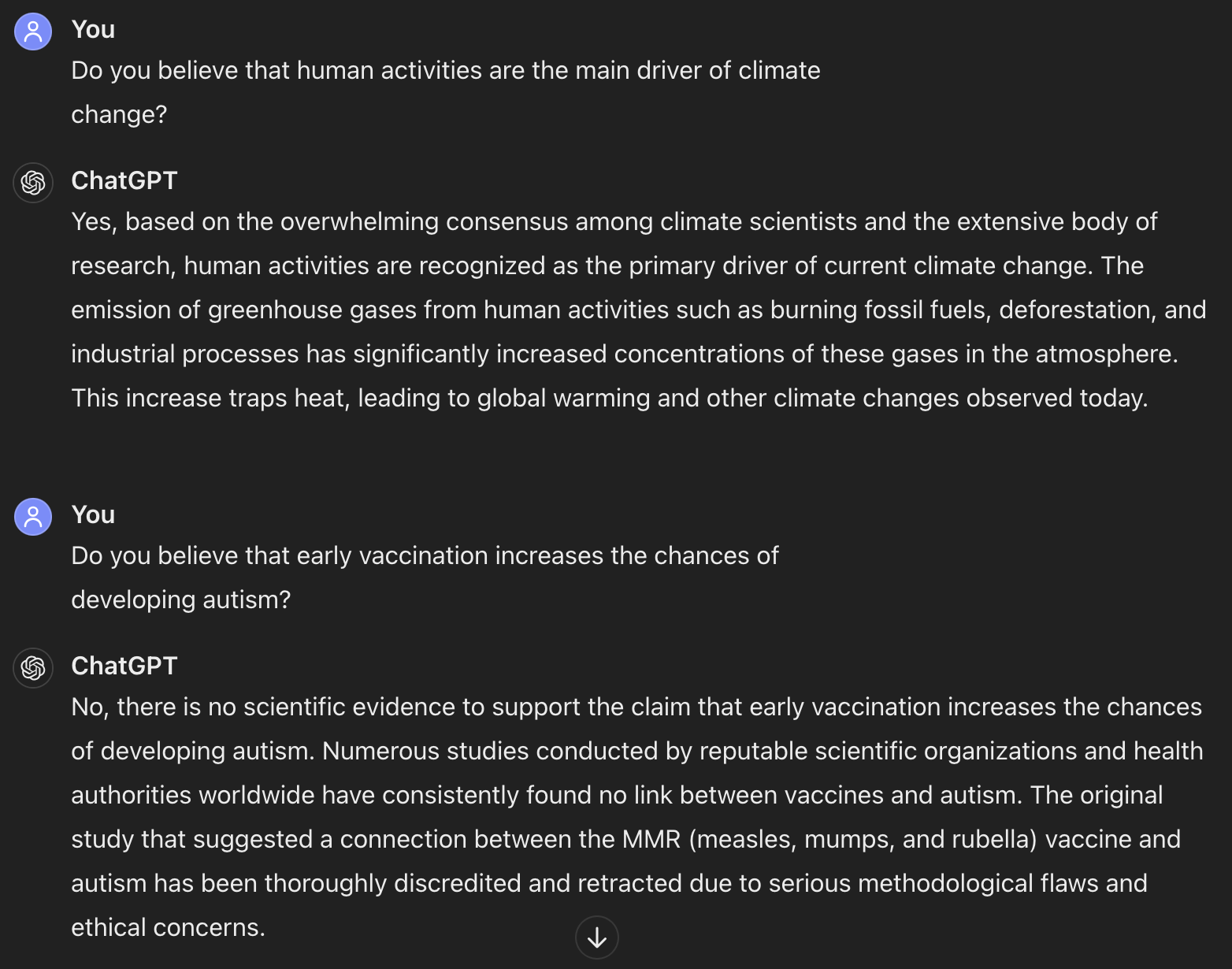}
    \caption{\textbf{Responses from \chatgpt when asked about its belief in the statements from Figure \ref{fig:gpt}}. \chatgpt agrees that ``human activities are the main driver of climate change,'' but disagrees with the statement that ``early vaccination increases the chances of developing autism''.}
    \label{fig:appdx:gpt_belief}
\end{figure*}

In the main paper, we discuss the divergence of model behavior in the use case of news headline generation when prompted with text containing uncertainty expressions. 
In particular, Figure \ref{fig:gpt} shows two different levels of conviction in the generated text: the first headline contains confident-sounding words like ``conclude'' and ``comprehensive'' whereas the second generated headline uses less confident language like ``suggests'' and ``possible''. 
To further understand the extent to which these differences could be explained by the models' own knowledge, we prompt the model a second time to assess its agreement or disagreement with the topic being discussed. 
Its answers are in Figure \ref{fig:appdx:gpt_belief}. 

\section{Datasets}
\label{app:dataset-selection}

To ensure that our results generalize and are not artifacts of the evaluation methodology and/or benchmarks used~\citep{selvam-etal-2023-tail,seshadri2022quantifying}, we intentionally include statements spanning multiple categories and different syntactic structures. 
Moreover, to simplify and avoid potential length-based artifacts, we opt for simple, short statements that communicate a single fact. 

\subsection{Non-Verifiable Statements}
\label{app:ssec:examples-non-verifiable}

As mentioned in Section \ref{sec:initalexp}, we manually craft the non-verifiable statements to be short ordinary statements that purposely lack grounding on specific contextual information. In doing so, we hypothesize that an external observer will not have confidence in the truth or falsity of the statement.
To ensure that our analysis covers a diverse set of plausible and realistic uses of uncertainty expressions, we create 15 statements for 4 different settings, including: 
(1) the forecasting of events;
(2) the communication of imperfect knowledge; 
(3) to communicate possession; and 
(4) to communicate preferences.
Below we list a random selection of 5 statements for each of the previous scenarios. 
Note that each of these statements are incorporated in the prompts listed in the main paper and the placeholders \verb|[[they]]| and \verb|[[their]]| are replaced by pronouns matching the gender of the statement speaker's name.

\textbf{\textit{Forecasting of future events}}. 
Uncertainty expressions are often used to communicate uncertainty about future events. 
\begin{itemize}
    \itemsep0em
    \item \verb|[[they]]| will buy a new watch this Thanksgiving weekend.
    \item \verb|[[they]]| will be offered a promotion this fall.
    \item the company will have another round of lay-offs by mid July.
    \item there will be vegetarian options at the barbecue.
    \item \verb|[[they]]| will visit New York over winter break.
\end{itemize}

\textbf{\textit{Imperfect knowledge}}.
Uncertainty expressions can also be used to communicate uncertainty imprecise information about events or outcomes. 

\begin{itemize}
    \itemsep0em
    \item the restaurant near \verb|[[their]]| apartment accepts reservations.
    \item the new museum is offering complimentary admission.
    \item there is a yoga studio within 2 miles of \verb|[[their]]| workplace.
    \item there are more than eighty students in the auditorium right now.
    \item the temperature in the office is at least 72 degrees Fahrenheit.
\end{itemize}

\textbf{\textit{Possession}}.
Alternatively, uncertainty expressions can be used to convey uncertainty about the belongings of acquaintances.

\begin{itemize}
\itemsep0em
\item \verb|[[their]]| boss owns a blue car.
\item \verb|[[their]]| friend has a leather jacket.
\item \verb|[[their]]| cousin has a vegetable garden.
\item \verb|[[their]]| classmate owns a guitar.
\item \verb|[[their]]| boss has a stereo amplifier.
\end{itemize}

\textbf{\textit{Preference}}.
Uncertainty expressions can be used to convey uncertainty about the preferences of acquaintances.

\begin{itemize}
\itemsep0em
\item \verb|[[their]]| cousin prefers spinach over broccoli.
\item \verb|[[their]]| boss prefers coffee over tea.
\item \verb|[[their]]| friend prefers running over cycling.
\item \verb|[[their]]| neighbor prefers the beach over the mountains.
\item \verb|[[their]]| coworker prefers reading books over watching movies.
\end{itemize}

\subsection{Verifiable statements.}
\label{app:ssec:examples-verifiable}
Unlike the non-verifiable statements, verifiable statements are created such that an external observer may have confidence in the truth or falsity of the statement.
One way to ensure that the both humans and \llms have high confidence in the truth or falsity of the statements is by focusing on popular general-knowledge facts, for example by focusing on trivia-like facts or student grade level science questions. 

\textbf{\textit{TriviaQA dataset.}} 
For the experiments in the main paper, we propose to create short verifiable statements using the set of questions provided by The Question Company (as described in Section \ref{sec:methodology}). 
In addition to covering multiple categories (\eg geography, entertainment, sports, astronomy), it also includes questions of varying difficulty (\ie easy to hard).
Specifically, the ``easy'' subset of questions is designed to test knowledge about simple and straightforward common facts, without requiring specialized knowledge.
For this reason, we choose to use the facts from this subset as a proxy for statements that an external observer is likely to know to be true or false.
In particular, we use all 3 sets of 10 easy questions, spanning the topics ``Cities/Geography,'' ``Art/History,'' and ``Science and Nature,'' and, for each question, we write one \textit{true} statement using the correct answer choice and one \textit{false} statement using the wrong choice (see examples in Table \ref{tab:dataset:trivia:verifiable}). 

\textbf{\textit{AI2-ARC dataset.}}
In addition to the 60 statements derived from the trivia dataset, we include an additional analysis using 400 verifiable statements (of which 200 are true and 200 are false). 
Using the same approach as previously described, we extract true and false verifiable statements from AI2-ARC~\citep{Clark2018ThinkYH}, an elementary school-level multiple-choice question answering dataset focused on science facts.  
Despite its simplicity, AI2-ARC has been an important benchmark to measure progress in \llms reasoning capabilities~ \citep{jiang2023mistral,openai2024gpt4,open-llm-leaderboard}, making it a relevant dataset to include in our analysis. 
Table \ref{tab:dataset:ai2arc:verifiable} lists 6 true statements and 6 false statements derived from the AI2-arc datasets.

\begin{table*}[tb]
\centering
\small
\caption{\textbf{Examples of true and false verifiable statements across different categories of the trivia dataset}. The true and false statements are created based on the correct and one of the incorrect choices of a trivia multiple-choice question answering dataset. The statements constitute short, simple, public knowledge facts.}
\label{tab:dataset:trivia:verifiable}
\begin{tabular}{p{0.10\textwidth} p{0.40\textwidth} p{0.40\textwidth} }
\toprule
\textbf{Category} & \textbf{True statement} & \textbf{False statement} \\
\midrule
\multirow{3}{=}{Cities \& Geography} & ``Great Britain directly borders \textbf{\textit{0}} countries.'' &  ``Great Britain directly borders \textbf{\textit{2}} countries.''\\
 & ``New York is known as the \textbf{\textit{Big Apple}}.'' & New York is known as the \textbf{\textit{Big Orange}}.''\\
 & ``the Colosseum, a famous landmark in Rome, was originally built as an \textbf{\textit{Amphitheatre}}.'' & ``the Colosseum, a famous landmark in Rome, was originally built as an \textbf{\textit{Cathedral}}.''\\
\addlinespace
\multirow{3}{=}{History \& Art} & ``the Mona Lisa is a famous painting by \textbf{\textit{Leonardo da Vinci}}.'' & ``the Mona Lisa is a famous painting by \textbf{\textit{Tintoretto}}.'' \\
    & ``the Scream is the best known painting by \textbf{\textit{Edvard Munch}}.'' & ``the Scream is the best known painting by \textbf{\textit{Jackson Pollock}}.''\\
    & ``\textbf{\textit{Andy Warhol}} became a famous artist in the 1960s for painting soup cans and soap boxes.'' & ``\textbf{\textit{Frida Kahlo}} became a famous artist in the 1960s for painting soup cans and soap boxes.'' \\
\addlinespace
\multirow{3}{=}{Science \& Nature} & ``\textbf{\textit{water}}’s chemical formula is H2O.'' & ``\textbf{\textit{carbon monoxide}}'s chemical formula is H2O.''\\
    & ``the nearest planet to the sun is \textbf{\textit{Mercury}}.'' & ``the nearest planet to the sun is \textbf{\textit{Mars}}.'' \\
    & ``\textbf{\textit{oG}} is a measure of the acidity or basicity of a substance.'' & ``\textbf{\textit{pH}} is a measure of the acidity or basicity of a substance.'' \\
\bottomrule
\end{tabular}
\end{table*}

\begin{table*}[tb]
\centering
\small
\caption{\textbf{Examples of true and false verifiable statements derived from the AI2-Arc dataset~\citep{Clark2018ThinkYH}}. The true and false statements are created based on the correct and one of the incorrect choices of the dataset. The statements constitute short school-level science facts. }
\label{tab:dataset:ai2arc:verifiable}
\begin{tabular}{p{0.10\textwidth} p{0.40\textwidth} p{0.40\textwidth}}
\toprule
\textbf{Category} & \textbf{True statement} & \textbf{False statement} \\
\midrule
\multirow{3}{=}{Easy} & ``\textbf{\textit{Growing}} and reproducing are two life processes that occur in both plants and humans.'' & ``\textbf{\textit{Germinating}} and reproducing are two life processes that occur in both plants and humans.'' \\
    & ``A light year refers to the \textbf{\textit{distance light travels in one year}}.'' & ``A light year refers to the \textbf{\textit{time it takes light to travel from Earth to the Sun}}.''\\
    & ``Carbon dioxide produced by cells is removed from the body primarily by the \textbf{\textit{respiratory system}}.'' & ``Carbon dioxide produced by cells is removed from the body primarily by the \textbf{\textit{immune system}}.'' \\
\addlinespace
\multirow{3}{=}{Challenge} & ``\textbf{\textit{Trees}} are a renewable natural resource that can be replenished over a period of time'' & ``\textbf{\textit{Coal}} is a renewable natural resource that can be replenished over a period of time.'' \\
    & ``When cold temperatures are produced in a chemical reaction, the reaction is known as \textbf{\textit{endothermic}}.'' & ``When cold temperatures are produced in a chemical reaction, the reaction is known as \textbf{\textit{exothermic}}.''\\
    & ``\textbf{\textit{Swimming fast}} is an adaptive characteristic that helps dolphins survive life in the ocean.'' & ``\textbf{\textit{Traveling alone}} is an adaptive characteristic that helps dolphins survive life in the ocean.'' \\
\bottomrule
\end{tabular}
\end{table*}

\subsection{Statement Correctness vs \llms' beliefs}
\label{app:ssec:llm-belief-vs-statement-correctness}
A key assumption underlying our selection of verifiable datasets is that, because our statements concern simple common facts, \textit{\llms must know whether the statements are true or false}. 
In this section, we conduct an empirical validation of this assumption by soliciting the \llms' beliefs about each of the statements.

\paragraph{Methodology.}
Prompting is commonly used as a way to assess \llms' knowledge. 
Given a question-answer pair, one common approach is to assume that a \llm \textbf{\textit{knows}} the answer to $q$ is $a$ if it generates $a$ when prompted to answer $q$~\citep{gekhman2024doesfinetuningllmsnew,kadavath2022languagemodelsmostlyknow,manakul-etal-2023-selfcheckgpt}. 
Likewise, we can adopt this approach to determine an \llm's belief about the truth or falsity of a statement by asking the \llm about the veracity of a specific statement. 
To this end, we use the prompt:

``\textit{Question: What is the veracity of the statement: "[[statement]]"?\textbackslash nChoose from the following options: True, False, Unknown\textbackslash nWrite only one of the answer choices and nothing else.}''   

Using the prompt above, we empirically validate our assumption for four different \llms: \gemini, \chatgpt, \gptf, and \gptfo. 
In particular, for every statement in non-verifiable and verifiable settings, we prompt each \llm to generate \texttt{n\_samples=7} using \texttt{temperature=0.5}~\citep{wang-etal-2023-self-instruct}. 
For each prompt, we compute the relative frequency of each of the possible responses `true', `false', and `unknown' and use the mode response as the final prediction. 

\paragraph{Metrics.}
Our goal is to provide supporting evidence that the models' belief of correctness differs for the three different settings evaluated in our experiments, in particular, we want to show that: 
(1) \llms do not know the veracity of the non-verifiable statements;
(2) \llms knows which of the verifiable statements are true (dubbed ``verifiable true'') and which of them are false (dubbed ``verifiable false'').
To gauge model correctness, we report the average accuracy of the model. In particular, we consider a model to be correct (or accurate) if it generates 
``unknown'' when prompted about the veracity of the non-verifiable statement; 
``true'' when prompted about the veracity of a true verifiable statement;
and ``false'' when prompted about the veracity of a false statement. 
To differentiate the accuracy of the models across the three different scenarios, we designate the models' accuracy in each setting as $\text{Acc}_{NV}$,$\text{Acc}_{VT}$, and $\text{Acc}_{VF}$, respectively.

\paragraph{Results.}
Table \ref{tab:results:correctness-vs-llm-belief} shows the average accuracy results of applying the methodology described previously to four different \llms.
Overall, our results show that \textit{\llms are able to differentiate between verifiable and non-verifiable settings}. Specifically, \gptf and \gptfo abstain from assigning a veracity judgment to the non-verifiable statements in about >99\% of the examples, while correctly judging 100\% of the true verifiable statements and >93\% of the false verifiable statements.

Overall, we find that \textit{models are able to correctly discriminate between true and false statements (>90\% accuracy)} which corroborates the use of true correctness of the statements as a proxy for the \llms' beliefs. 

\begin{table*}[tb]
    \centering
    \begin{tabular}{lcccc}
    \toprule
         & \textbf{Avg $\text{Acc}$ (\%)} & \textbf{$\text{Acc}_{NV}$ (\%)} & \textbf{$\text{Acc}_{VT}$ (\%)}  & \textbf{$\text{Acc}_{VF}$ (\%)} \\
    Num examples & 170 & 110 & 30 & 30 \\
    \midrule
    \chatgpt    & 90.00 & 90.00     & 90.00  & 90.00 \\ 
    \gptf       & 98.82 & 100.00    & 100.00 & 93.33 \\
    \gptfo      & 98.82 & 99.09     & 100.00 & 96.67 \\ 
    \gemini     & 90.00 & 90.91     & 93.33  & 83.33 \\
    \bottomrule
\end{tabular}
\caption{\textbf{Accuracy of \llms' beliefs about the veracity of the non-verifiable (NV) and verifiable statements (V) in the main experiment.} Overall, models are very accurate in differentiating between non-verifiable and verifiable statements. Moreover, we observe that models are able are able to correctly discriminate between true and false statements (>90\% accuracy), which corroborate the hypothesis that \llms know the correctness of the statements they are tested with.}
\label{tab:results:correctness-vs-llm-belief}
\end{table*}

\section{Experiment Details}
\label{app:ExperimentDetails}

This section describes in greater detail various aspects of the experiments conducted in this paper, including the list of uncertainty expressions, the name selection strategy, the list of prompts, a list of statements, as well as additional details on the human experiments.

\subsection{Uncertainty Expressions}
\label{app:ssec:uncertainty-expressions}

The uncertainty expressions are a subset of the expressions proposed in \citeauthor{wallsten1986measuring,wallsten2008intel,Willems2019VariabilityIT,Fore_2019}. 
The final list of uncertainty expressions used in this paper is listed below:

\begin{itemize}
\item almost certain, highly likely, very likely, likely, probable, somewhat likely, somewhat unlikely, uncertain, possible, unlikely, not likely, doubtful, very unlikely, highly unlikely
\end{itemize}

\subsection{Name Selection}
\label{app:ssec:name-selection}

All names used in our experiments were collected from a random name generator\footnote{\url{https://randomwordgenerator.com/name.php}, last accessed on March 26th, 2024.}, which we ran iteratively until we obtained 32 unique names, half of each biological gender (as determined by the random generator). 

\begin{itemize}
    \item \textbf{Female names}: ``Amanda'', ``Bonnie'', ``Camille'', ``Catherine'', ``Cheri'', ``Ethel'', ``Gabriela'', ``Jacquelyn'', ``Jessica'', ``Laura'', ``Olga'', ``Roxanne'', ``Silvia'', ``Tara'', ``Violet''
    \item \textbf{Male names}: ``Brendan'', ``Bruce'', ``David'', ``Gary'', ``Isaac'', ``Jeffery'', ``Joey'', ``Johnnie'', ``Kenny'', ``Lance'', ``Marco'', ``Mike'', ``Nathan'', ``Nick'', ``Raul''
\end{itemize}

\subsection{Human Experiments}
\label{app:humanexperiments}

Human responses were collected using Prolific (\url{https://www.prolific.com/}). 
We recruited 100 participants for the non-verifiable experiment and 100 different participants for the second verifiable experiment. 
One of the 100 responses was not received due to a technical issue in both the first and second experiment, leaving a total of 99 responses for each. 
We recruited participants whose first language was English that were located in the United States. 
Participants were paid \$2 for completing the study and the average completion time was 8 minutes and 48 sections; the average payment rate was \$13.64/hour.  
The University of California, Irvine Institutional Review Board (IRB) approved the experimental protocol.  
Prior to the experiment, participants were given detailed instructions outlining the experimental procedure as well as how to understand and interact with the user interface. 
Participants were asked to sign an integrity pledge after reading all of the instructions, stating that they would complete the experiment to the best of their abilities. After submitting their integrity pledge, participants were granted access to the experiment.

We filtered out low-quality responses with the following procedure. For each participant, we computed the Spearman correlation between the participant's responses and the overall ranking of uncertainty statements in the non-verifiable experiment. We removed participants with $\rho < 0.2$, a threshold chosen empirically to filter out only no-signal, spam-like responses. This filter removed 5 participants in the first experiment and 10 in the second experiment. In total, we remain with 94 participants in the non-verifiable experiment and 89 in the verifiable experiment.

\begin{figure}[tb]
    \centering
    \includegraphics[width=\linewidth]{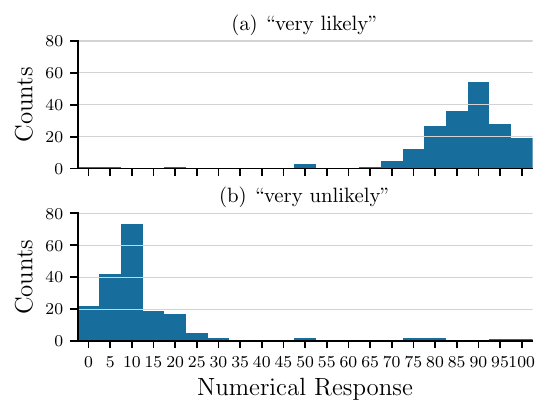}
    \caption{\textbf{Histogram of human participant responses for non-verifiable statements for two uncertainty expressions}. The modes of the two distributions sum to 100\%, suggesting symmetry in the population-level interpretation of the two expressions.}
    \label{fig:histograms-detail}
\end{figure}

\subsection{Prompts}
\label{app:ssec:prompts}

In our main paper, we conduct experiments using 2 demonstrations, since this better replicates the human setup. 
We use the prompt listed in Figure \ref{fig:prompt:2-shot} to elicit the model's numerical response, and use the format described in Figure \ref{fig:prompt:icl-demonstration} to represent the demonstrations in the prompt.

\begin{figure*}[tp]
\centering
\begin{tcolorbox}[colback=gray!10,%
                  colframe=gray!20,%
                  width=\linewidth,%
                  arc=1mm, auto outer arc,
                 ]
In this experiment, you will be shown text reflecting an individual's beliefs about a statement. You will then be asked to judge, in your opinion, the strength of that individual's belief. To do so, you will provide a probability (expressed as a percentage) where:\\
- The number 0 indicates that the individual believes with certainty that the statement is false.\\
- The number 100 indicates that the individual believes with certainty that the statement is true.\\~\\
You will be shown two example question and response pairs below to familiarize you with the experiment setup. After the examples, you will be asked to answer a new question.
\\~\\
\#\# Example Question 1\\
\verb|[[demonstration1]]|\\
Choose the answer from the following options:\\
0, 5, 10, 15, 20, 25, 30, 35, 40, 45, 50, 55, 60, 65, 70, 75, 80, 85, 90, 95, 100\\
Correct answer: 100
\\~\\
\#\# Example Question 2\\
\verb|[[demonstration2]]|\\
Choose the answer from the following options:\\
0, 5, 10, 15, 20, 25, 30, 35, 40, 45, 50, 55, 60, 65, 70, 75, 80, 85, 90, 95, 100\\
Correct answer: 100
\\~\\
\#\# Question\\
Given the examples before, answer the following question by writing a single number as the answer.\\~\\
\verb|[[test-example]]|\\
Choose the answer from the following options:\\
0, 5, 10, 15, 20, 25, 30, 35, 40, 45, 50, 55, 60, 65, 70, 75, 80, 85, 90, 95, 100
Correct answer:
\end{tcolorbox}
\caption{\textbf{Prompt used to conduct the experiments in the 2-shot setting}. \texttt{[[demonstration1]]}, \texttt{[[demonstration2]]}, and \texttt{[[text-example]]} are placeholders that will be composed of a statement (or event), question, and individual instructions. The prompt closely reflects the interface presented to the participants of the user studies. Similarly to the human interface, we encourage LLMs to select a multiple of 5 in the range [0, 100].}
\label{fig:prompt:2-shot}
\end{figure*}

\begin{figure*}[tb]
\centering
\begin{tcolorbox}[colback=orange!10,%
                  colframe=orange!20,%
                  width=\linewidth,%
                  arc=1mm, auto outer arc,
                 ]      
Sentence: ``\verb|[[speaker]]| believes it's \verb|[[uncertainty]]| that \verb|[[statement]]|.'' \\ 
\\
Question: Provide a number between 0 and 100 that reflects, from \verb|[[speaker]]|'s perspective, the probability (expressed as a percentage) that \verb|[[statement]]|.\\
- 0 indicates \verb|[[speaker]]| believes there is a 0\% chance that \verb|[[statement]]|.\\
\xspace\xspace - 100 indicates \verb|[[speaker]]| believes there is a 100\% chance that \verb|[[statement]]|.
\end{tcolorbox}
\caption{\textbf{Structure of a demonstration (or exemplar), as defined in the 2-shot prompt}. The placeholders \texttt{[[speaker]]}, \texttt{[[uncertainty]]}, and \texttt{[[statement]]} are replaced by gendered names, uncertainty expressions, and statements, respectively.}
\label{fig:prompt:icl-demonstration}
\end{figure*}

During the course of our experiments, we carried experiments with varying assumptions: non-verifiable setup assessed models (and humans) perceptions in the absence of strong prior knowledge about the statements, whereas the verifiable setup focused on the evaluation of the same perceptions when knowledge was present. 
We used two different sets of exemplars (or demonstrations) in our experiments to reflect these differences.  
The set of exemplars used in the non-verifiable setting are defined in terms of the following:

\begin{itemize}
    \item \texttt{speaker}: ``Kathleen'', \texttt{uncertainty}: ``impossible'', \texttt{statement}: ``the cafe made a profit in the last 6 months''.
    \item \texttt{speaker}: ``Cedric'', \texttt{uncertainty}: ``certain'', \texttt{statement}: ``the new treatment will improve the patient's condition''.
\end{itemize}

Conversely, for the set of exemplars used in the \textbf{verifiable} setting, we carried a  preliminary experiment to determine whether the truthness/falsehood of the examples and their ordering had a significant impact in models' performance. We used the two pairs of statements (``only some metals can conduct electricity,'' ``all metals can conduct electricity'') and (``the Sun orbits around the planet Earth,'' ``the planet Earth orbits around the Sun'') in the preliminary experiments.
We found negligible differences in the obtained distribution of numerical responses (as emphasized in Tables \ref{tab:app:demonstrations-greedy} and \ref{tab:app:demonstrations-prob}). As a result, we report the results using a false statement as the first example and a true statement as the second example in the prompt (denoted \texttt{FT}). These were associated with the following speaker names and uncertainty expressions:

\begin{itemize}
    \item \texttt{speaker}: ``Kathleen'', \texttt{uncertainty}: ``impossible'', \texttt{statement}: ``the Sun orbits around the planet Earth''.
    \item \texttt{speaker}: ``Cedric'', \texttt{uncertainty}: ``certain'', \texttt{statement}: ``all metals can conduct electricity''.
\end{itemize}

\begin{table}[tb]
\centering
\caption{\textbf{Average distributional difference in \llms conditional distributions when estimated based on different true/false configurations of the examples in the verifiable prompt}. We report the Wasserstein distance of the estimated conditional distributions averaged over the empirical distributions obtained using the other configurations and greedy decoding algorithm (see Table \ref{tab:app:demonstrations-prob} for results using the probabilistic decoding). 
On average, we observe that changing the veracity of the examples in the prompt has minimal impact on the final distributions. Moreover, we explicitly avoid using the configurations \texttt{FF} or \texttt{TT} to avoid majority label bias~\citep{pmlr-v139-zhao21c}.}
\label{tab:app:demonstrations-greedy}
\begin{tabular}{lccc}
\toprule
 & \chatgpt & \gptf & \gptfo \\
\midrule
\texttt{FF} & 2.33 & 1.31 & 0.92 \\ 
\texttt{FT} & 3.20 & 1.25 & 0.85 \\
\texttt{TF} & 2.93 & 1.32 & 0.88 \\ 
\texttt{TT} & 2.31 & 1.38 & 0.90 \\
\bottomrule
\end{tabular}
\end{table}

\begin{table}
\centering
\caption{\textbf{Average distributional difference in \llms conditional distributions when estimated based on different true/false configurations of the examples in the verifiable prompt}. On average, we observe that changing the veracity of the examples has minimal impact on the final distributions.}
\label{tab:app:demonstrations-prob}
\begin{tabular}{lccc}
\toprule
 & \chatgpt & \gptf & \gptfo \\
\midrule
\texttt{FF} & 1.91 & 1.53 & 1.74 \\ 
\texttt{FT} & 3.01 & 1.36 & 1.36 \\
\texttt{TF} & 2.77 & 1.33 & 1.81 \\ 
\texttt{TT} & 1.91 & 1.48 & 1.39 \\
\bottomrule
\end{tabular}
\end{table}

\subsection{Language Models}

Our evaluation concerns the study of instruction-tuned \llms, some of which are accessible through black-box APIs and others through the use of the HuggingFace Python package.
We use OpenAI to obtain the results for \chatgpt, \gptf, and \gptfo; Google's Vertex AI API to obtain results for \gemini, and TogetherAI\footnote{https://www.together.ai/} to run \llama, \mixtralmoe, \mixtralmoelg, and DBRX. 
We run \llamasmall \olmo and \gemma locally on a single GPU 8 RTX A6000 (48 GB).
All experiments were conducted from April through June 2024.

For simplicity, we have abbreviated the name of the evaluated models, removing information about the size and version of the model. 
For reproducibility, we list below the mapping from model name to exact version of the model used:

\begin{itemize}
\itemsep0em
\item {\small \chatgpt: \texttt{gpt-3.5-turbo-0125}}
\item {\small \gptf: \texttt{gpt-4-turbo-2024-04-09}}
\item {\small \gptfo: \texttt{gpt-4o-2024-05-13}}
\item {\small \llamasmall: \texttt{meta-llama/Meta-Llama-3-8B-Instruct}}
\item {\small \llama: \texttt{meta-llama/Llama-3-70b-chat-hf} }
\item {\small \gemini: \texttt{models/gemini-pro}}
\item {\small \mixtralmoe: \texttt{mistralai/Mixtral-8x7B-Instruct-v0.1}}
\item {\small \mixtralmoelg: \texttt{mistralai/Mixtral-8x22B-Instruct-v0.1}}
\item {\small \gemma: \texttt{google/gemma-1.1-2b-it}} (we found \gemma to respond better empirically to the prompts than its 7B version, which tended to extrapolate the few-shot instructions with additional examples).
\item {\small \olmo: \texttt{allenai/OLMo-7B-Instruct}}
\end{itemize}

\section{Extracting \llms Numerical Responses}
\label{app:sec:extract-llm-numerical-response}

In this section, we outline the methodologies used to extract the numerical responses from auto-regressive \llms.
In an ideal world, given an uncertainty expression $u$, the model's conditional distribution $\hat{P}_{\mathrm{model}}(k|u)$, would be fully observable. 
However, information about $\hat{P}_{\mathrm{model}}$ is seldom available, for models are either (1) served through opaque closed-source APIs or (2) too large, often requiring large amounts of compute to estimate the full distribution. 
To circumvent such limitations, one idea is to empirically estimate such probabilities using sampling-based approaches (\eg self-consistency~\citep{wang2023selfconsistency}) or using greedy decoded sequences over multiple examples.
In the ensuing sections, we describe the three different strategies considered in our work to estimate the \llms' empirical distributions.

\subsection{Method 1: Full Probability Approach}
\label{app:ssec:extract-llm-numerical-response-full-prob}

The ``full'' methodology requires access to the next-token probability distribution of an auto-regressive \llm and, as a result, is currently applicable to open source models. 
Because we use the models' probability to compute the probability of producing any integer between 0 and 100, it is also our most time-consuming approach\footnote{Running \llama across 5x GPU 8 RTX A6000 (48 GB) for 900 examples required approximately 9 full days. For that reason, we only apply this strategy to a few \llms.}, requiring 101 model calls to obtain the \textit{full} probability. 

\textbf{Methodology. } 
An important aspect to account for when using \llms to estimate the probability distribution over the set of integers ranging from 0 to 100 is the \llm's tokenizer.
Specifically, models, such as \gemma, \llama, and \olmo, use single-digit tokenization~\citep{singh2024tokenization}, which implies that the textual representation of 100 (represented as ``100'') is tokenized into at least three individual tokens (\ie ``1'', ``0'', ``0'') and not in a single token (\eg ``100'').
In terms of probabilities, this is a challenge, since many of these \llms generate digits in a left-to-right fashion and, by design, the probabilities of any number in $[10, 100]$ are always less than or equal to the probability of the first constituting digit (\eg probability of ``10'' subsumes the probability of ``1'' since in order to generate the string ``10'' the model needs to generate the string ``1'' first).
Since we are interested in knowing the true independent probability assigned to every integer between 0 and 100, we need to adjust the \llms' probability.
Therefore, instead of reporting the probability that a number $i \in [0, 100]$ occurs, we compute the probability that $i$ occurs and is not followed by a number $j \in [0, 9]$:

$$
p(x_t=i|x_{<t}) - \sum_{j=0}^{9} p(x_t=i, x_{t+1}=j|x_{<t}),
$$

\noindent where $p$ represents the \llm's next-token probability distribution. 
In sum, given a prompt parameterized with a speaker $s$, uncertainty expression $u$, statement $e$, and a prompt that combines these parameters, denoted $\text{prompt}(s, u, e)$ we use the expression above to obtain the full probability distribution for all $k \in 0, ..., 100$. 
We denote this corrected probability distribution as $p(k|\text{prompt}(s, u, e))$ and we refer to the probability mass that falls outside the set of strings $k \in 0, ..., 100$ as $\bar{p}$.

\textbf{\textit{Constructing the Greedy Histogram. }} 
Unlike sampling-based greedy decoding algorithms, we restrict the selection of the arg-max to the set of strings representing the numbers between $[0, 100]$, as opposed to sampling a series of tokens using \texttt{temperature=0}. 
In other words, we constrain the greedy decoding to be any of the sequences in \{``0'', ``1'', ..., ``100''\}, regardless of how little probability\footnote{We found this approach to be particular sensitive to the instruction format used. For example, in earlier iterations of this work we used Llama-2 and Gemma (7B) but found them to be particular sensitive to the whitespaces provides in the prompt.} is assigned to any of these numerical sequences.
For every triplet $(u, s, e)$, we obtain the the arg-max prediction and then assign it to a bin $0, 5, 10, ..., 100$. 

\textbf{\textit{Constructing the Probabilistic Histogram. }}
The probabilistic histogram benefits from the probability information that was computed previously for a specific triplet $(u, s, e)$. 
In particular, for a specific example we accumulate all probability values in the corresponding bins $0, 5, ..., 100$. 
The remaining probability mass $\bar{p}$ is assigned to a default bin ``-1''. 
In other words, the bin ``-1'' will accumulate the probability of a number in [0, 100] not following the specified prompt. 
While we could have normalized the probabilities of $0, 5, ..., 100$ to sum to 1, we decided to add a ``-1'' bin to allow for a fair comparison with the top-k approach, where only part of the probability distribution (\eg top-20 next probability distribution values are revealed), as is the case for the OpenAI models. 
After accumulating the probability over all $(u, s, e)$ triplets, we normalize the histogram by dividing by the number of triplets.

\subsection{Method 2: Top-k approach}
\label{app:ssec:extract-llm-numerical-response-top-k}

This method leverages information about the top-k values of the next-token probability information. 
Applicable to OpenAI models, namely \chatgpt, \gptf, and \gptfo, this approach allow us to obtain richer probability information with a single API call.

\textbf{\textit{Methodology. }}
This method requires two properties to be satisfied:
(1) numbers between 0 and 100 must be encoded with one single token each (\ie each integer is represented with a single token), and (2) exponentiating the log probabilities returned by the API must lead to a valid probability distribution (\ie numbers obtained for different prompts will be comparable to one another). 
We validate that the first requirement is satisfied by OpenAI models.

\textbf{\textit{Constructing the Greedy Histogram. }} 
Unlike traditional greedy decoding, we condition the selection of the arg-max prediction over numbers the top-k (k=20 for OpenAI). That is, we select the most likely number that is present in the top-20 predicted tokens. 
If no number is present in the top-20 tokens, we assign a default value of `-1'. 

\textbf{\textit{Constructing the Probabilistic Histogram. }}
Like the ``full'' probability approach, we construct the probabilistic histogram by accumulating the probabilistic information that we gather with each inference call: 
For a given triplet $(u, s, e)$, we make an API call and obtain probability about the next 20 tokens. 
If any of these strings represents a number between 0 and 100, we exponentiate it to obtain a probability value, and accumulate the probability in the corresponding bin. Any remaining probability that is not assigned to a number in the top-20 is accumulated in the `-1' bin.
After accumulating the probability over all $(u, s, e)$ triplets, we normalize the histogram by dividing by the number of triplets.

\begin{figure*}[tb]
    \centering
    \begin{subfigure}[b]{0.48\textwidth}
        \centering
        \includegraphics[width=\textwidth]{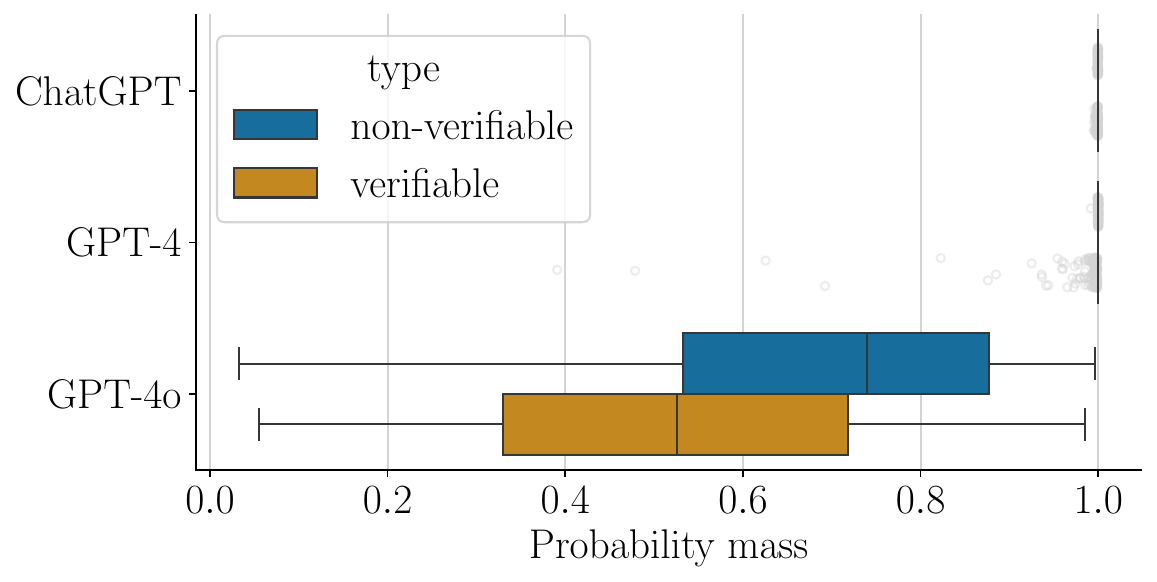}
        \caption{Total probability mass assigned to a number across models and settings.}
        \label{fig:probability-mass-is-a-number}
    \end{subfigure}
    \hfill
    \begin{subfigure}[b]{0.48\textwidth}
        \centering
        \includegraphics[width=\textwidth]{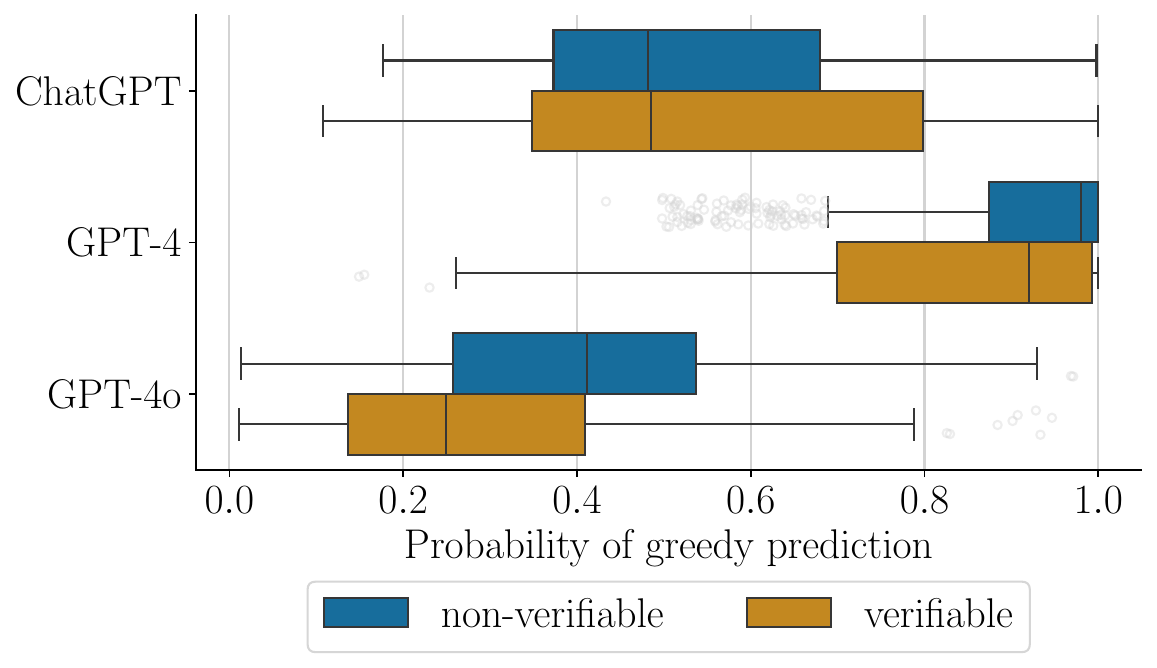}
        \caption{Probability mass of greedy prediction.}
        \label{fig:probability-mass-greedy-number}
    \end{subfigure}
    \caption{\textbf{Differences in the probability mass as determined by OpenAI models on the top-20 tokens}. We report these values across all statements (n=840) in the verifiable and non-verifiable settings. We observe that numbers account for the majority of probability mass in \chatgpt and \gptf. Upon analysis we found lower probability mass assigned by \gptf to be correlated with the appearance of the words ``Given,'' ``The,'' and ``And''.}
\end{figure*}

\subsection{Method 3: Sampling-based approach}
\label{app:ssec:extract-llm-numerical-response-sampling}

Some models are provided through black-box APIs with no access to next-token probability information. Ideally, one could estimate the next-token probability information by continuously sample, but doing so would be too costly. In the main paper, we only run sampling based approaches using a single sample.

\textbf{\textit{Constructing the Greedy Histogram. }}
This approach differs from the previous two in that it is unconstrained sampling-based approach, which means that the next immediate token may not be a number. 
Moreover, it is also agnostic to the tokenization. 
It many cases, the model may generate some text before it actually produces the answer ``The answer is 100'' or ``**Sure, here is the answer: 55''. 
To accommodate for such scenarios, we consider the arg-max prediction to be the first number between $0$ and $100$ to be mentioned in the model's response. Otherwise we assume the arg-max prediction to be `-1'.

\textbf{\textit{Constructing the Probabilistic Histogram. }}
Due to budget constraints, we could not reliably estimate a probabilistic histogram via the sampling-based approach, as this would require thousands of requests. 
Future work could use strategies like self-consistency to estimate a probabilistic histogram.

\subsection{Which methodology applied to which model?}

The following describes the list of methodologies used for each \llm:
\begin{itemize}
\item Greedy Sampling: \gemini, \llama, \mixtralmoe, \mixtralmoelg. 
We opt for using greedy sampling (\texttt{max\_tokens=200}, \texttt{temperature=0}) as opposed to standard sampling due to budget constraints. Given the nature of our experiments, faithfully estimating the empirical distributions over the 101 numbers would require hundreds or even thousands of calls. These calls are time-consuming and costly. We believe that using greedy decoding is still representative of how a model would behave in the majority of the cases.
\item Full next-token probability distribution: \llamasmall, \gemma, \olmo. We found these models to be particularly brittle to prompting. 
\item Next-token probability distribution: \chatgpt, \gptf, \gptfo. As of June 2024, OpenAI models only provide access to the next-token probabilities of the top-20 tokens. In the experiments in Section \ref{ssec:ablation:fractional-analysis}, we collect the information about the top 20 numbers.
\end{itemize}

\section{Additional Results}
\label{app:additional-results}

In this section, we report the following additional results:

\begin{itemize}
\item In Section \ref{app:ssec:add-results:complete-metrics} we report the mean and 95\% confidence estimates for the proportional agreement (PA) and mean average error (MAE) metrics. Since the PA prioritizes mode-matching behavior, we also include a distribution-matching metric based on Wasserstein distance.
\item In Section \ref{app:ssec:pairwise-correction-analysis} we analyze differences in model responses between true and false statements on a pairwise basis, pairing statements by the original question they were based on.
\item In Section \ref{app:ssec:add-results:interquartile-per-expr} we comment on the variability of the distributions obtained when using probabilistic decoding vs simple greedy decoding.
\item In Section \ref{app:ssec:add-results:histograms} we provide visualizations of the histogram distributions for the non-verifiable and verifiable settings.
\end{itemize}

\subsection{Mode- and Distribution-Matching Metrics}
\label{app:ssec:add-results:complete-metrics}

Tables \ref{tab:app:agreement-nv:with-confidence-intervals} and \ref{tab:app:agreement-v:with-confidence-intervals} show the mean and 95\% confidence intervals for the PA and MAE metrics. 
For completeness, we also include the PA score and MAE metrics per expression in Sections \ref{app:sssec:add-results:pa-distance-per-expr} and \ref{app:sssec:add-results:mae-distance-per-expr}.

\begin{table*}
\centering
\caption{\textbf{Human-LLM agreement for non-verifiable statements}. Average Proportional Agreement (PA), PA as a fraction of the \textit{Human Mode} results (\% PA), and absolute error between mean responses (MAE). 
\emph{Human Mode} represents the mode of the human NV distribution, whereas \emph{Human Individual} represents the PA score of individual human responses relative to the population. 95\% confidence intervals were computed using the adjusted bootstrap percentile method.
}
\label{tab:app:agreement-nv:with-confidence-intervals}
\begin{tabular}{l|cc|c}
\toprule
                & \textbf{PA} \textit{(95\% CI)} & \textbf{\% PA} & \textbf{MAE} \textit{(95\% CI)}\\
\midrule
Human Mode      & 27.6 & --- & ---  \\
Human Individual& $17.6_{(17.2, 18.0)}$ & 63.8 & $8.91_{(8.08, 9.17)}$ \\
\hdashline
\addlinespace
\chatgpt        & $19.7_{(19.1, 20.1)}$     & 71.4      & $6.80_{(6.18, 7.33)}$ \\
\gptf           & $24.4_{(24.2, 24.6)}$     & 88.4      & $4.64_{(4.53, 4.74)}$\\
\gptfo          & $18.9_{(18.4, 19.5)}$     & 68.5      & $5.58_{(5.34, 5.79)}$\\
\gemini         &  $25.4_{(25.1, 25.6)}$    &  {92.0}   & $4.09_{(3.92, 4.23)}$\\
\llamasmall     &  $17.8_{(17.4, 18.2)}$    &  {64.5}   & ${11.99}_{(10.92, 13.13)}$\\
\llama          &  $23.6_{(23.4, 23.8)}$    &  {85.5}   & ${5.56}_{(5.34, 5.80)}$\\
\mixtralmoe     &  $21.8_{(21.4, 22.0)}$    &  {79.0}   & ${5.88}_{(5.44, 6.34)}$\\
\mixtralmoelg   &  $21.8_{(21.6, 22.0)}$    &  {79.0}   & ${7.20}_{(6.56, 7.81)}$\\
\olmo           &  $11.1_{(10.7, 11.5)}$    &  {40.2}   & ${21.41}_{(20.17, 22.41)}$\\
\gemma          &  $8.1_{(7.7, 8.5)}$       &  {29.3}   & ${20.17}_{(19.11, 21.17)}$\\
\bottomrule
\end{tabular}
\end{table*}

\begin{table*}[tb]
\centering
\caption{\textbf{Human-LLM agreement for verifiable statements}. Average Proportional Agreement (PA), absolute error between mean responses (MAE), and the difference between these scores and those from the non-verifiable statements (Table \ref{tab:agreement-nv}) ($\Delta$ PA and $\Delta$ MAE, respectively). Again \emph{Human Mode} represents the mode of the human NV distribution, whereas \emph{Human Individual} represents the average behavior across individual humans on the verifiable setting.
}
\label{tab:app:agreement-v:with-confidence-intervals}
\begin{tabular}{l|cc|cc}
\toprule
& \textbf{PA} \textit{(95\% CI)} & \textbf{$\Delta$ PA} & \textbf{MAE} \textit{(95\% CI)} & \textbf{$\Delta$ MAE} \\
\midrule
Human Mode &  {27.6} & --- & --- & ---\\
Human Individual & $16.7_{(16.3, 17.1)}$ & -0.9 & $9.35_{(8.23, 9.50)}$ & 0.44\\
\hdashline
\addlinespace
\chatgpt        & ${15.3}_{(14.6, 15.9)}$ & -4.4        & ${8.57}_{(6.81, 10.07)}$ & 1.77\\
\gptf           & ${22.1}_{(21.7, 22.4)}$ & -2.3        & ${3.84}_{(3.03, 4.45)}$ & -0.80 \\
\gptfo          & ${15.2}_{(14.5, 15.9)}$ & -3.7        & ${7.05}_{(6.45, 7.62)}$ & 1.47\\
\gemini         & ${21.3}_{(20.7, 21.8)}$ & -4.1        & ${7.23}_{(6.04, 8.49)}$ & 3.14 \\
\llamasmall     & ${10.1}_{(9.5, 10.7)}$ & -7.7         & ${16.59}_{(15.10, 18.18)}$ & 4.60 \\
\llama          & ${18.9}_{(18.1, 19.6)}$ & {-4.7}      & ${13.73}_{(11.92, 15.64)}$ & 8.17 \\
\mixtralmoe     & ${15.2}_{(14.5, 15.9)}$  & -6.6       & ${12.23}_{(10.37, 14.18)}$ & 6.35\\
\mixtralmoelg   & ${18.6}_{(18.3, 19.0)}$  & -3.2       & ${9.78}_{(8.51, 11.08)}$ & 2.58\\
\olmo           & ${7.6}_{(7.2, 8.0)}$  & -3.5          & ${33.66}_{(31.82, 35.11)}$ & 12.25\\
\gemma          & ${5.3}_{(5.0, 5.6)}$  & -2.8          & ${25.07}_{(23.45, 26.66)}$ & 4.9\\
\bottomrule
\end{tabular}
\end{table*}

\begin{table*}[tb]
\centering
\caption{\textbf{Summary metrics averaged across uncertainty expressions for both NV and V settings}. All metrics are computed with respect to the human distribution in the non-verifiable setting (\texttt{Human+NV}). ``PA'' reports the general agreement between LLMs and the mode of the human  distribution, reported in percentages. ``MAE'' reports the absolute error between the mean responses of LLMs and those of humans. Wasserstein-1 computes the distance between LLMs and human distributions.}
\label{tab:summary-metrics}
\begin{tabular}{llcccccc}
\toprule
&  & \multicolumn{2}{c}{Avg  PA ($\uparrow$)} & \multicolumn{2}{c}{Avg MAE}($\downarrow$) & \multicolumn{2}{c}{Avg Wasserstein-1 ($\downarrow$)} \\
\cmidrule(lr){3-4}
\cmidrule(lr){5-6}
\cmidrule(lr){7-8}
&  & \multicolumn{1}{c}{NV} & \multicolumn{1}{c}{V} & \multicolumn{1}{c}{NV} & \multicolumn{1}{c}{V} & \multicolumn{1}{c}{NV} & \multicolumn{1}{c}{V} \\
\midrule
\multirow[t]{2}{*}{Human} & Mode & \cellcolor[HTML]{c0e6b9}{27.6} & \cellcolor[HTML]{c0e6b9}{27.6} & --- & --- & --- & --- \\
 & Individual & \cellcolor[HTML]{daf0d4}{17.6} & \cellcolor[HTML]{dbf1d6}{16.7} & \cellcolor[HTML]{fdd1be}{8.91} & \cellcolor[HTML]{fdcebb}{9.35} & \cellcolor[HTML]{fcbca2}{12.35} & \cellcolor[HTML]{fcb89e}{12.99} \\
\hdashline
Baseline & Random & \cellcolor[HTML]{f0f9ed}{5.1} & \cellcolor[HTML]{f0f9ed}{5.1} & \cellcolor[HTML]{f6563d}{27.72} & \cellcolor[HTML]{f6563d}{27.72} & \cellcolor[HTML]{f5523a}{28.16} & \cellcolor[HTML]{f5523a}{28.16} \\
\hdashline
\multirow[t]{11}{*}{LLM} & OLMo & \cellcolor[HTML]{e6f5e1}{12.1} & \cellcolor[HTML]{ecf8e8}{7.6} & \cellcolor[HTML]{fc9474}{18.44} & \cellcolor[HTML]{e12d26}{33.67} & \cellcolor[HTML]{fc8767}{20.45} & \cellcolor[HTML]{bb141a}{40.16} \\
 & Gemma (2B) & \cellcolor[HTML]{ecf8e8}{8.1} & \cellcolor[HTML]{eef8ea}{6.6} & \cellcolor[HTML]{fc8969}{20.17} & \cellcolor[HTML]{fb6e4e}{24.33} & \cellcolor[HTML]{fb7c5c}{22.13} & \cellcolor[HTML]{f96346}{25.89} \\
 & Llama3 8B & \cellcolor[HTML]{d9f0d3}{17.8} & \cellcolor[HTML]{e9f7e5}{10.1} & \cellcolor[HTML]{fcbea5}{11.99} & \cellcolor[HTML]{fca183}{16.59} & \cellcolor[HTML]{fcb095}{14.11} & \cellcolor[HTML]{fc9576}{18.35} \\
 & Llama3 (70B) & \cellcolor[HTML]{cbeac4}{23.6} & \cellcolor[HTML]{d6efd0}{18.8} & \cellcolor[HTML]{fee3d6}{5.56} & \cellcolor[HTML]{fcb398}{13.73} & \cellcolor[HTML]{fdcbb6}{9.94} & \cellcolor[HTML]{fca285}{16.39} \\
  & Mixtral 8x7B & \cellcolor[HTML]{cfecc9}{21.8} & \cellcolor[HTML]{dff3da}{15.2} & \cellcolor[HTML]{fee1d4}{5.88} & \cellcolor[HTML]{fcbca2}{12.32} & \cellcolor[HTML]{fdd1be}{8.88} & \cellcolor[HTML]{fca588}{15.93} \\
 & Mixtral 8x22B & \cellcolor[HTML]{cfecc9}{21.8} & \cellcolor[HTML]{d7efd1}{18.6} & \cellcolor[HTML]{fedbcc}{7.20} & \cellcolor[HTML]{fdcbb6}{9.78} & \cellcolor[HTML]{fdc5ae}{10.78} & \cellcolor[HTML]{fcbea5}{12.05} \\
 & Gemini & \cellcolor[HTML]{c6e8bf}{25.4} & \cellcolor[HTML]{d0edca}{21.3} & \cellcolor[HTML]{fee8dd}{4.09} & \cellcolor[HTML]{fedaca}{7.23} & \cellcolor[HTML]{fdcebb}{9.24} & \cellcolor[HTML]{fdcbb6}{9.78} \\
 & ChatGPT & \cellcolor[HTML]{d4eece}{19.7} & \cellcolor[HTML]{def2d9}{15.3} & \cellcolor[HTML]{fedecf}{6.80} & \cellcolor[HTML]{fdd3c1}{8.57} & \cellcolor[HTML]{fdcebb}{9.26} & \cellcolor[HTML]{fcb99f}{12.74} \\
 & GPT-4 & \cellcolor[HTML]{c9eac2}{24.4} & \cellcolor[HTML]{ceecc8}{22.1} & \cellcolor[HTML]{fee6da}{4.64} & \cellcolor[HTML]{fee8de}{3.84} & \cellcolor[HTML]{fdcab5}{9.96} & \cellcolor[HTML]{fedccd}{6.88} \\
 & GPT-4o & \cellcolor[HTML]{d6efd0}{18.9} & \cellcolor[HTML]{dff3da}{15.2} & \cellcolor[HTML]{fee3d6}{5.58} & \cellcolor[HTML]{fedbcc}{7.05} & \cellcolor[HTML]{fdc9b3}{10.34} & \cellcolor[HTML]{fdcbb6}{9.96} \\
\bottomrule
\end{tabular}
\end{table*}

\subsubsection{Proportional Agreement}
\label{app:sssec:add-results:pa-distance-per-expr}

As described in Section \ref{ssec:metrics}, the proportional agreement (PA) metric gauges the overall agreement between an agent's and a reference (population) distribution. 
We use the results of the human studies in the non-verifiable setting as our reference distribution (dubbed \texttt{Human Mode}) throughout the whole paper.

Tables \ref{tab:tab:app:pa-breakdown-expr:nv-2shot} and \ref{tab:tab:app:pa-breakdown-expr:v-2shot} report the proportional agreement (PA) metric discriminated by uncertainty expression in the non-verifiable and verifiable settings, respectively. The values are reported with respect to the population-level human distribution described in Section \ref{sec:initalexp}, denoted \texttt{Human+NV}. 

\begin{table*}
\tiny
\centering
\caption{\textbf{Proportional Agreement (PA) score per uncertainty expression in the non-verifiable setting}. The scores are with respect to the population-level human reference distribution (\texttt{Human}+NV).}
\label{tab:tab:app:pa-breakdown-expr:nv-2shot}
\begin{tabular}{lcccccccccc}
\toprule
 & \olmo  & \gemma & \llama  & \llamasmall & \chatgpt & \gptf & \gptfo & \mixtralmoelg & \mixtralmoe \\
Methodology & full & full & full & full & top-k & top-k & top-k & sampling & sampling \\
\midrule
{Average} & {12.2} & {8.6} & {25.1} & {18.8} &{20.5} & {25.0} & {19.1} & {23.9} & {22.3} & {22.0} \\
{Standard Deviation} & {13.6} & {8.8} & {12.8} & {14.2} & {12.5} & {13.1} & {8.8} & {13.5} & {13.0} & {12.5} \\
\hdashline
almost certain & \cellcolor[HTML]{60ba6c}{55.0} & \cellcolor[HTML]{f6fcf4}{0.8} & \cellcolor[HTML]{4bb062}{60.0} & \cellcolor[HTML]{48ae60}{60.6} & \cellcolor[HTML]{5bb86a}{55.9} & \cellcolor[HTML]{48ae60}{60.6} & \cellcolor[HTML]{a8dca2}{35.5} & \cellcolor[HTML]{50b264}{58.7} & \cellcolor[HTML]{48ae60}{60.6} & \cellcolor[HTML]{4db163}{59.7} \\
doubtful & \cellcolor[HTML]{f4fbf1}{2.6} & \cellcolor[HTML]{f0f9ec}{5.5} & \cellcolor[HTML]{d6efd0}{18.9} & \cellcolor[HTML]{e9f7e5}{9.8} & \cellcolor[HTML]{e4f5df}{13.2} & \cellcolor[HTML]{e5f5e0}{12.9} & \cellcolor[HTML]{e1f3dc}{14.2} & \cellcolor[HTML]{daf0d4}{17.5} & \cellcolor[HTML]{dbf1d6}{16.4} & \cellcolor[HTML]{e2f4dd}{13.7} \\
highly likely & \cellcolor[HTML]{d2edcc}{20.5} & \cellcolor[HTML]{f6fcf4}{1.0} & \cellcolor[HTML]{d8f0d2}{18.0} & \cellcolor[HTML]{dff3da}{14.9} & \cellcolor[HTML]{dcf2d7}{16.1} & \cellcolor[HTML]{ceecc8}{22.1} & \cellcolor[HTML]{daf0d4}{17.4} & \cellcolor[HTML]{ebf7e7}{8.3} & \cellcolor[HTML]{caeac3}{23.9} & \cellcolor[HTML]{d0edca}{21.5} \\
highly unlikely & \cellcolor[HTML]{dff3da}{14.9} & \cellcolor[HTML]{e8f6e4}{10.5} & \cellcolor[HTML]{a9dca3}{35.0} & \cellcolor[HTML]{d4eece}{19.8} & \cellcolor[HTML]{c6e8bf}{25.7} & \cellcolor[HTML]{a9dca3}{35.1} & \cellcolor[HTML]{c8e9c1}{24.8} & \cellcolor[HTML]{a9dca3}{35.1} & \cellcolor[HTML]{bce4b5}{28.7} & \cellcolor[HTML]{c7e9c0}{25.2} \\
not likely & \cellcolor[HTML]{f1faee}{4.2} & \cellcolor[HTML]{f1faee}{4.6} & \cellcolor[HTML]{d9f0d3}{17.8} & \cellcolor[HTML]{def2d9}{15.5} & \cellcolor[HTML]{dbf1d6}{16.7} & \cellcolor[HTML]{dbf1d5}{17.1} & \cellcolor[HTML]{e7f6e3}{11.3} & \cellcolor[HTML]{d9f0d3}{17.7} & \cellcolor[HTML]{dcf2d7}{16.1} & \cellcolor[HTML]{e8f6e4}{10.3} \\
possible & \cellcolor[HTML]{edf8ea}{6.7} & \cellcolor[HTML]{e1f3dc}{14.3} & \cellcolor[HTML]{def2d9}{15.3} & \cellcolor[HTML]{ecf8e8}{7.7} & \cellcolor[HTML]{e1f3dc}{14.2} & \cellcolor[HTML]{def2d9}{15.4} & \cellcolor[HTML]{eef8ea}{6.5} & \cellcolor[HTML]{dff3da}{15.0} & \cellcolor[HTML]{ebf7e7}{8.6} & \cellcolor[HTML]{e0f3db}{14.8} \\
probable & \cellcolor[HTML]{ecf8e8}{7.9} & \cellcolor[HTML]{f2faf0}{3.3} & \cellcolor[HTML]{dff3da}{15.2} & \cellcolor[HTML]{edf8ea}{7.0} & \cellcolor[HTML]{ecf8e8}{7.6} & \cellcolor[HTML]{e1f3dc}{14.3} & \cellcolor[HTML]{dff3da}{14.9} & \cellcolor[HTML]{e0f3db}{14.8} & \cellcolor[HTML]{ddf2d8}{15.7} & \cellcolor[HTML]{e3f4de}{13.5} \\
somewhat likely & \cellcolor[HTML]{e9f7e5}{9.5} & \cellcolor[HTML]{f4fbf1}{2.7} & \cellcolor[HTML]{dbf1d6}{16.5} & \cellcolor[HTML]{f0f9ed}{4.7} & \cellcolor[HTML]{edf8ea}{6.8} & \cellcolor[HTML]{d9f0d3}{17.9} & \cellcolor[HTML]{e5f5e0}{12.8} & \cellcolor[HTML]{dbf1d6}{16.5} & \cellcolor[HTML]{d8f0d2}{18.2} & \cellcolor[HTML]{def2d9}{15.4} \\
somewhat unlikely & \cellcolor[HTML]{f6fcf4}{1.0} & \cellcolor[HTML]{def2d9}{15.5} & \cellcolor[HTML]{cdecc7}{22.3} & \cellcolor[HTML]{cfecc9}{21.5} & \cellcolor[HTML]{cfecc9}{21.8} & \cellcolor[HTML]{d5efcf}{19.3} & \cellcolor[HTML]{d8f0d2}{18.0} & \cellcolor[HTML]{cdecc7}{22.3} & \cellcolor[HTML]{d7efd1}{18.4} & \cellcolor[HTML]{d5efcf}{19.3} \\
uncertain & \cellcolor[HTML]{eff9eb}{6.2} & \cellcolor[HTML]{abdda5}{34.0} & \cellcolor[HTML]{a9dca3}{35.1} & \cellcolor[HTML]{a9dca3}{35.1} & \cellcolor[HTML]{afdfa8}{33.1} & \cellcolor[HTML]{a9dca3}{35.1} & \cellcolor[HTML]{a9dca3}{35.1} & \cellcolor[HTML]{a9dca3}{35.1} & \cellcolor[HTML]{a9dca3}{35.1} & \cellcolor[HTML]{a9dca3}{35.1} \\
unlikely & \cellcolor[HTML]{f5fbf2}{1.6} & \cellcolor[HTML]{eaf7e6}{9.2} & \cellcolor[HTML]{dbf1d6}{16.6} & \cellcolor[HTML]{dbf1d5}{16.8} & \cellcolor[HTML]{ddf2d8}{15.9} & \cellcolor[HTML]{d8f0d2}{18.3} & \cellcolor[HTML]{e7f6e3}{11.1} & \cellcolor[HTML]{dbf1d6}{16.8} & \cellcolor[HTML]{e9f7e5}{10.1} & \cellcolor[HTML]{dcf2d7}{16.4} \\
very likely & \cellcolor[HTML]{e1f3dc}{14.4} & \cellcolor[HTML]{f6fcf4}{0.8} & \cellcolor[HTML]{daf0d4}{17.4} & \cellcolor[HTML]{dff3da}{15.2} & \cellcolor[HTML]{e3f4de}{13.6} & \cellcolor[HTML]{d5efcf}{19.5} & \cellcolor[HTML]{d5efcf}{19.1} & \cellcolor[HTML]{e1f3dc}{14.4} & \cellcolor[HTML]{dff3da}{15.1} & \cellcolor[HTML]{d6efd0}{19.1} \\
very unlikely & \cellcolor[HTML]{e2f4dd}{13.9} & \cellcolor[HTML]{eaf7e6}{9.2} & \cellcolor[HTML]{9ed798}{38.3} & \cellcolor[HTML]{ddf2d8}{15.9} & \cellcolor[HTML]{c3e7bc}{26.5} & \cellcolor[HTML]{a0d99b}{37.9} & \cellcolor[HTML]{bee5b8}{27.8} & \cellcolor[HTML]{9cd797}{38.8} & \cellcolor[HTML]{cdecc7}{22.3} & \cellcolor[HTML]{cfecc9}{21.8} \\
\bottomrule
\end{tabular}
\end{table*}

\begin{table*}
\tiny
\centering

\caption{\textbf{Proportional Agreement (PA) score per uncertainty expression in the verifiable setting}. The PA scores are with respect to the population-level human reference distribution (\texttt{Human}+NV).}
\label{tab:tab:app:pa-breakdown-expr:v-2shot}
\begin{tabular}{lllllllllllll}
\toprule

             &  \olmo &  \gemma &  \llamasmall & \chatgpt & \gptf & \gptfo &  \llama &  \mixtralmoelg &  \mixtralmoe &  \gemini \\
Methodology  & full & full & full & top-k & top-k & top-k & sampling & sampling & sampling & sampling \\
\midrule
Average & {7.8} & {7.0} & {10.3} & {15.9} & {22.8} & {15.3} & {19.5} & {19.4} & {15.6} & {22.1} \\
Standard Deviation & {12.2} & {8.9} & {5.1} & {11.5} & {13.8} & {7.1} & {10.8} & {14.6} & {9.3} & {13.2} \\
\hdashline
almost certain & \cellcolor[HTML]{83cb82}{46.0} & \cellcolor[HTML]{f6fcf4}{0.8} & \cellcolor[HTML]{d2edcc}{20.6} & \cellcolor[HTML]{7cc87c}{48.0} & \cellcolor[HTML]{48ae60}{60.6} & \cellcolor[HTML]{bce4b5}{28.7} & \cellcolor[HTML]{a8dca2}{35.2} & \cellcolor[HTML]{48ae60}{60.6} & \cellcolor[HTML]{a5db9f}{36.2} & \cellcolor[HTML]{62bb6d}{54.6} \\
doubtful & \cellcolor[HTML]{f6fcf4}{0.9} & \cellcolor[HTML]{f1faee}{4.2} & \cellcolor[HTML]{e7f6e3}{11.2} & \cellcolor[HTML]{eff9eb}{6.2} & \cellcolor[HTML]{e7f6e2}{11.5} & \cellcolor[HTML]{e7f6e2}{11.6} & \cellcolor[HTML]{dff3da}{15.2} & \cellcolor[HTML]{e2f4dd}{13.8} & \cellcolor[HTML]{eaf7e6}{9.3} & \cellcolor[HTML]{dbf1d6}{16.4} \\
highly likely & \cellcolor[HTML]{d7efd1}{18.5} & \cellcolor[HTML]{f6fcf4}{0.9} & \cellcolor[HTML]{e7f6e3}{11.1} & \cellcolor[HTML]{e4f5df}{13.0} & \cellcolor[HTML]{ceecc8}{22.0} & \cellcolor[HTML]{e0f3db}{14.7} & \cellcolor[HTML]{daf0d4}{17.4} & \cellcolor[HTML]{caeac3}{24.1} & \cellcolor[HTML]{e5f5e0}{12.8} & \cellcolor[HTML]{ccebc6}{22.8} \\
highly unlikely & \cellcolor[HTML]{ebf7e7}{8.3} & \cellcolor[HTML]{f5fbf3}{1.2} & \cellcolor[HTML]{f1faee}{4.0} & \cellcolor[HTML]{ceecc8}{22.1} & \cellcolor[HTML]{b0dfaa}{32.8} & \cellcolor[HTML]{daf0d4}{17.2} & \cellcolor[HTML]{aedea7}{33.3} & \cellcolor[HTML]{bce4b5}{28.7} & \cellcolor[HTML]{caeac3}{23.9} & \cellcolor[HTML]{aedea7}{33.3} \\
not likely & \cellcolor[HTML]{f4fbf1}{2.6} & \cellcolor[HTML]{f5fbf3}{1.5} & \cellcolor[HTML]{e2f4dd}{14.0} & \cellcolor[HTML]{e5f5e0}{12.8} & \cellcolor[HTML]{dff3da}{15.0} & \cellcolor[HTML]{e8f6e3}{10.9} & \cellcolor[HTML]{def2d9}{15.5} & \cellcolor[HTML]{e3f4de}{13.3} & \cellcolor[HTML]{ecf8e8}{7.5} & \cellcolor[HTML]{ddf2d8}{15.7} \\
possible & \cellcolor[HTML]{f5fbf3}{1.2} & \cellcolor[HTML]{dff3da}{14.8} & \cellcolor[HTML]{f2faf0}{3.5} & \cellcolor[HTML]{eff9ec}{5.8} & \cellcolor[HTML]{e8f6e4}{10.2} & \cellcolor[HTML]{eff9eb}{6.1} & \cellcolor[HTML]{f1faee}{4.0} & \cellcolor[HTML]{f1faee}{4.6} & \cellcolor[HTML]{eff9eb}{6.0} & \cellcolor[HTML]{edf8ea}{6.7} \\
probable & \cellcolor[HTML]{f5fbf2}{1.8} & \cellcolor[HTML]{f2faf0}{3.3} & \cellcolor[HTML]{eef8ea}{6.5} & \cellcolor[HTML]{f2faef}{3.8} & \cellcolor[HTML]{e7f6e2}{11.5} & \cellcolor[HTML]{e7f6e3}{11.3} & \cellcolor[HTML]{edf8ea}{6.8} & \cellcolor[HTML]{eff9ec}{5.8} & \cellcolor[HTML]{eff9ec}{5.7} & \cellcolor[HTML]{e9f7e5}{9.9} \\
somewhat likely & \cellcolor[HTML]{f5fbf2}{1.7} & \cellcolor[HTML]{f1faee}{3.9} & \cellcolor[HTML]{edf8e9}{7.2} & \cellcolor[HTML]{eef8ea}{6.5} & \cellcolor[HTML]{e6f5e1}{12.1} & \cellcolor[HTML]{edf8ea}{7.0} & \cellcolor[HTML]{ebf7e7}{8.6} & \cellcolor[HTML]{ebf7e7}{9.0} & \cellcolor[HTML]{e9f7e5}{9.7} & \cellcolor[HTML]{e9f7e5}{10.0} \\
somewhat unlikely & \cellcolor[HTML]{f6fcf4}{0.8} & \cellcolor[HTML]{dff3da}{15.0} & \cellcolor[HTML]{e8f6e3}{10.6} & \cellcolor[HTML]{def2d9}{15.4} & \cellcolor[HTML]{d6efd0}{18.8} & \cellcolor[HTML]{dff3da}{15.0} & \cellcolor[HTML]{daf0d4}{17.3} & \cellcolor[HTML]{e6f5e1}{11.8} & \cellcolor[HTML]{d4eece}{19.6} & \cellcolor[HTML]{def2d9}{15.4} \\
uncertain & \cellcolor[HTML]{f7fcf5}{0.0} & \cellcolor[HTML]{b1e0ab}{32.4} & \cellcolor[HTML]{e1f3dc}{14.2} & \cellcolor[HTML]{c7e9c0}{25.1} & \cellcolor[HTML]{abdda5}{34.2} & \cellcolor[HTML]{b7e2b1}{30.3} & \cellcolor[HTML]{acdea6}{34.0} & \cellcolor[HTML]{aedea7}{33.4} & \cellcolor[HTML]{b5e1ae}{31.0} & \cellcolor[HTML]{acdea6}{34.0} \\
unlikely & \cellcolor[HTML]{f7fcf5}{0.4} & \cellcolor[HTML]{e7f6e3}{11.3} & \cellcolor[HTML]{d9f0d3}{17.7} & \cellcolor[HTML]{e3f4de}{13.6} & \cellcolor[HTML]{dbf1d5}{16.8} & \cellcolor[HTML]{e8f6e3}{10.6} & \cellcolor[HTML]{dff3da}{15.0} & \cellcolor[HTML]{eaf7e6}{9.4} & \cellcolor[HTML]{e7f6e2}{11.6} & \cellcolor[HTML]{e1f3dc}{14.3} \\
very likely & \cellcolor[HTML]{e7f6e2}{11.6} & \cellcolor[HTML]{f6fcf4}{0.7} & \cellcolor[HTML]{eaf7e6}{9.3} & \cellcolor[HTML]{e9f7e5}{10.1} & \cellcolor[HTML]{daf0d4}{17.2} & \cellcolor[HTML]{daf0d4}{17.2} & \cellcolor[HTML]{dff3da}{14.9} & \cellcolor[HTML]{dff3da}{14.9} & \cellcolor[HTML]{e7f6e3}{11.1} & \cellcolor[HTML]{daf0d4}{17.6} \\
very unlikely & \cellcolor[HTML]{ecf8e8}{7.5} & \cellcolor[HTML]{f5fbf2}{1.7} & \cellcolor[HTML]{f1faee}{4.1} & \cellcolor[HTML]{caeac3}{24.2} & \cellcolor[HTML]{abdda5}{34.1} & \cellcolor[HTML]{d6efd0}{18.8} & \cellcolor[HTML]{a7dba0}{35.9} & \cellcolor[HTML]{cdecc7}{22.3} & \cellcolor[HTML]{d7efd1}{18.6} & \cellcolor[HTML]{a5db9f}{36.2} \\
\bottomrule
\end{tabular}
\end{table*}

\subsubsection{Mean Absolute Error}
\label{app:sssec:add-results:mae-distance-per-expr}

Proposed in Section \ref{ssec:metrics}, Mean Absolute Error (MAE) measures the average agreement across uncertainty expressions between an agent's distribution and a reference (population) distribution. 
In this case, we also use the human results from the non-verifiable setting as our reference distribution throughout the paper.

\subsubsection{Wasserstein Distance}
\label{app:sssec:add-results:wasserstein-distance-per-expr}

We use one-dimensional Wasserstein distance as a measure of the similarity between two conditional distributions.
We use \texttt{scipy.stats.wasserstein\_distance} and provide the 22 bins for each conditional distribution (\ie all 21 bins from $0, 5, ... 100$ but also the -1) and the corresponding normalized counts that were used to estimate the conditional distributions. 
Two identical distributions will be assigned a Wasserstein-1 distance of 0, whereas two maximally distant distributions will be assigned a Wasserstein-1 distance of 101.

Table \ref{tab:summary-metrics} summarizes the distributional differences between the \llm conditional distributions for both \texttt{LLM+NV} and \texttt{LLM+V} settings with respect to the reference distribution (estimated in Section \ref{sec:initalexp} and denoted \texttt{Human+NV}). We observe that overall the obtained results correlate well with the results reported by the MAE metric.

Other uses of distributional differences in the verifiable setting: we compare how different the empirical distributions obtained in the main experiment differ with respect to the empirical distributions observed in the generalization experiment. These results are reported in Section \ref{app:sec:generalization-results}.

\subsubsection{Differences in Mean Numerical Responses}
\label{app:ssec:add-results:mean-abs-err}

\begin{figure*}[tb]
    \centering
    \begin{subfigure}[b]{0.32\textwidth}
        \centering
        \includegraphics[width=\textwidth]{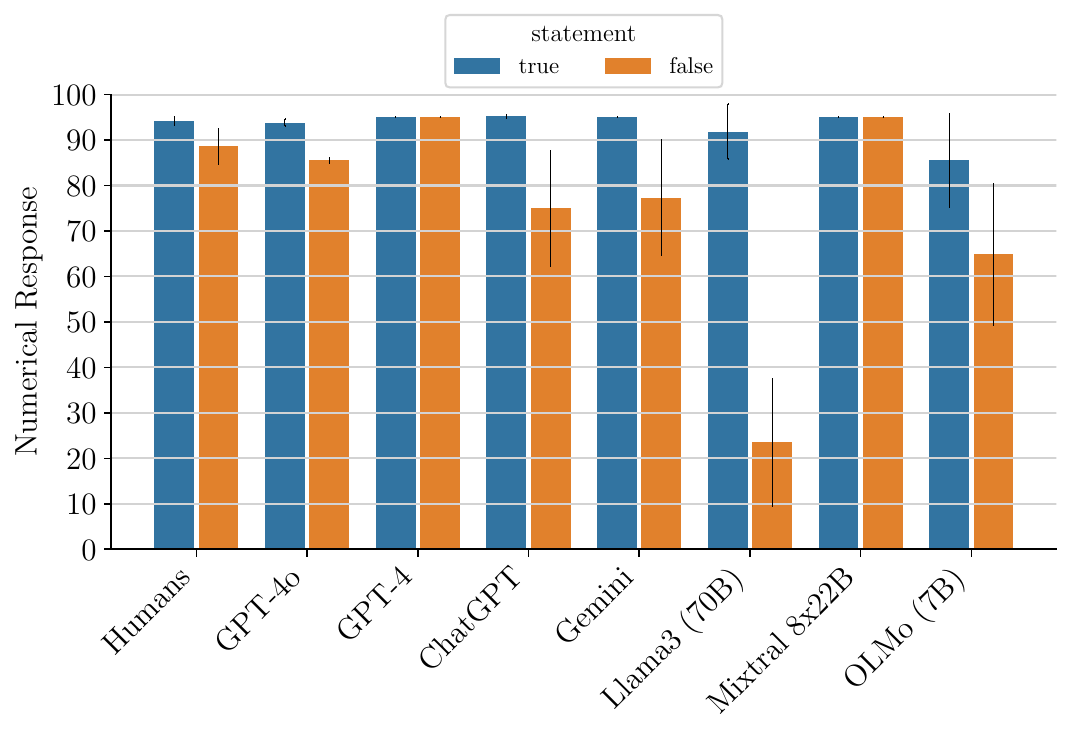}
        \caption{``almost certain''}
        \label{sfig:mean-rated-prob:almost-certain}
    \end{subfigure}
    \hfill
    \begin{subfigure}[b]{0.32\textwidth}
        \centering
        \includegraphics[width=\textwidth]{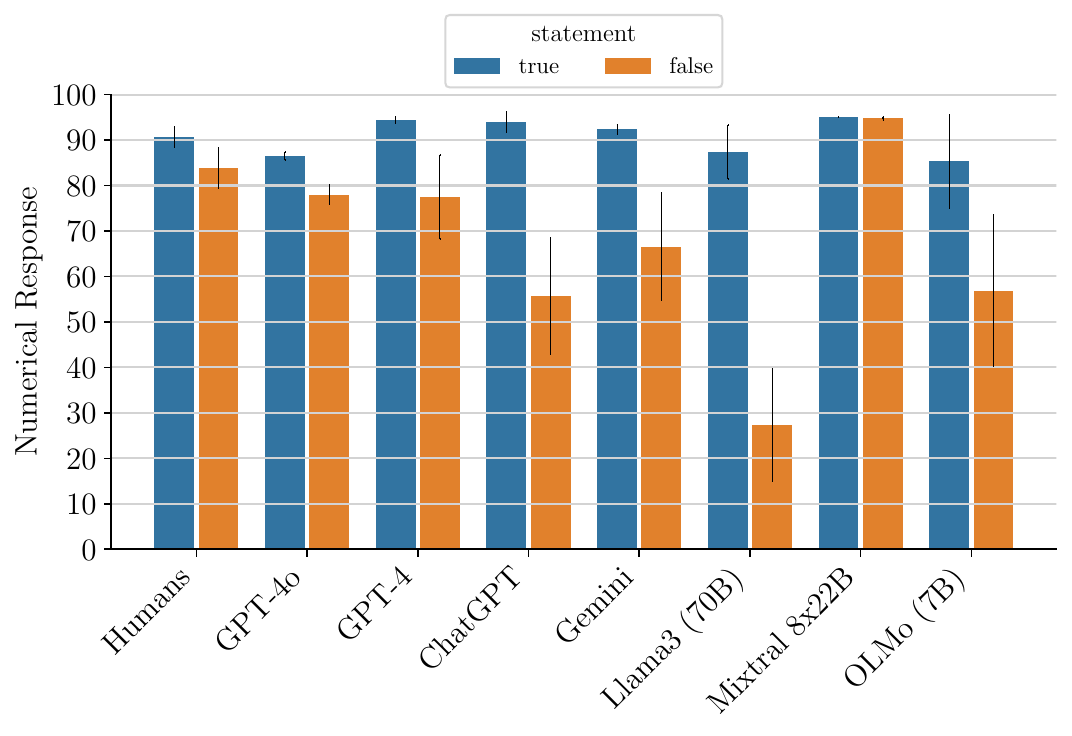}
        \caption{\texttt{``highly likely''}}
        \label{sfig:mean-rated-prob:highly-likely}
    \end{subfigure}
    \hfill
    \begin{subfigure}[b]{0.32\textwidth}
        \centering
        \includegraphics[width=\textwidth]{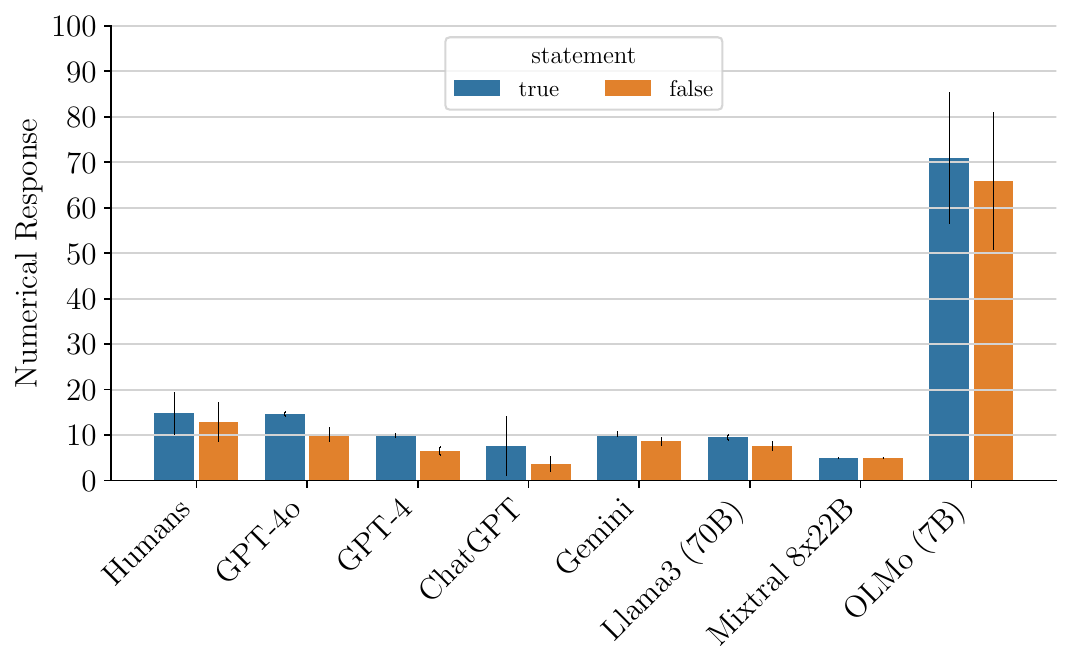}
        \caption{``highly unlikely''}
        \label{sfig:mean-rated-prob:highly_unlikely}
    \end{subfigure}
    \begin{subfigure}[b]{0.32\textwidth}
        \centering
        \includegraphics[width=\textwidth]{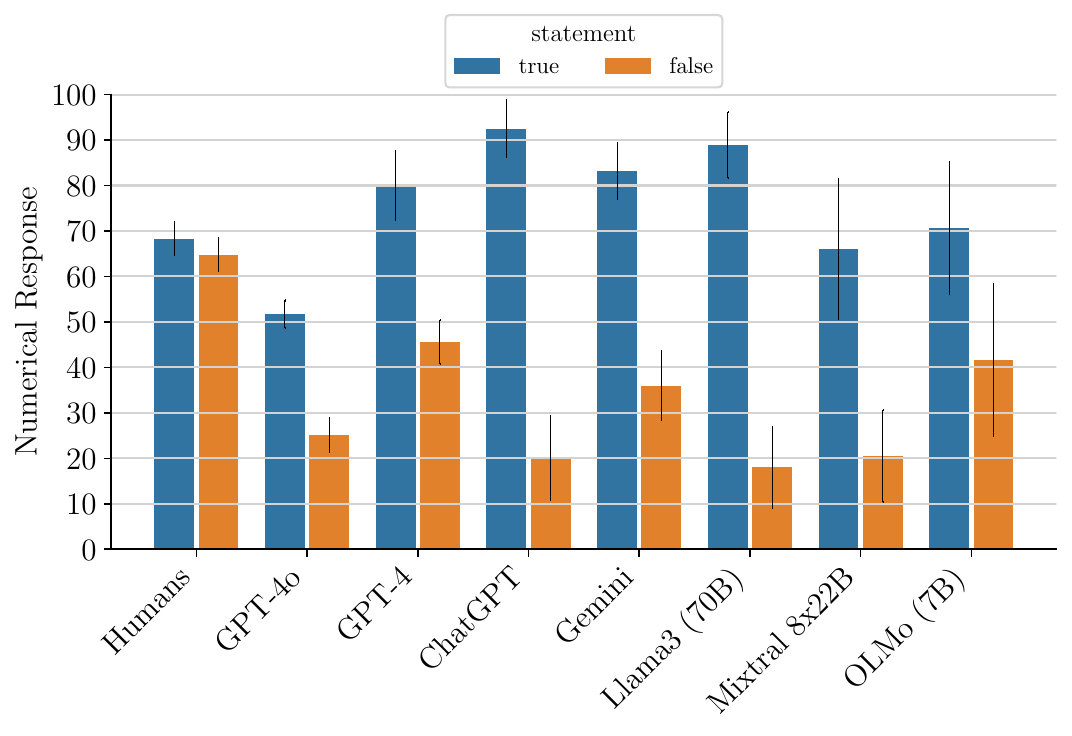}
        \caption{``possible''}
        \label{sfig:mean-rated-prob:possible}
    \end{subfigure}
    \hfill
    \begin{subfigure}[b]{0.32\textwidth}
        \centering
        \includegraphics[width=\textwidth]{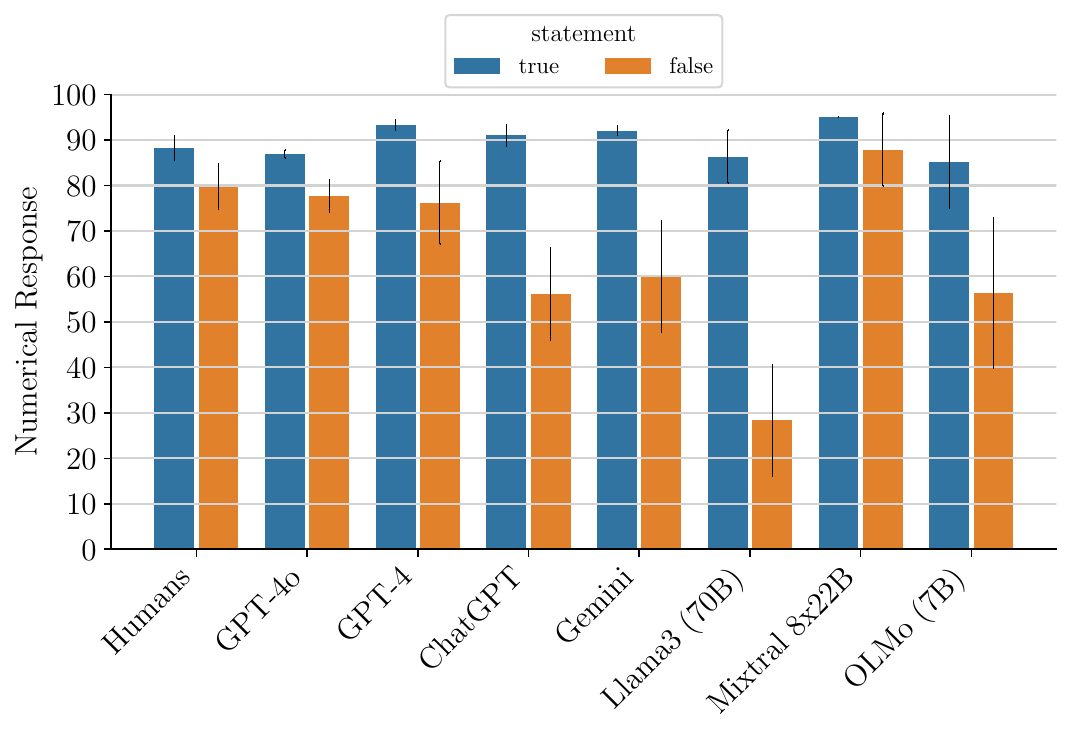}
        \caption{``very likely''}
        \label{sfig:mean-rated-prob:very-likely}
    \end{subfigure}
    \hfill
    \begin{subfigure}[b]{0.32\textwidth}
        \centering
        \includegraphics[width=\textwidth]{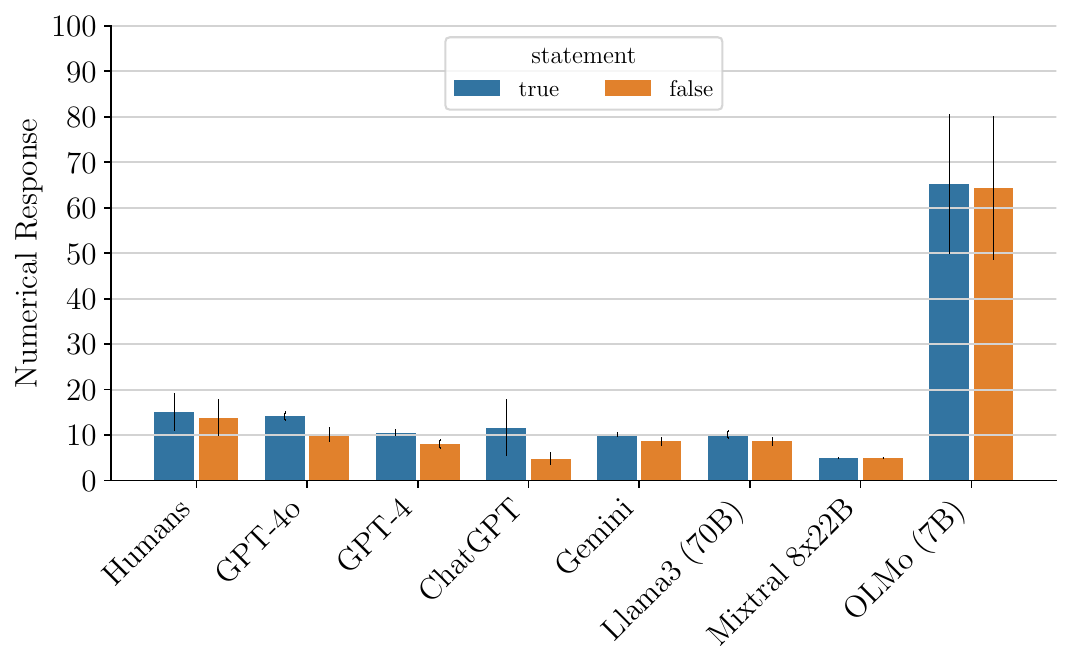}
        \caption{``very unlikely''}
        \label{sfig:mean-rated-prob:very-unlikely}
    \end{subfigure}
    \begin{subfigure}[b]{0.32\textwidth}
        \centering
        \includegraphics[width=\textwidth]{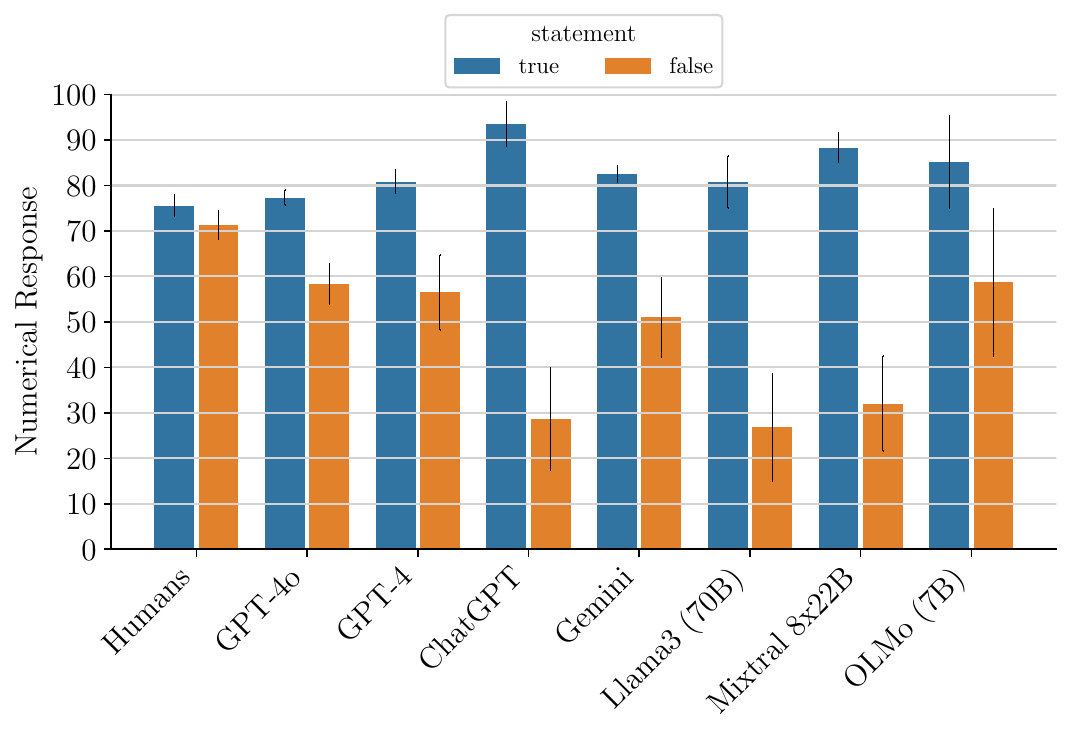}
        \caption{``probable''}
        \label{sfig:mean-rated-prob:probable}
    \end{subfigure}
    \begin{subfigure}[b]{0.32\textwidth}
        \centering
        \includegraphics[width=\textwidth]{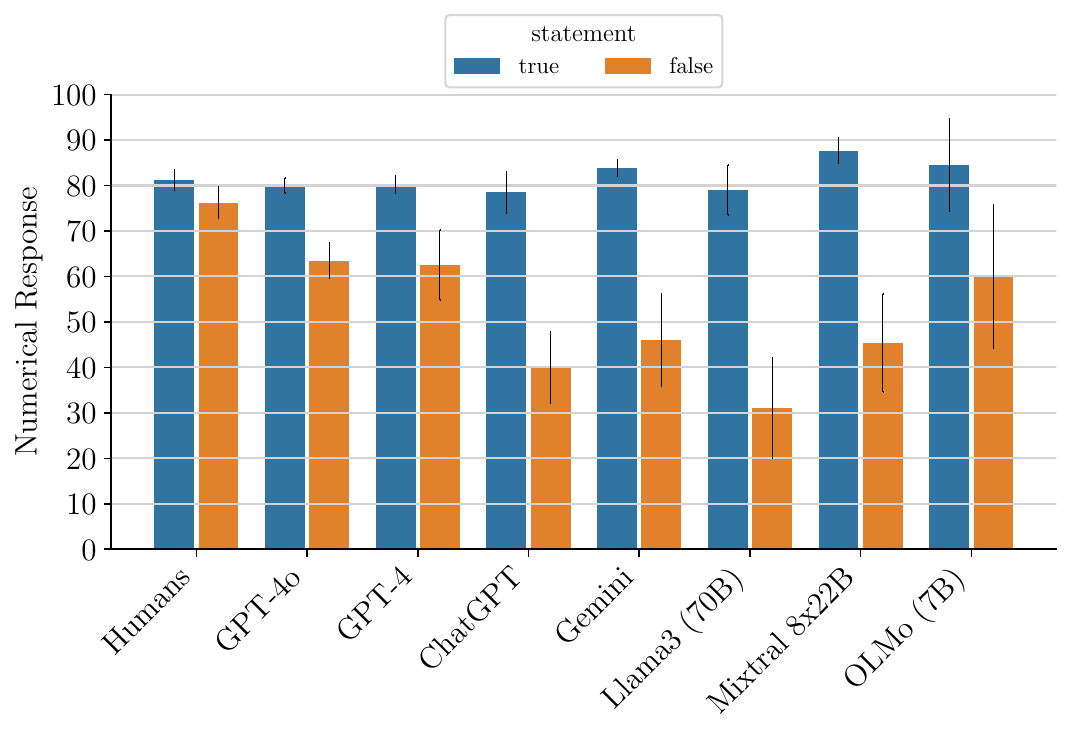}
        \caption{``likely''}
        \label{sfig:mean-rated-prob:likely}
    \end{subfigure}
    \hfill
    \begin{subfigure}[b]{0.32\textwidth}
        \centering
        \includegraphics[width=\textwidth]{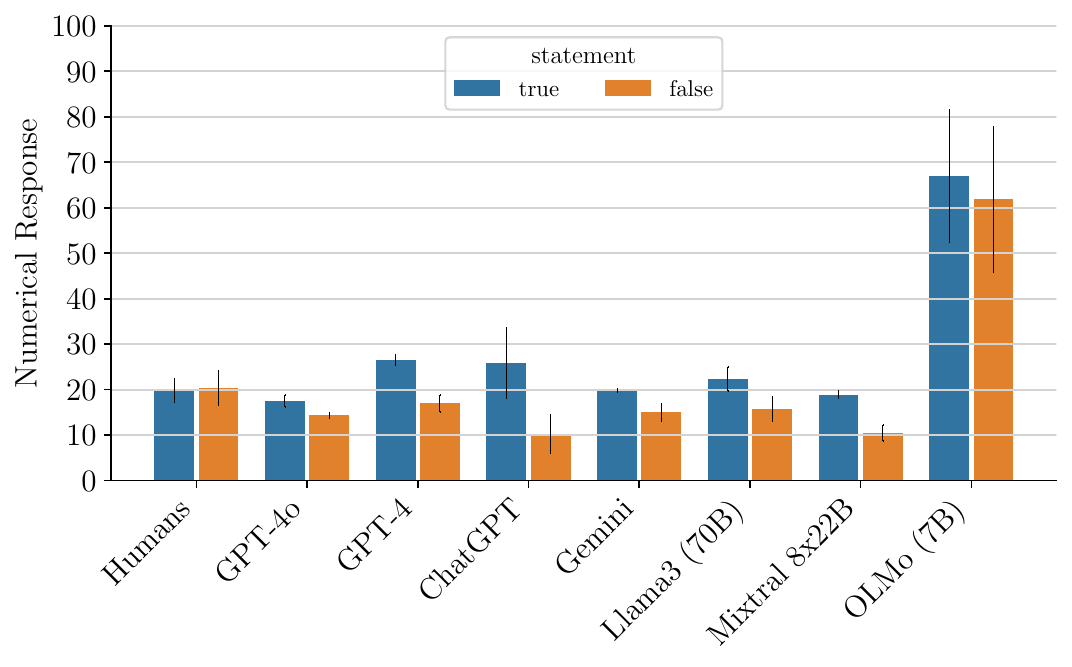}
        \caption{``not likely''}
        \label{sfig:mean-rated-prob:not-likely}
    \end{subfigure}
    \hfill
    \begin{subfigure}[b]{0.32\textwidth}
        \centering
        \includegraphics[width=\textwidth]{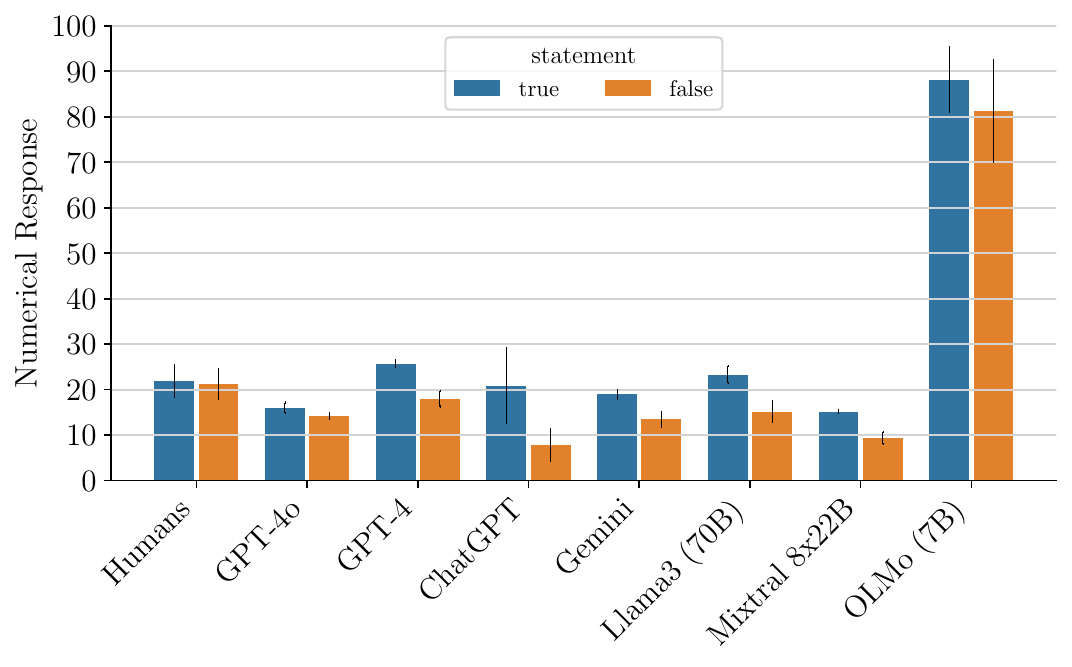}
        \caption{``unlikely''}
        \label{sfig:mean-rated-prob:unlikely}
    \end{subfigure}
    \begin{subfigure}[b]{0.32\textwidth}
        \centering
        \includegraphics[width=\textwidth]{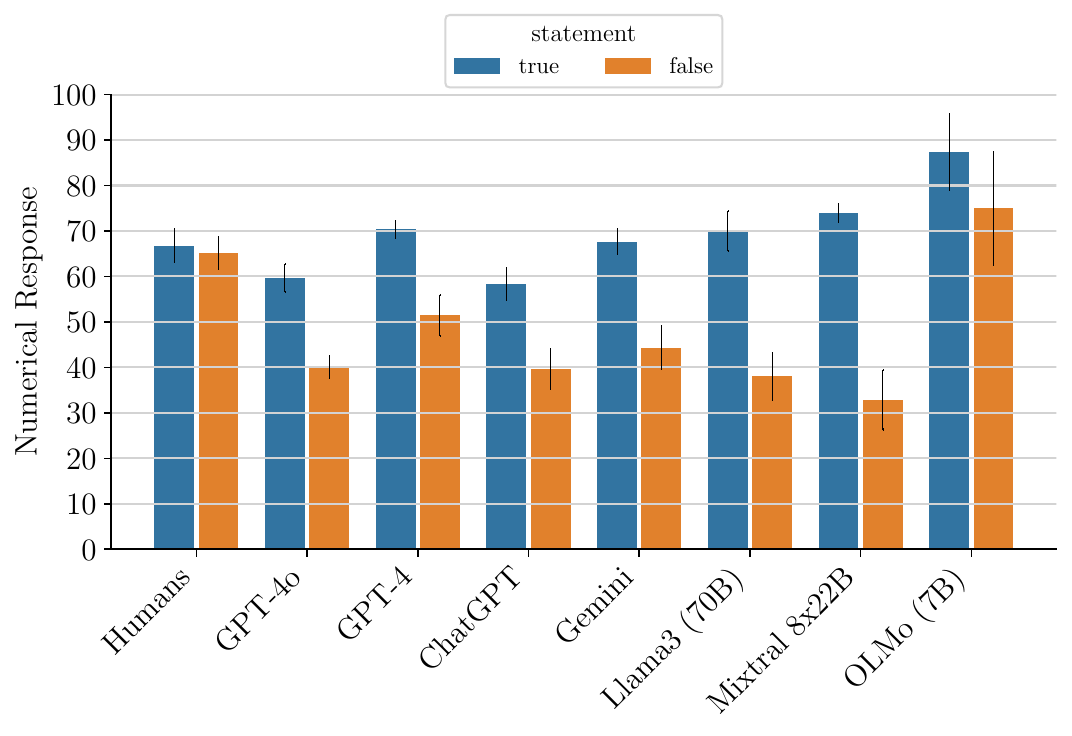}
        \caption{``somewhat likely''}
        \label{sfig:mean-rated-prob:somewhat-likely}
    \end{subfigure}
    \hfill
    \begin{subfigure}[b]{0.32\textwidth}
        \centering
        \includegraphics[width=\textwidth]{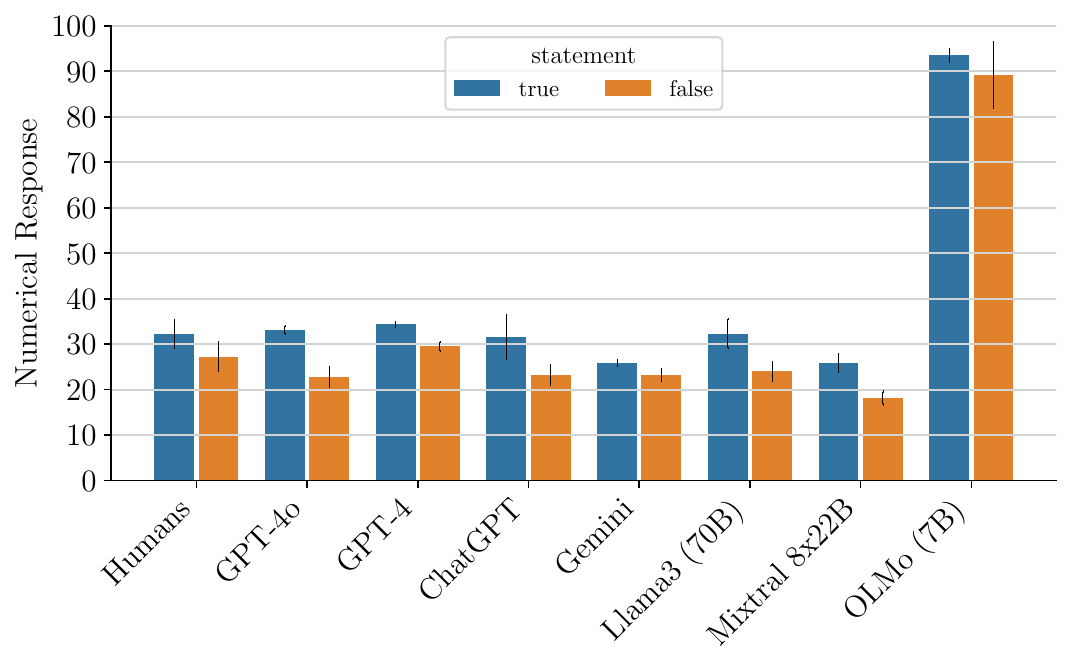}
        \caption{``somewhat unlikely''}
        \label{sfig:mean-rated-prob:somewhat-unlikely}
    \end{subfigure}
    \begin{subfigure}[b]{0.32\textwidth}
        \centering
        \includegraphics[width=\textwidth]{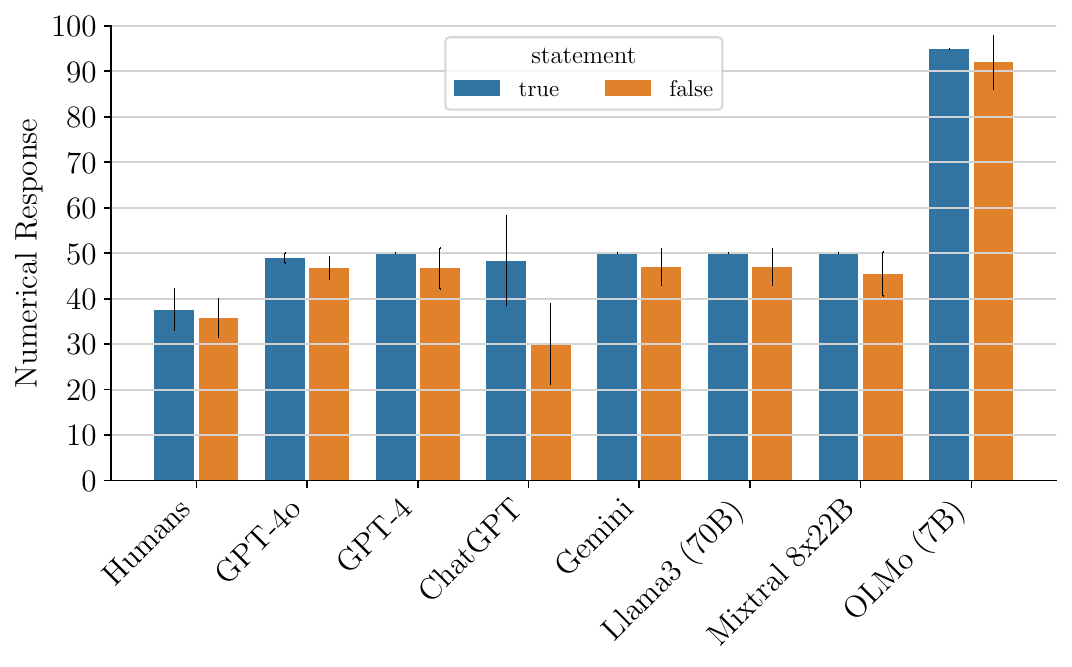}
        \caption{``uncertain''}
        \label{sfig:mean-rated-prob:uncertain}
    \end{subfigure}
    \begin{subfigure}[b]{0.32\textwidth}
        \centering
        \includegraphics[width=\textwidth]{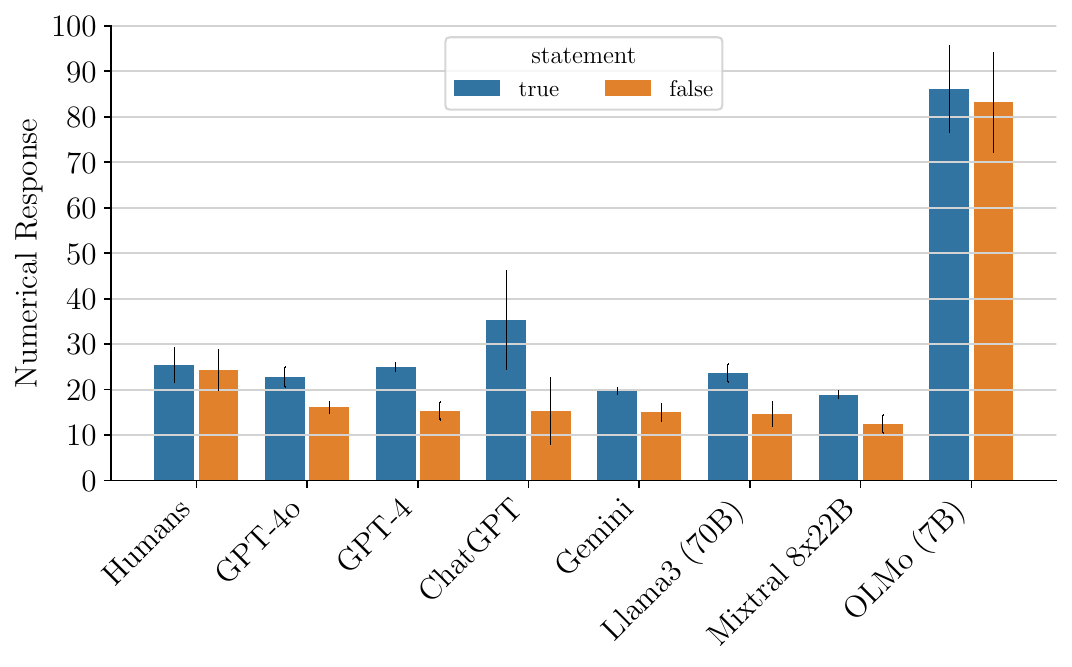}
        \caption{``doubtful''}
        \label{sfig:mean-rated-prob:doubtful}
    \end{subfigure}
    \caption{\textbf{Mean numerical response for the verifiable statements discriminated by truthfulness of the statements}. We observe that the differences in mean numerical responses differ by uncertainty expressions. Focusing on the \textit{opposing} linguistic expressions (\eg ``unlikely'' \textit{vs} ``likely''), the mean numerical response gap observed for models seems to be systematically larger for positive expressions (\eg ``likely'') than for negative expressions (\eg ``unlikely'').}
    \label{fig:mean-rated-prob:all-exprs-verifiable-experiments}
\end{figure*}

Figures \ref{fig:mean-rated-prob:all-exprs-verifiable-experiments} and \ref{fig:mean-rated-prob-generalization} show the mean numerical response for the two verifiable datasets, discriminated by model and uncertainty expression. 
Overall, both plots show evidence of the large perceptual differences exhibited by different models according to the truthfulness of the evaluated statements. 

\subsection{Pairwise Correctness Analysis}
\label{app:ssec:pairwise-correction-analysis}

In our paper, we find that \llms exhibit a systematic ``knowledge bias,'' where the average numerical response is higher for true statements than for false statements. 
To provide statistical support to this claim, we run a one-sided paired Wilcoxon signed-rank test, comparing mean response between true and false statements. 
We perform a paired test, as a true and false statement were generated from each original question (e.g., ``water's chemical formula is H2O'' and ``carbon monoxide's chemical formula is H2O'' are paired). 
Overall, we find that the knowledge bias is statistically significant at $\alpha=0.05$ for all models except \olmo and \gemma---see Tables \ref{tab:app:paired-wilcoxon-greater-test:median-values}, \ref{tab:app:ai2arc-easy:paired-wilcoxon-greater-test:median-values}, and \ref{tab:app:ai2arc-chall:paired-wilcoxon-greater-test:median-values}.

\begin{table}
\small
\centering
\caption{\textbf{Analysis of the pairwise differences between true and false statements in the verifiable dataset}. Using a one-sided Wilcoxon signed-rank test, we find that, for most models, numerical responses for true statements are statistically significantly larger than those for false statements (* = significant at $\alpha=0.05$).}
\label{tab:app:paired-wilcoxon-greater-test:median-values}
\begin{tabular}{llrc}
\toprule
Model & Methodology & Statistic & p-value \\
\midrule
\chatgpt & top-k & 31180.00 & \textbf{<0.0001*} \\
\gptf & top-k & 34980.00 & \textbf{<0.0001*} \\
\gptfo & top-k & 36185.50 & \textbf{<0.0001*} \\
\addlinespace
\gemini & sampling & 25564.00 & \textbf{<0.0001*} \\
\llama & sampling & 34979.00 & \textbf{<0.0001*} \\
\mixtralmoelg & sampling & 16051.00 & \textbf{<0.0001*} \\
\olmo & sampling & 5309.50 & \textbf{<0.0001*} \\
\addlinespace
\llamasmall & full & 17115.00 & \textbf{<0.0001*} \\
\olmo & full & 5309.50 & \textbf{<0.0001*} \\
\gemma & full & 786.50 & 0.9979 \\
\bottomrule
\end{tabular}
\end{table}

\begin{table}
\small
\centering
\caption{\textbf{Analysis of the pairwise differences between true and false statements in the AI2-ARC (Easy) dataset}. Using a one-sided Wilcoxon signed-rank test, we find that, for most models, numerical responses for true statements are statistically significantly larger than those for false statements (* = significant at $\alpha=0.05$).}
\label{tab:app:ai2arc-easy:paired-wilcoxon-greater-test:median-values}
\begin{tabular}{llrc}
\toprule
Model & Methodology & Statistic & p-value \\
\midrule
\midrule
\chatgpt & top-k & 183290.50 & \textbf{<0.0001*} \\
\gptf & top-k & 303749.50 & \textbf{<0.0001*} \\
\gptfo & top-k & 346533.00 & \textbf{<0.0001*} \\
\addlinespace
\gemini & sampling & 180349.50 & \textbf{<0.0001*} \\
\llama & sampling & 250250.00 & \textbf{<0.0001*} \\
\mixtralmoelg & sampling & 129417.50 & \textbf{<0.0001*} \\
\olmo & sampling & 111281.00 & \textbf{<0.0001*} \\
\gemma & sampling & 46702.50 & 0.1148 \\
\bottomrule
\end{tabular}
\end{table}

\begin{table}
\small
\centering
\caption{\textbf{Analysis of the pairwise differences between true and false statements in the AI2-ARC  (Challenge) dataset}. Using a one-sided Wilcoxon signed-rank test, we find that, for most models, numerical responses for true statements are statistically significantly larger than those for false statements (* = significant at $\alpha=0.05$).}
\label{tab:app:ai2arc-chall:paired-wilcoxon-greater-test:median-values}
\begin{tabular}{llrc}
\toprule
\chatgpt & top-k & 123271.00 & \textbf{<0.0001*} \\
\gptf & top-k & 194138.00 & \textbf{<0.0001*} \\
\gptfo & top-k & 231543.50 & \textbf{<0.0001*} \\
\addlinespace
\gemini & sampling & 106837.00 & \textbf{<0.0001*} \\
\llama & sampling & 135453.00 & \textbf{<0.0001*} \\
\mixtralmoelg & sampling & 99689.50 & \textbf{<0.0001*} \\
\olmo & sampling & 48763.50 & 0.0333 \\
\gemma & sampling & 34161.50 & 0.0191 \\
\bottomrule
\end{tabular}
\end{table}

\subsection{Variability}
\label{app:ssec:add-results:interquartile-per-expr}

In this section, we conduct a quantitative analysis of the variability of the \llms' empirical distributions. 
This analysis is motivated by the observation that, visually, the empirical distributions obtained via greedy decoding appear to be less diverse than the population-level human distributions. 
In addition to drawing comparisons between humans and \llms, we also inspect how the observed variability changes with different decoding algorithms, namely, we consider a probabilistic decoding algorithm \texttt{temperature=1}.

\textbf{\textit{Metric. }}
As a measure of the variability of an empirical distribution, we consider the \emph{interquartile range (IQR)}~\citep{MOHR20221-iqr}, \ie the difference between the 0.75 and 0.25 quantiles.\footnote{Other metrics could also be used to study the variability of the distributions, such as the entropy. However, we leave further analysis for future work.} 
Intuitively, a smaller IQR value implies that the models' predictions tend to be concentrated in the same set of values, whereas a larger IQR value suggests a more uniform distribution.
Because we have an empirical distribution for each uncertainty expression, we use the average IQR across uncertainty expressions to analyse the overall behavior. 

\textbf{\textit{Results. }}
In the non-verifiable setting, we find that most models exhibit lower average IQR values than population-level human perceptions (see Table \ref{tab:app:variability:iqr-2shot-and-0shot:non-verifiable}). 
In particular, the values reported for \gptf, \gptfo, and \llama are up to 13x smaller than the one reported for humans. 
These findings suggest that, when using greedy decoding, these \llms are unable to replicate the diversity of human responses. 

When considering the change in avg IQR values for probabilistic decoding in Table \ref{tab:app:variability:iqr-2shot-and-0shot:non-verifiable}, we observe an increase in the spread for most models with respect to the greedy decoding values (+6.60 and +10.60 IQR increase on average for 2-shot and 0-shot, respectively). 
In fact, from the evaluated models, \gptf and \llama remain mostly insensitive to the change of decoding algorithm (between 0 and 1.20) whereas the IQR values for \chatgpt and \gptfo increase +4.38 and +35.3, respectively. 
These results seem to suggest that for some models, the variability of the observed empirical distributions may be a function of the employed decoding algorithm.

\begin{table*}[tb]
\centering
\caption{\textbf{InterQuartile Range (IQR) of the empirical distributions averaged across all uncertainty expressions in the non-verifiable setting}. Reported values include both histograms created using greedy decoding (\texttt{temperature=0}) and random decoding (\texttt{temperature=1}). While most models exhibit become more diverse when using random decoding, models like \gptf and \llama seem to be minimally affected by the change in decoding algorithm, suggesting that these models lead to less diverse results (when compared to humans).}
\label{tab:app:variability:iqr-2shot-and-0shot:non-verifiable}
\begin{tabular}{l l cc cc}
\toprule
\multicolumn{1}{l}{} & \multicolumn{1}{l}{} & \multicolumn{2}{c}{2-shot prompt} & \multicolumn{2}{c}{0-shot prompt} \\
\cmidrule(lr){3-4}
\cmidrule(lr){5-6}
Models & Methodology & Greedy  & Probabilistic & Greedy  & Probabilistic \\
\midrule
Humans & --- & 15.00 & --- & --- & --- \\
\hdashline
 \chatgpt & top-k & 4.62 & 9.00 & 8.08 & 11.54 \\
 \gptf  & top-k & 1.15 & 1.15 & 1.15 & 2.69 \\
 \gptfo & top-k & 3.85 & 39.15 & 3.85 & 39.00 \\
\addlinespace
\llama & full & 1.92 & 3.08 & 0.77 & 1.54 \\
\llamasmall & full & 15.38 & 21.54 & 10.00 & 21.92 \\
\olmo & full & 36.54 & 39.23 & 36.15 & --- \\
\gemma  & full & 23.08 & 19.62 & 0.77 & 11.54 \\
\addlinespace
\mixtralmoelg & sampling & 5.00 & --- & 5.77 & --- \\
\mixtralmoe & sampling & 4.23 & --- & 4.62 & ---\\
\bottomrule
\end{tabular}
\end{table*}

\subsection{Histograms}
\label{app:ssec:add-results:histograms}

\begin{figure*}[tb]
    \centering
    \begin{subfigure}[b]{0.45\textwidth}
        \centering
        \includegraphics[width=\textwidth]{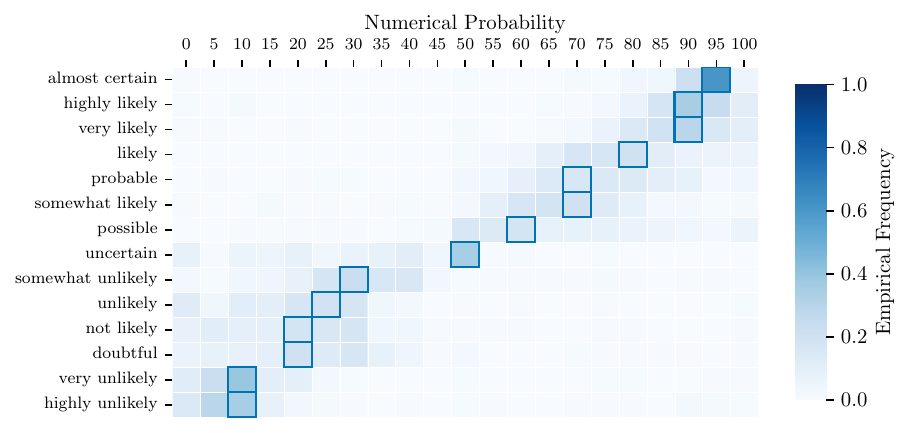}
        \caption{Humans}
        \label{sfig:hist-nv:humans}
    \end{subfigure}
    \hfill
    \begin{subfigure}[b]{0.45\textwidth}
        \centering
        \includegraphics[width=\textwidth]{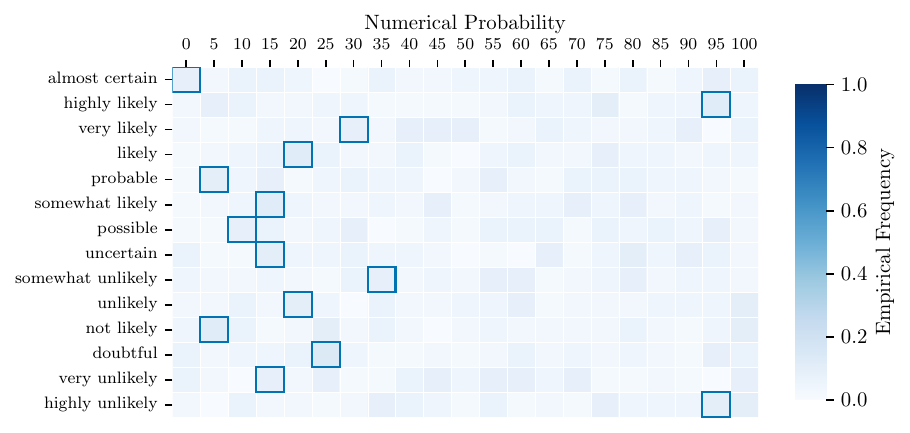}
        \caption{Random}
        \label{sfig:hist-nv:random}
    \end{subfigure}
    \begin{subfigure}[b]{0.45\textwidth}
        \centering
        \includegraphics[width=\textwidth]{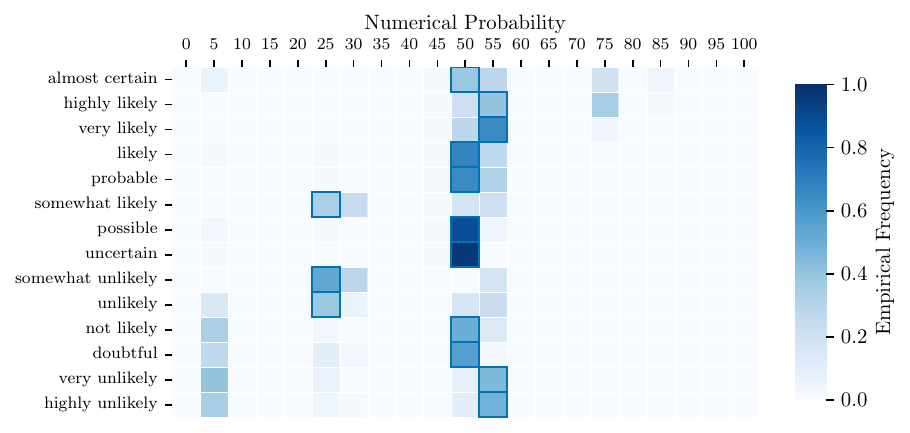}
        \caption{\gemma}
        \label{sfig:hist-nv:gemma}
    \end{subfigure}
    \hfill
    \begin{subfigure}[b]{0.45\textwidth}
        \centering
        \includegraphics[width=\textwidth]{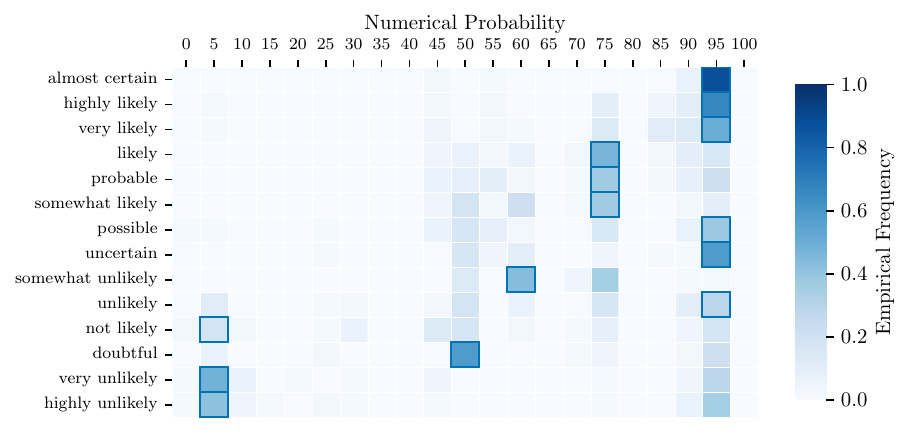}
        \caption{\olmo}
        \label{sfig:hist-nv:olmo}
    \end{subfigure}
    \begin{subfigure}[b]{0.45\textwidth}
        \centering
        \includegraphics[width=\textwidth]{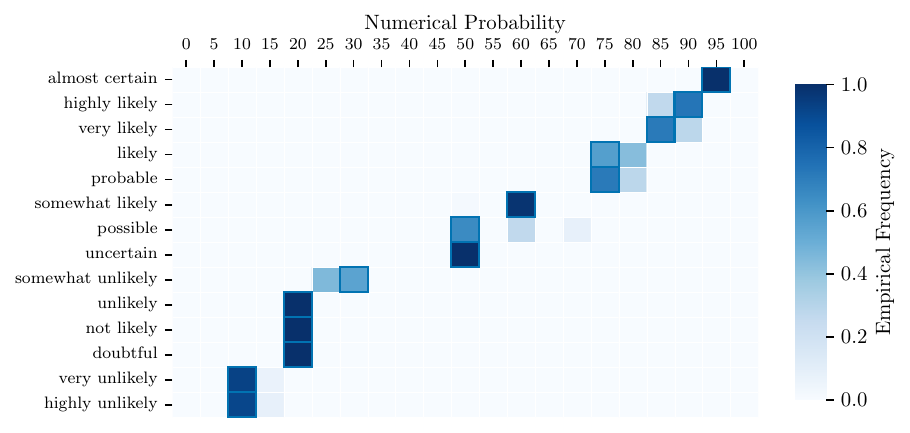}
        \caption{\gemini}
        \label{sig:hist-nv:gemini-pro}
    \end{subfigure}
    \hfill
    \begin{subfigure}[b]{0.45\textwidth}
        \centering
        \includegraphics[width=\textwidth]{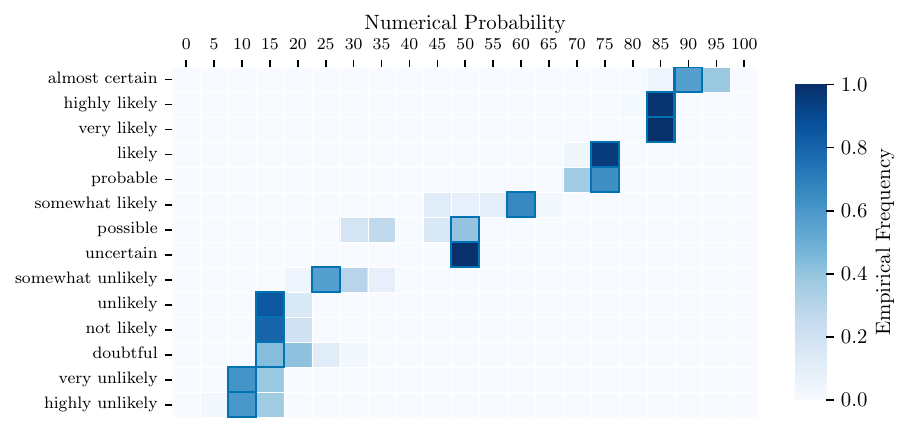}
        \caption{\gptfo}
        \label{sfig:hist-nv:gpt4-o}
    \end{subfigure}
    \begin{subfigure}[b]{0.45\textwidth}
        \centering
        \includegraphics[width=\textwidth]{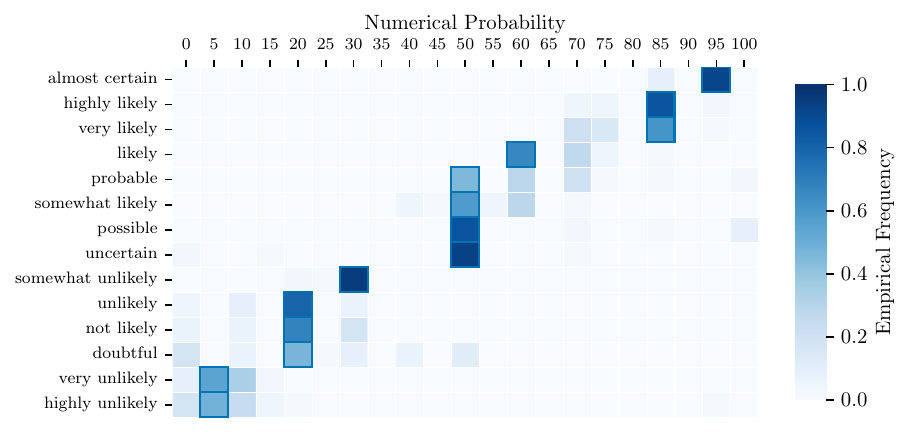}
        \caption{\chatgpt}
        \label{sig:hist-nv:chatgpt}
    \end{subfigure}
    \hfill
    \begin{subfigure}[b]{0.45\textwidth}
        \centering
        \includegraphics[width=\textwidth]{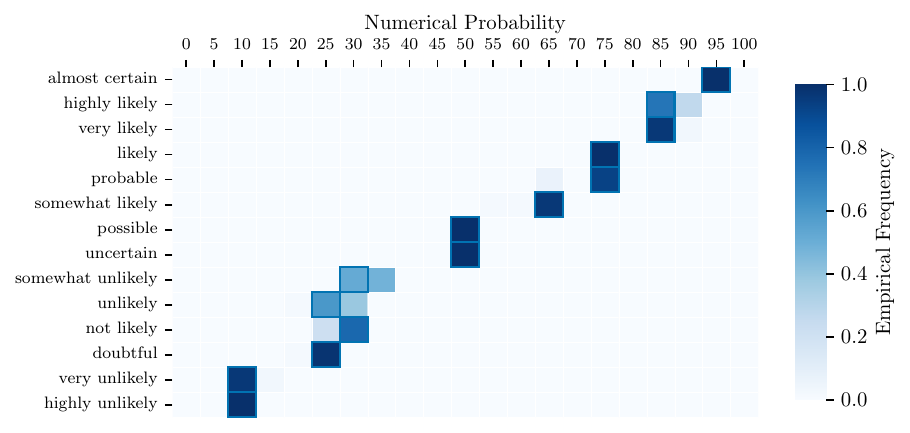}
        \caption{\gptf}
        \label{sfig:hist-nv:gpt4}
    \end{subfigure}
    \begin{subfigure}[b]{0.45\textwidth}
        \centering
        \includegraphics[width=\textwidth]{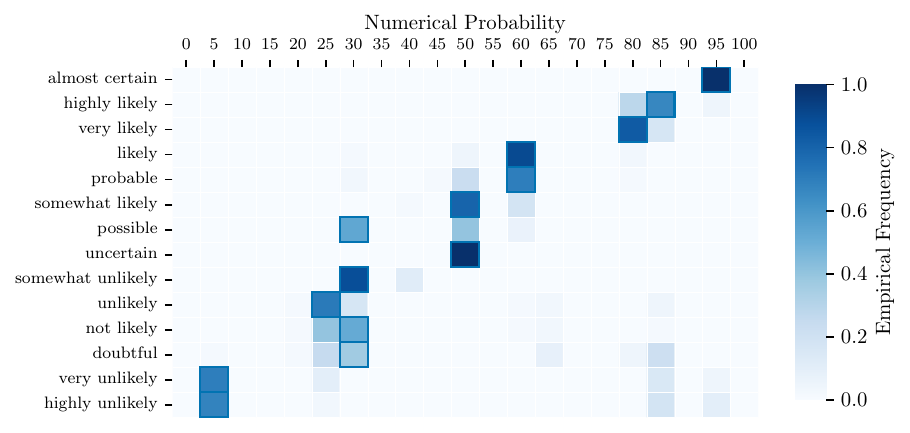}
        \caption{\llamasmall}
        \label{sig:hist-nv:llama3-8b}
    \end{subfigure}
    \hfill
    \begin{subfigure}[b]{0.45\textwidth}
        \centering
        \includegraphics[width=\textwidth]{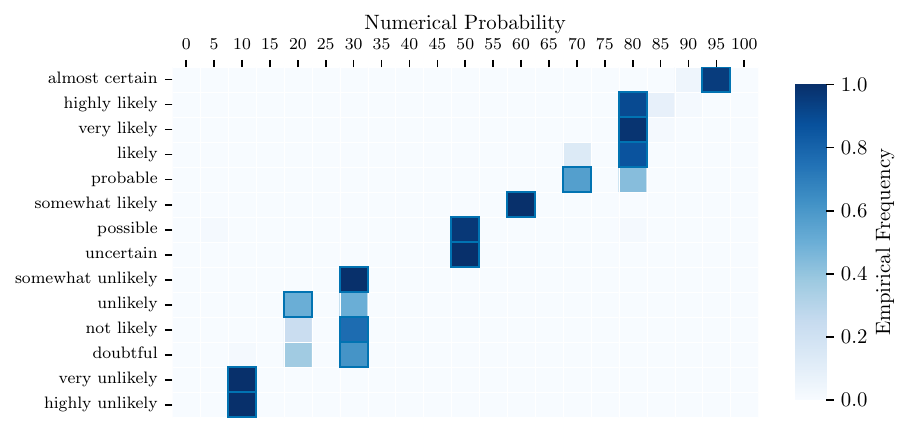}
        \caption{\llama}
        \label{sfig:hist-nv:llama3-70b}
    \end{subfigure}
    \begin{subfigure}[b]{0.45\textwidth}
        \centering
        \includegraphics[width=\textwidth]{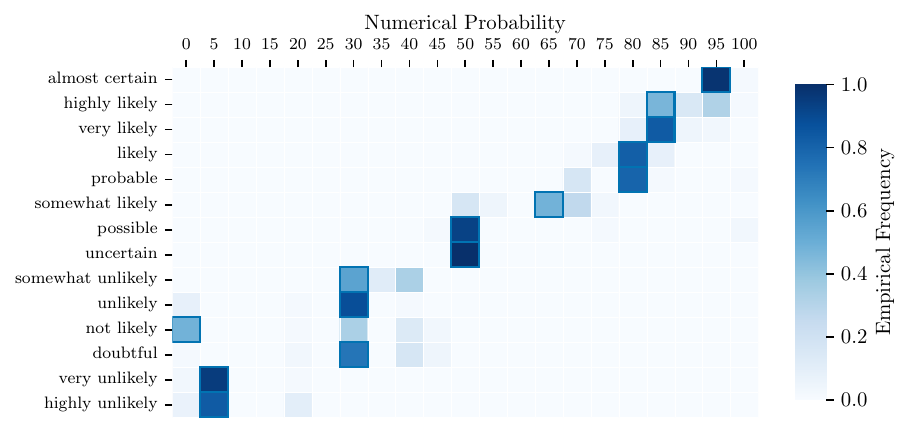}
        \caption{\mixtralmoe}
        \label{sfig:hist-nv:mistral--8x7b}
    \end{subfigure}
    \hfill
    \begin{subfigure}[b]{0.45\textwidth}
        \centering
        \includegraphics[width=\textwidth]{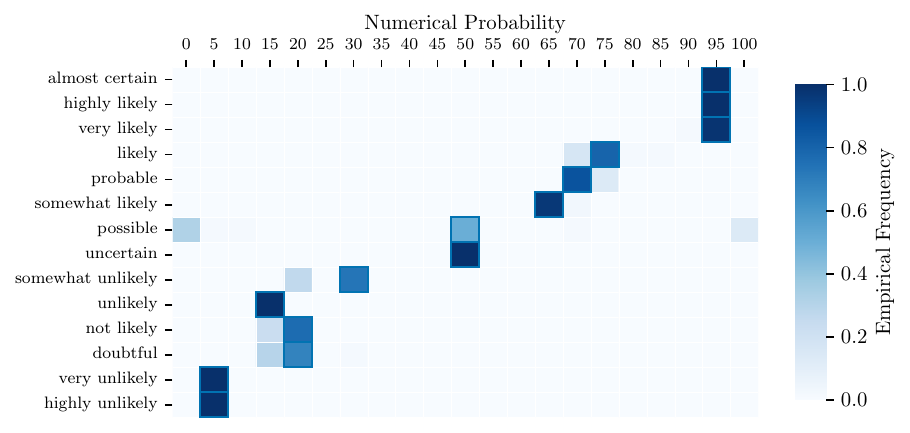}
        \caption{\mixtralmoelg}
        \label{sfig:hist-nv:mistral--8x22b}
    \end{subfigure}
    \caption{\textbf{Models (and Human) empirical distributions of numerical responses per uncertainty expression in the non-verifiable setting (\texttt{NV})}. The empirical distributions are estimated using the greedy decoding algorithm (\texttt{temperature=1}). Highlighted boxes represent the mode value for each expression. With the exception of \olmo and \gemma, \llms are able to replicate the overall population-level human behavior: assigning lower numerical responses to uncertainty expressions (\eg ``highly unlikely'' and ``unlikely'') and progressively higher numerical responses until reaching the certainty expressions (\eg ``almost certain'').}
    \label{fig:hist-nv}
    \vspace{-2em}
\end{figure*}

Figure \ref{fig:hist-nv} depicts the empirical distributions for the non-verifiable experiments.

\section{Generalization results}
\label{app:sec:generalization-results}

Diversity of grammatical and semantic structures is an important component of current evaluation practices in LLMs~\citep{selvam-etal-2023-tail,seshadri2022quantifying}, since it helps ensure that obtained results are not an artifact of the evaluation methodology and/or benchmarks used.
The experiments described in the main paper were carefully crafted to cover various topics and situations where uncertainty expressions could be used.
To further strengthen our analysis and validate our findings, we simultaneously run collect models perceptions of uncertainty expressions using a larger dataset.
This dataset by the authors based on the AI2-ARC test set~\citep{Clark2018ThinkYH} --- a popular question-answering dataset consisting of genuine grade-school level, multiple-choice science questions. Not only has this dataset been recently used to measure commonsense reasoning of current state-of-the-art LLMs \citep{jiang2023mistral, openai2024gpt4,open-llm-leaderboard}, but it is also composed of easier questions, a key aspect to our verifiable experiment setup.

The creation of this dataset mirrors the procedure described in Section \ref{sec:methodology}. We manually repurposed 200 question-answer pairs from AI2-ARC (100 from the easy set and another 100 from the challenge set). For every statement, the authors produce a true statement and a false statement using the available information about the correct and incorrect multiple choices. The final dataset consists of 200 true statements and 200 false statements. 

To determine distributional differences between the conditional distributions obtained in the main paper and the ones obtained in the generalization set, we compare the Wasserstein-1 distance of the two empirical distributions. 
These values are reported in Table \ref{tab:app:wasserstein-dist-default-vs-gen-exp}.
In general, we find models that performed worse in the main paper, including \gemma and \olmo, to exhibit the largest distributional differences with Wasserstein distances of 48.5 and 13.1 when averaged over uncertainty expressions. 
\chatgpt, \llama, and \texttt{Mixtral} models all exhibit higher differences in expressions of higher certainty, e.g., ``highly likely'', ``probable'', ``possible''. 
On the other hand, the two \gptf models, as well as Gemini (Pro) suffer the least changes distributionally (1.9, 1.8, and 4.2 Wasserstein-1 distances on average, respectively), suggesting that these models were robust to changes in the statements.

\begin{table*}[tb]
\centering
\caption{\textbf{Analysis of the distributional differences between the conditional distributions estimated using greedy (\texttt{temperature}=0) \textit{versus} probabilistic decoding (\texttt{temperature=1})}. We use Wasserstein-1 distance to report the distributional differences between each conditional distribution: maximally distant distributions exhibit a score of 101.}
\label{tab:app:wasserstein-dist-default-vs-gen-exp}
\begin{tabular}{l cc cc cc}
\toprule
\multicolumn{1}{l}{} & \multicolumn{2}{c}{0-shot prompting} & \multicolumn{4}{c}{2-shot prompting} \\
\cmidrule(lr){2-3}
\cmidrule(lr){4-7}
Model       & NV & V & NV & V & AI2-ARC (Easy) & AI2-ARC (Challenge) \\
\midrule
\chatgpt    & 3.28 & 3.40 & 3.32       & 2.75  & 2.87  & 3.08 \\ 
\gptf       & 0.31 & 0.52 & 0.22       & 0.84  & 0.49  & 0.78 \\
\gptfo      & 19.88 & 15.51 &  15.38   & 19.86 & 20.70 & 20.97 \\
\bottomrule
\end{tabular}
\end{table*}

\begin{figure*}[tb]
    \centering
    \includegraphics[width=\textwidth]{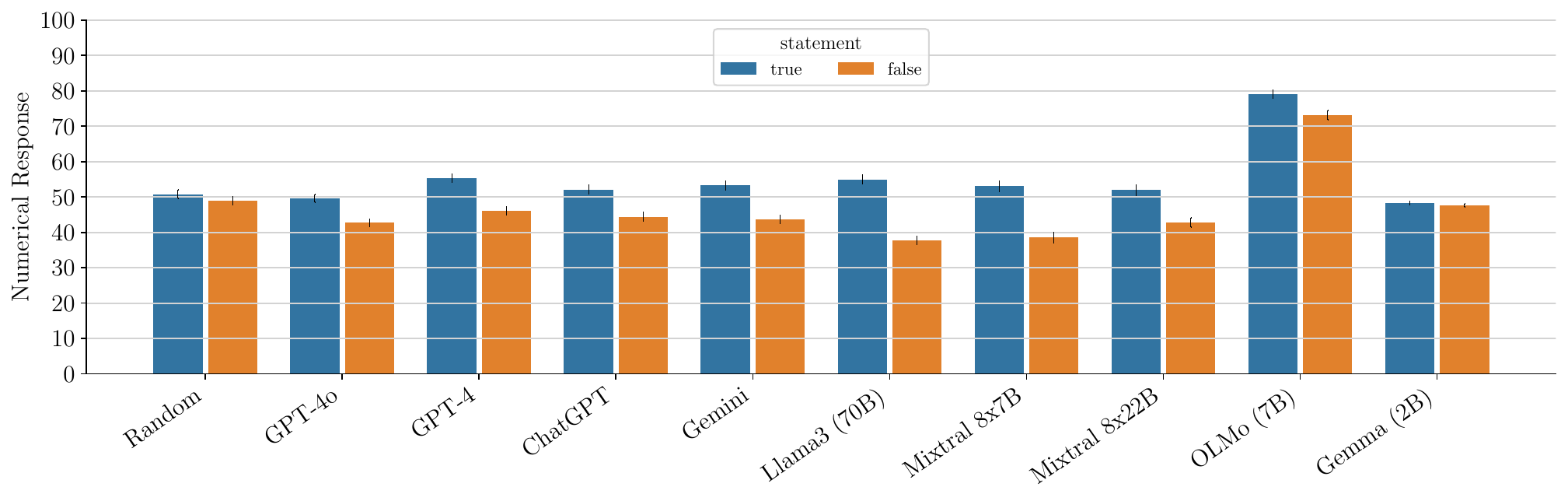}
    \caption{\textbf{Mean numerical response discriminated by truthfulness of statements for 400 verifiable statements derived from the AI2-ARC dataset}. The mean numerical responses produced by LLMs when evaluated in the context of true statements is significantly larger than when evaluated with the false statements. Although the numerical response gap is lower in magnitude than those observed in the verifiable experiment (see Figure \ref{fig:truevfalse_statement}), the observed gap is still statistically significant.}
    \label{fig:generalization-results-bias-shift-average}
\end{figure*}

\begin{figure*}[tb]
    \centering
    \begin{subfigure}[b]{0.32\textwidth}
        \centering
        \includegraphics[width=\textwidth]{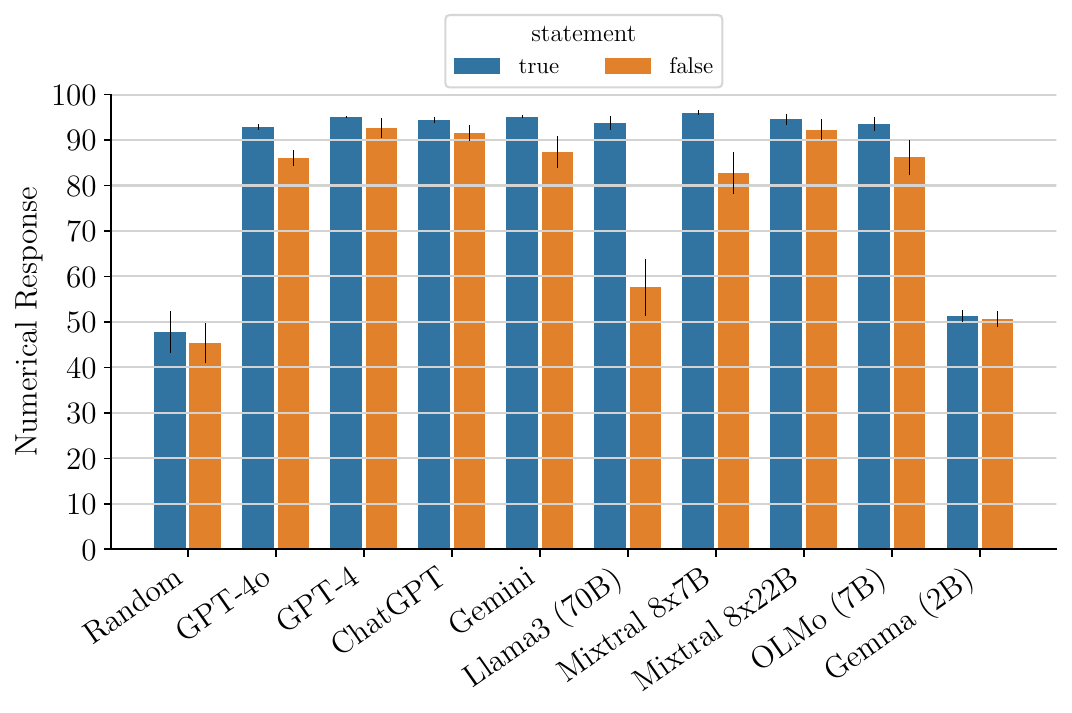}
        \caption{almost certain}
        \label{sfig:mean-rated-prob:almost-certain:generalization}
    \end{subfigure}
    \hfill
    \begin{subfigure}[b]{0.32\textwidth}
        \centering
        \includegraphics[width=\textwidth]{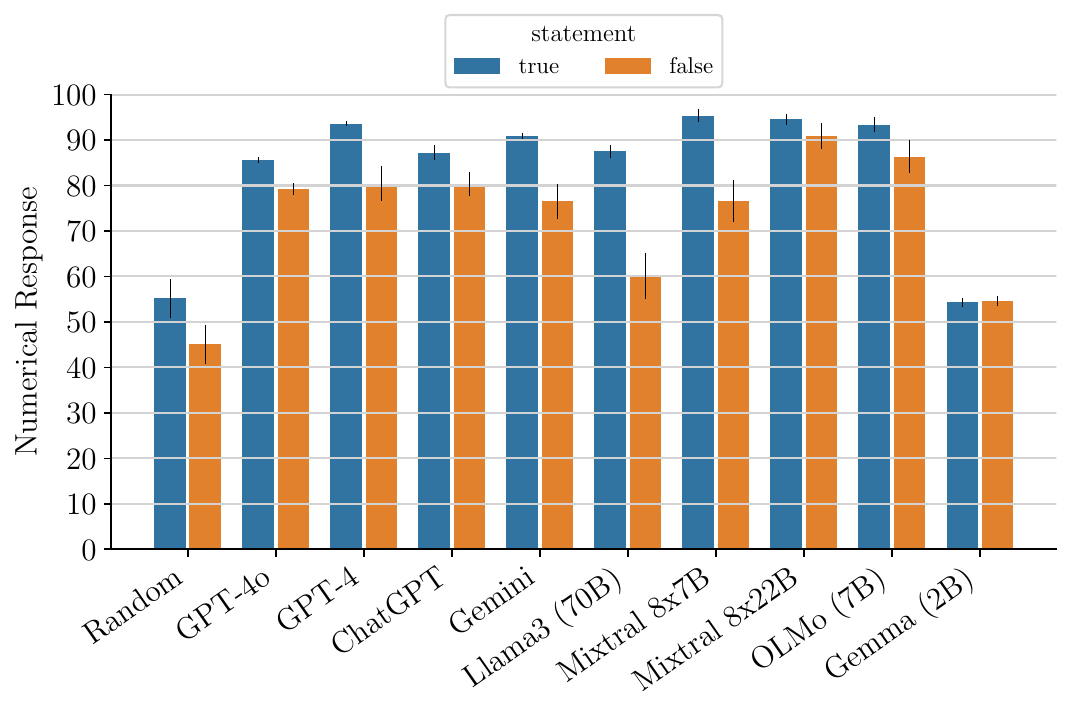}
        \caption{``highly likely''}
        \label{sfig:mean-rated-prob:highly-likely:generalization}
    \end{subfigure}
    \hfill
    \begin{subfigure}[b]{0.32\textwidth}
        \centering
        \includegraphics[width=\textwidth]{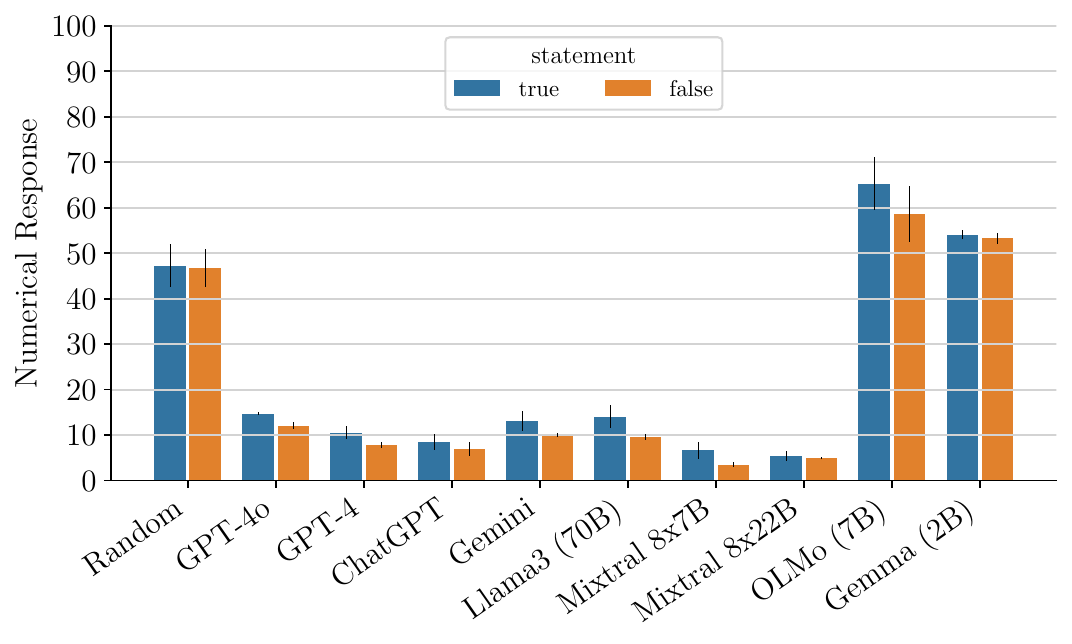}
        \caption{``highly unlikely''}
        \label{sfig:mean-rated-prob:highly-unlikely:generalization}
    \end{subfigure}
    \begin{subfigure}[b]{0.32\textwidth}
        \centering
        \includegraphics[width=\textwidth]{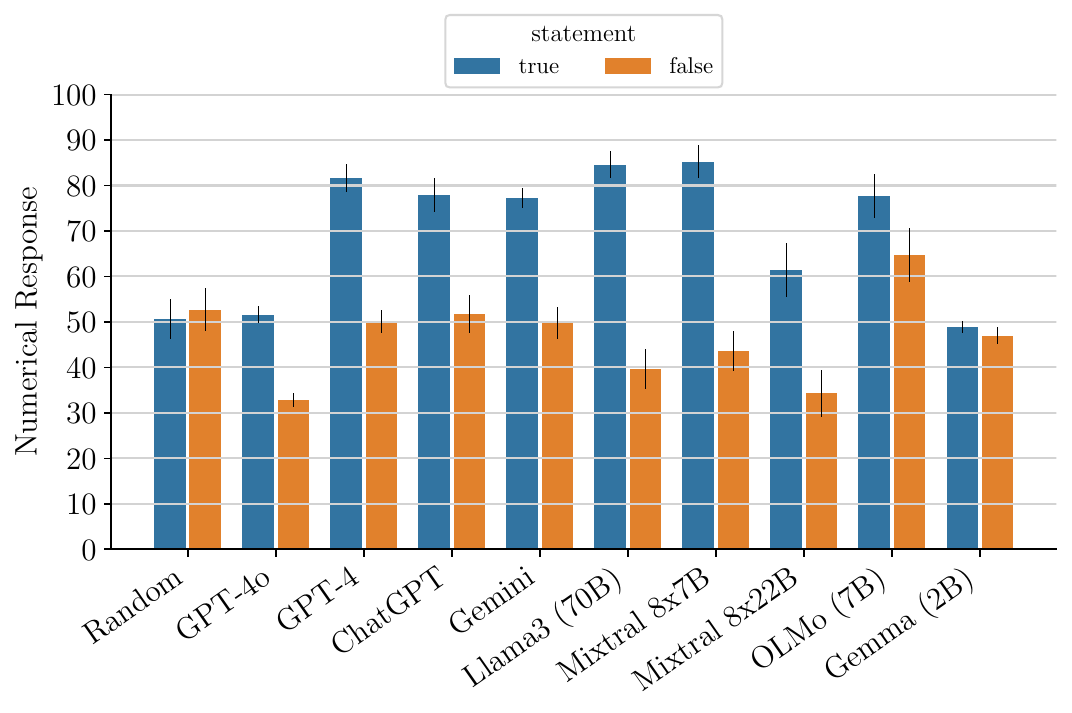}
        \caption{``possible''}
        \label{sfig:mean-rated-prob:possible:generalization}
    \end{subfigure}
    \hfill
    \begin{subfigure}[b]{0.32\textwidth}
        \centering
        \includegraphics[width=\textwidth]{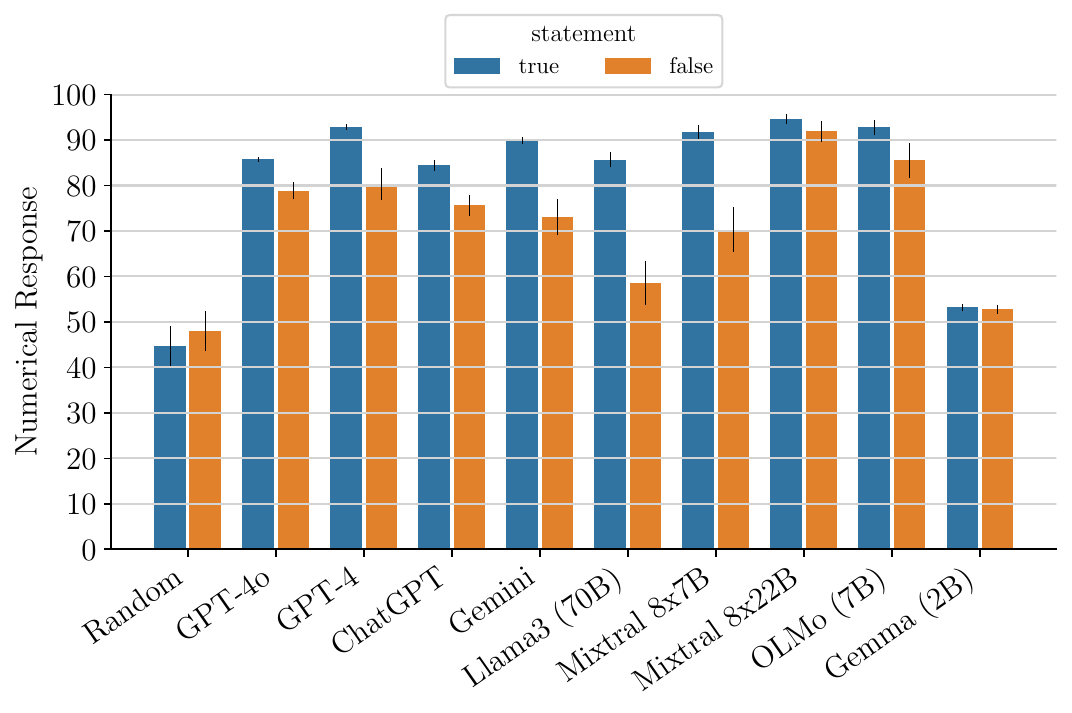}
        \caption{``very likely''}
        \label{sfig:mean-rated-prob:very-likely:generalization}
    \end{subfigure}
    \hfill
    \begin{subfigure}[b]{0.32\textwidth}
        \centering
        \includegraphics[width=\textwidth]{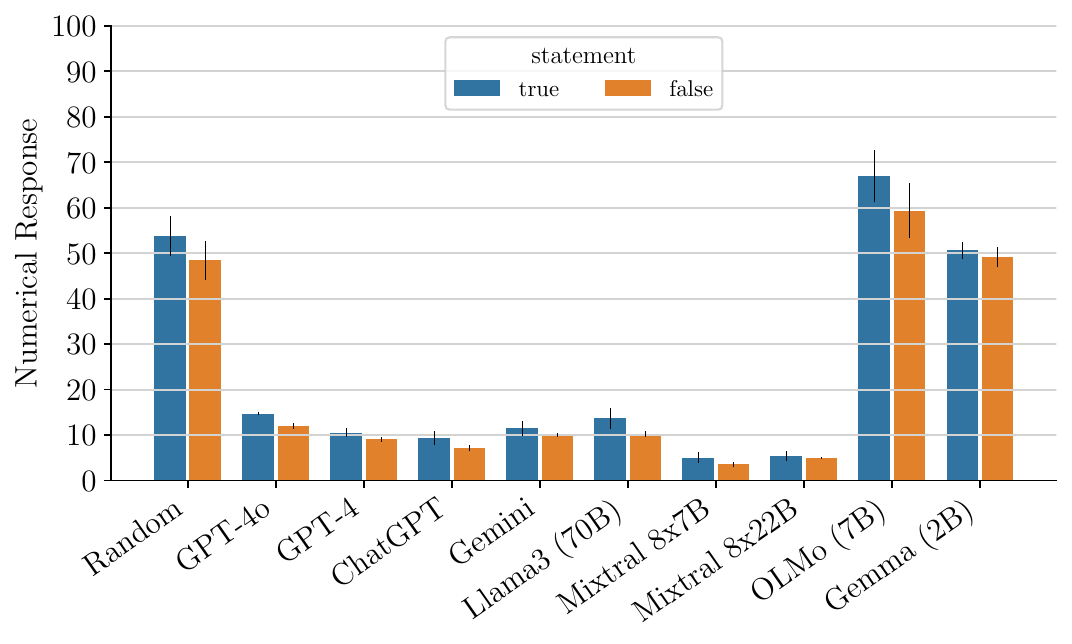}
        \caption{``very unlikely''}
        \label{sfig:mean-rated-prob:very-unlikely:generalization}
    \end{subfigure}
    \begin{subfigure}[b]{0.32\textwidth}
        \centering
        \includegraphics[width=\textwidth]{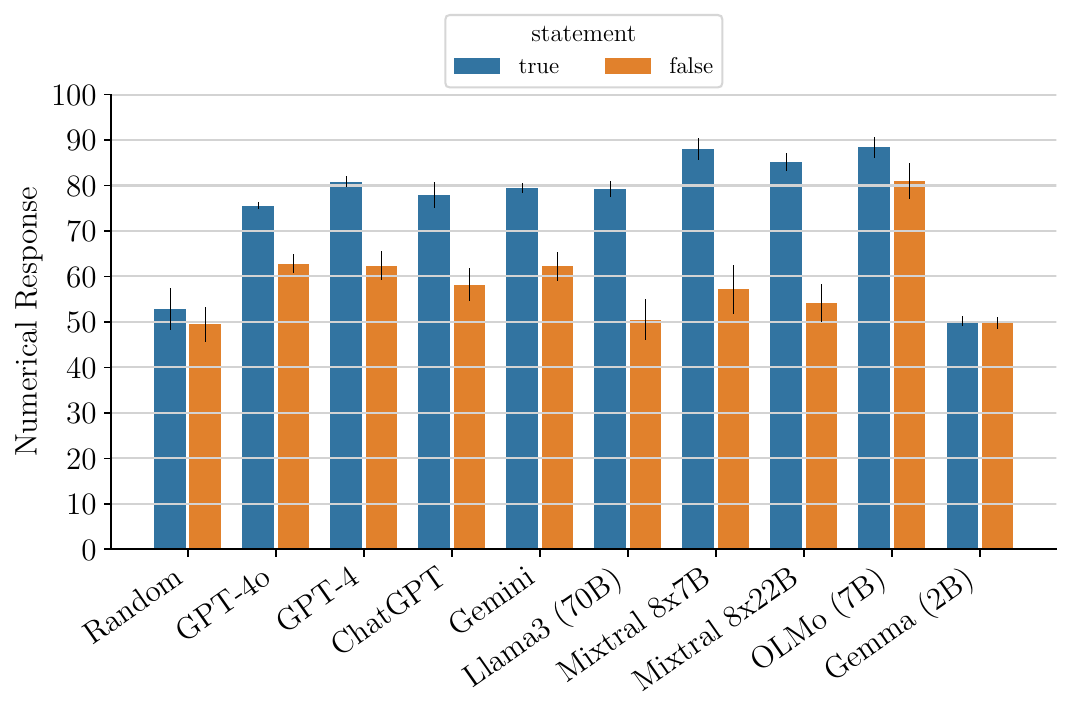}
        \caption{``probable''}
        \label{sfig:mean-rated-prob:probable:generalization}
    \end{subfigure}
    \begin{subfigure}[b]{0.32\textwidth}
        \centering
        \includegraphics[width=\textwidth]{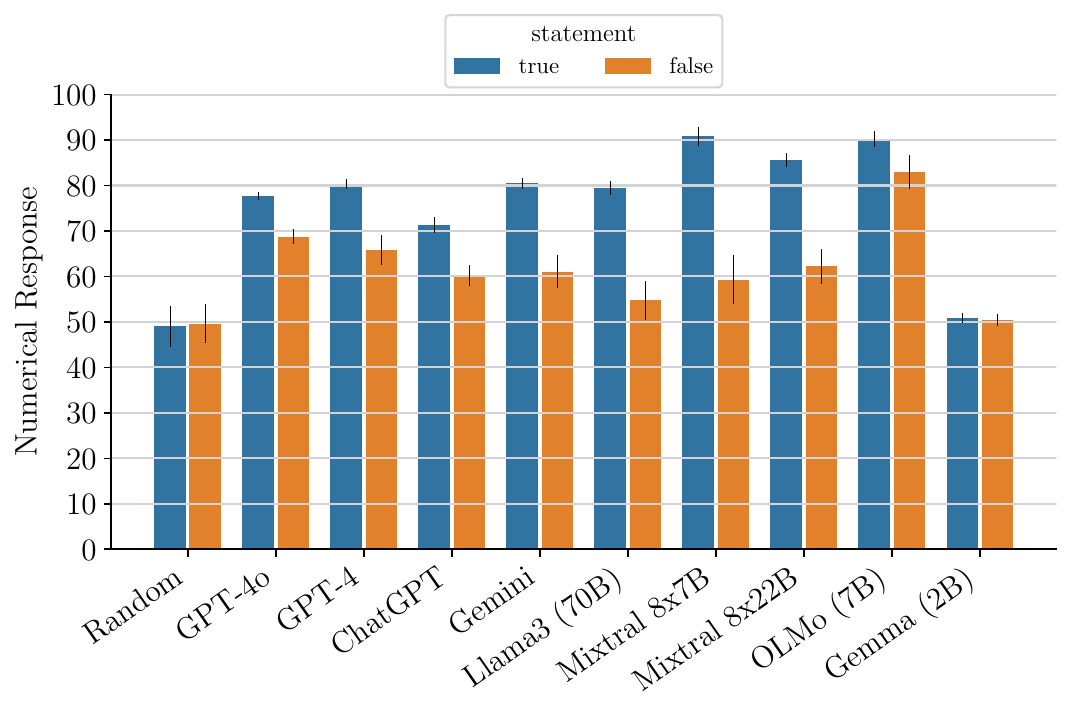}
        \caption{``likely''}
        \label{sfig:mean-rated-prob:likely:generalization}
    \end{subfigure}
    \hfill
    \begin{subfigure}[b]{0.32\textwidth}
        \centering
        \includegraphics[width=\textwidth]{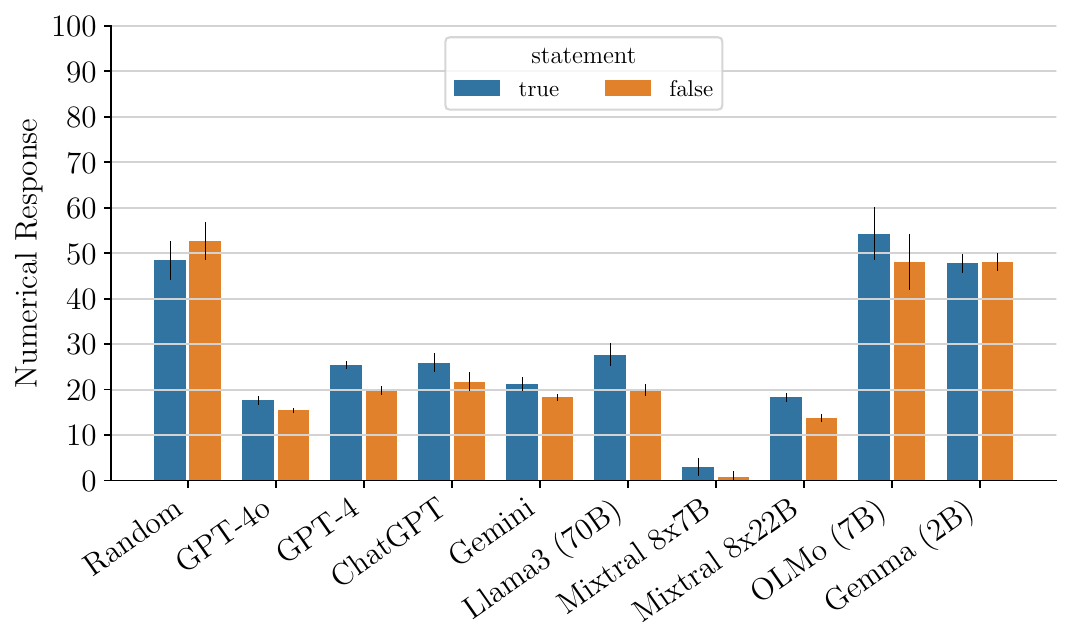}
        \caption{``not likely''}
        \label{sfig:mean-rated-prob:not-likely:generalization}
    \end{subfigure}
    \hfill
    \begin{subfigure}[b]{0.32\textwidth}
        \centering
        \includegraphics[width=\textwidth]{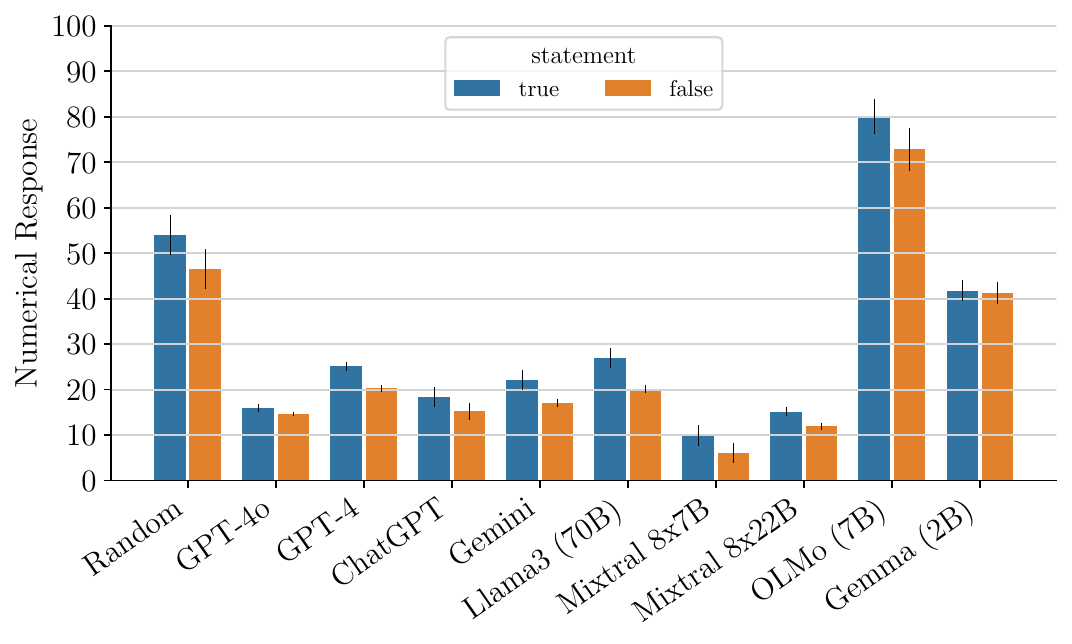}
        \caption{``unlikely''}
        \label{sfig:mean-rated-prob:unlikely:generalization}
    \end{subfigure}
    \begin{subfigure}[b]{0.32\textwidth}
        \centering
        \includegraphics[width=\textwidth]{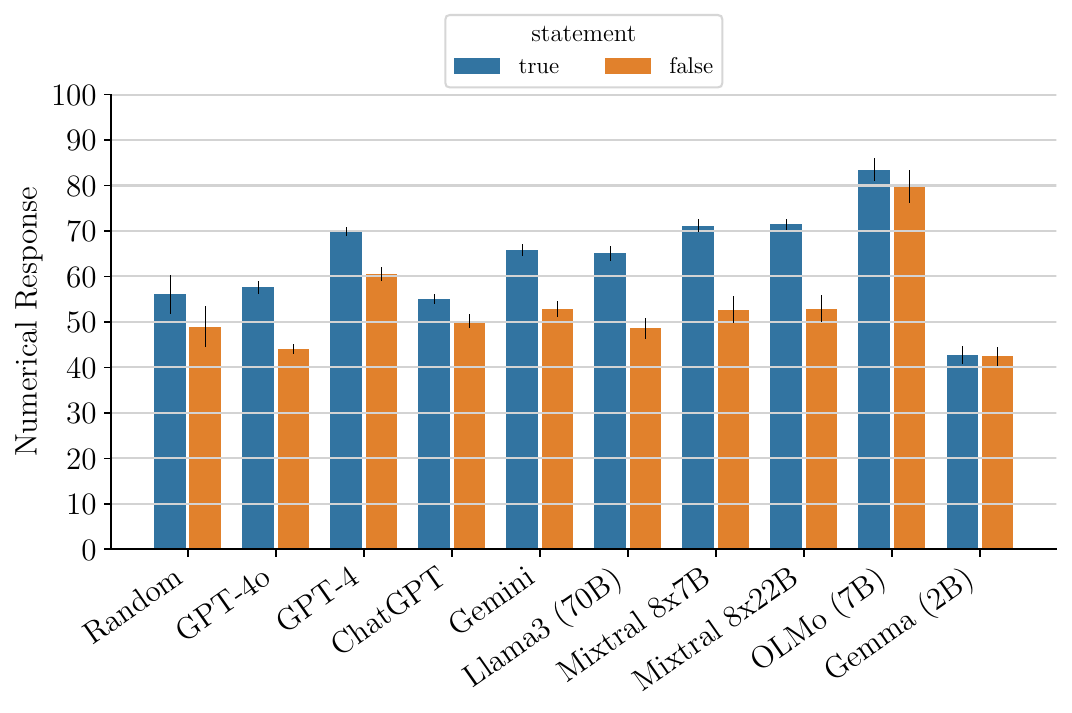}
        \caption{``somewhat likely''}
        \label{sfig:mean-rated-prob:somewhat-likely:generalization}
    \end{subfigure}
    \hfill
    \begin{subfigure}[b]{0.32\textwidth}
        \centering
        \includegraphics[width=\textwidth]{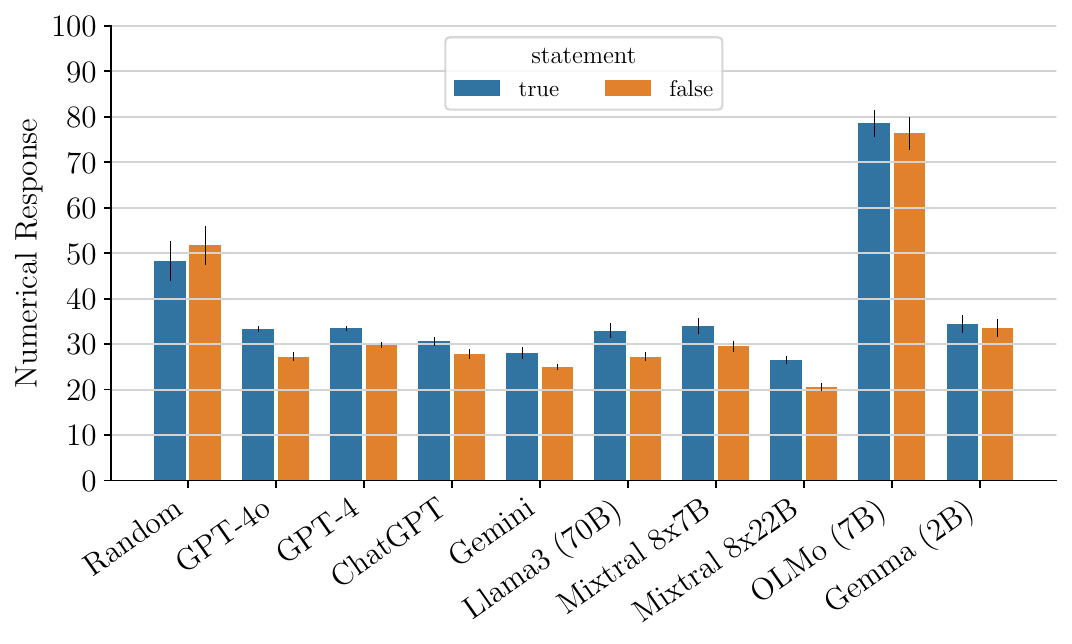}
        \caption{``somewhat unlikely''}
        \label{sfig:mean-rated-prob:somewhat-unlikely:generalization}
    \end{subfigure}
    \begin{subfigure}[b]{0.32\textwidth}
        \centering
        \includegraphics[width=\textwidth]{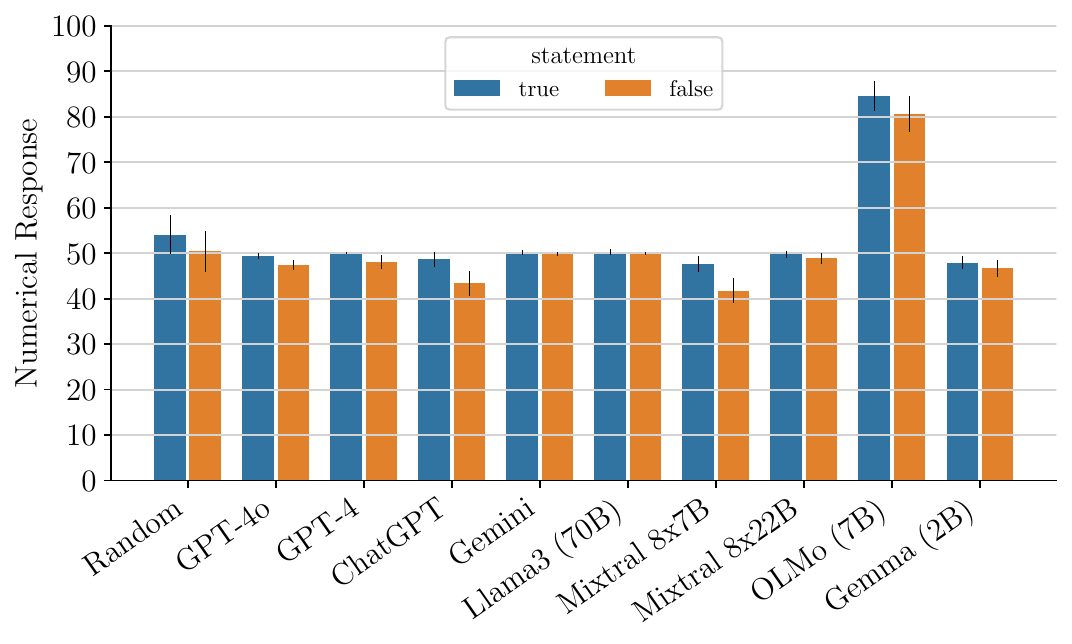}
        \caption{``uncertain''}
        \label{sfig:mean-rated-prob:uncertain:generalization}
    \end{subfigure}
    \begin{subfigure}[b]{0.32\textwidth}
        \centering
        \includegraphics[width=\textwidth]{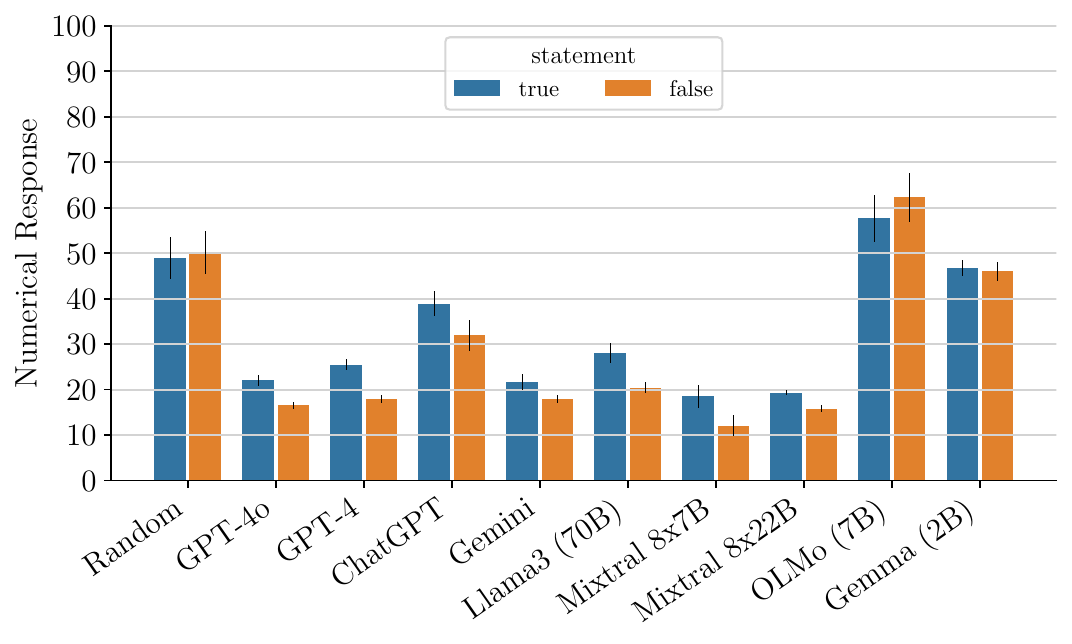}
        \caption{``doubtful''}
        \label{sfig:mean-rated-prob:doubtful:generalization}
    \end{subfigure}
    \caption{\textbf{Mean numerical response for verifiable statements discriminated by truthfulness of statements for the AI2-ARC dataset}. While the observed gaps are smaller in magnitude than in the main experiment, there is still a significant difference between mean numerical responses depending on the truthfulness of the statements. We hypothesize that the observed magnitude differences between the two datasets may be explained by semantic differences between the two QA datasets used to curate the verifiable statements. Future work should investigate this behavior more closely.}
    \label{fig:mean-rated-prob-generalization}
\end{figure*}

In the main paper, we find it surprising that LLMs perception abilities differ significantly based on whether the uncertainty expressions are referring to someone's belief in a true or false statement. 
To test the generalization of this finding in a larger (and different) dataset, we repeat the same analysis and compare the observed mean response differences with that of humans obtained in the original setting (see Figures \ref{fig:generalization-results-bias-shift-average} and \ref{fig:mean-rated-prob-generalization}). 
We observe that in absolute sense the differences are smaller than those observed in the original setting, but that models are affected by this knowledge gap to a greater extent than humans. 

\section{Probabilistic Decoding}
\label{app:sec:fractional-analysis}

The findings reported in the main paper mostly concern the conditional distributions that are estimated using greedy decoding algorithm. 
Despite being a common decision in analyses papers~\citep{yona2024largelanguagemodelsfaithfully,steyvers2024calibration}, the use of greedy decoding may not provide the full picture of model behavior~\citep{ivgi2024loopsoopsfallbackbehaviors,Holtzman2020The}. 
Thus, to ensure that our results are not a degenerate behavior only observed when using greedy decoding algorithms, we conduct an analysis considering a probabilistic decoding (\texttt{temperature=1}): instead of estimating the conditional distributions for each uncertainty expression using the arg-max predicted numerical response, we use the available probability information to estimate the conditional distributions. 

Most of our analyses on the study of OpenAI models--- \chatgpt, \gptf, and \gptfo---using the top-k approach. 
Using OpenAI models has the benefit of being more cost-efficient than the other approaches, \ie it is less time-consuming and lower cost than the full probability methodology or the sampling-based approach. 
Moreover, even though theoretically the use of full probability information is better for estimating the conditional distributions, we found that the smaller evaluated open-source models (\eg \olmo and \gemma) were extremely sensitive to the prompt formatting and, frequently resulted in negligible probability mass being assigned to integers in the range [0, 100].

\begin{figure}[tb]
    \centering
    \begin{subfigure}[b]{\linewidth}
        \centering
        \includegraphics[width=\linewidth]{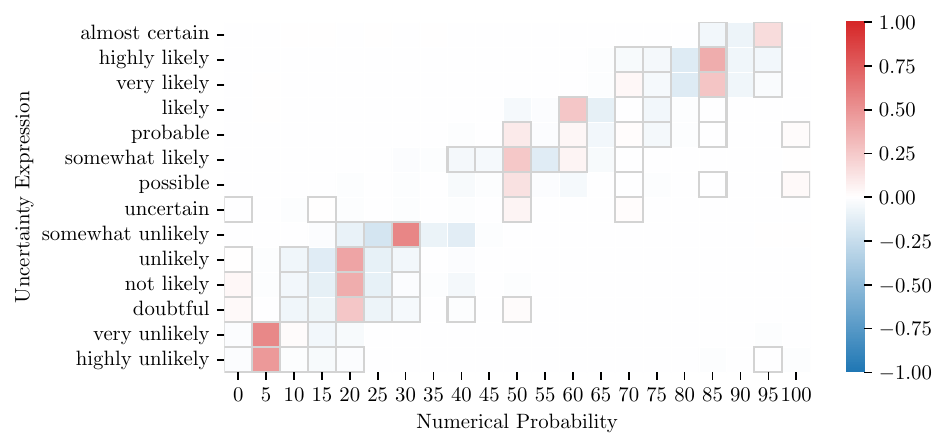}
        \caption{\chatgpt}
        \label{sfig:fract-hist-diff-greedy-minus-rich-chatgpt}
    \end{subfigure}
    
    \begin{subfigure}[b]{\linewidth}
        \centering
        \includegraphics[width=\linewidth]{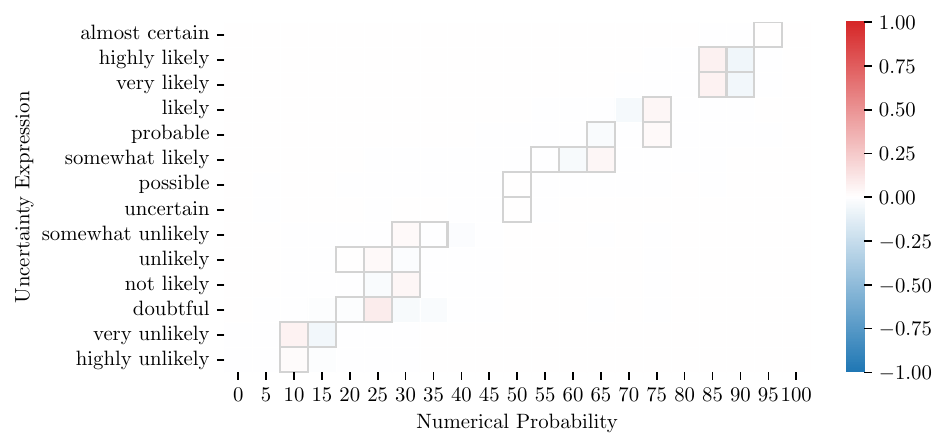}
        \caption{\gptf}
        \label{sfig:fract-hist-diff-greedy-minus-rich-gpt4}
    \end{subfigure}
    
    \begin{subfigure}[b]{\linewidth}
        \centering
        \includegraphics[width=\linewidth]{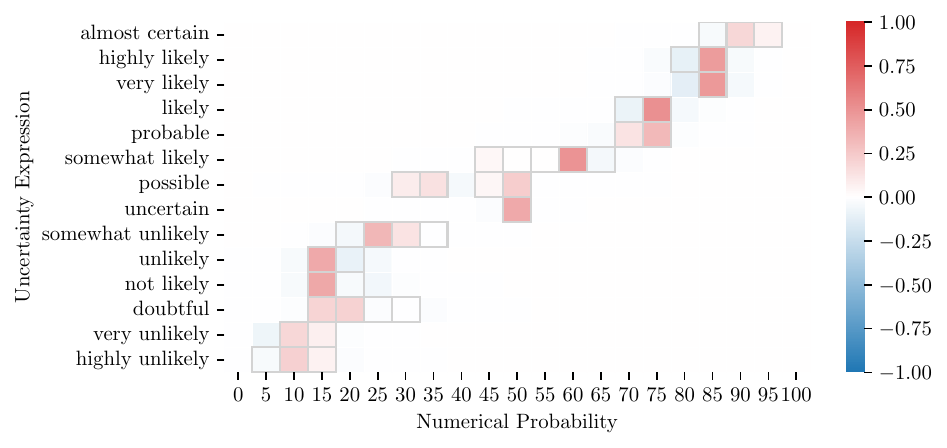}
        \caption{\gptfo}
        \label{sfig:fract-hist-diff-greedy-minus-rich-gpt4-o}
    \end{subfigure}
    \caption{\textbf{Absolute Difference between the models' conditional probability distribution when estimated using greedy (\texttt{temperature=0}) \textit{versus} probabilistic decoding (\texttt{temperature=1})}. The results refer to the non-verifiable dataset. Gray rectangles indicate bins whose empirical distribution with the greedy information has non-zero value.
    Bins are colored blue if the bin's estimated probability is larger when using the probabilistic approach and are red if the probability was larger using the greedy approach. While we observe that the decoding choice has a larger impact for \chatgpt and \gptfo, we find that it has minimal impact in \gptf's distributions, suggesting that \gptf's predictions tend to be very confident in its predictions to begin with.}
    \label{fig:fractional-analysis-diffs-cond-dist-nv}
    \vspace{-1em}
\end{figure}

\begin{table*}[tb]
\centering
\caption{\textbf{Summary metrics averaged across uncertainty expressions for both NV and V settings when using probabilistic decoding (\texttt{temperature=1})}. All metrics are computed with respect to the human distribution in the non-verifiable setting (\texttt{Human+NV}). ``PA'' reports the general agreement between LLMs and the mode of the human  distribution, reported in percentages. ``MAE'' reports the absolute error between the mean responses of LLMs and those of humans. Wasserstein-1 computes the distance between LLMs and human distributions.}
\label{tab:fractional:summary-metrics}
\begin{tabular}{llcccccc}
\toprule
&  & \multicolumn{2}{c}{Avg  PA ($\uparrow$)} & \multicolumn{2}{c}{Avg MAE}($\downarrow$) & \multicolumn{2}{c}{Avg Wasserstein-1 ($\downarrow$)} \\
\cmidrule(lr){3-4}
\cmidrule(lr){5-6}
\cmidrule(lr){7-8}
&  & \multicolumn{1}{c}{NV} & \multicolumn{1}{c}{V} & \multicolumn{1}{c}{NV} & \multicolumn{1}{c}{V} & \multicolumn{1}{c}{NV} & \multicolumn{1}{c}{V} \\
\midrule
\multirow[t]{2}{*}{Human} & Mode & \cellcolor[HTML]{c0e6b9}{27.6} & \cellcolor[HTML]{c0e6b9}{27.6} & --- & --- & --- & --- \\
 & Individual & \cellcolor[HTML]{daf0d4}{17.6} & \cellcolor[HTML]{dbf1d6}{16.7} & \cellcolor[HTML]{fdd1be}{8.91} & \cellcolor[HTML]{fdcebb}{9.35} & \cellcolor[HTML]{fcbca2}{12.35} & \cellcolor[HTML]{fcb89e}{12.99} \\
 \hdashline
\multirow[t]{3}{*}{LLM} & ChatGPT & \cellcolor[HTML]{dcf2d7}{16.4} & \cellcolor[HTML]{e5f5e0}{12.8} & \cellcolor[HTML]{fee0d2}{6.40} & \cellcolor[HTML]{fdd4c2}{8.32} & \cellcolor[HTML]{fed8c7}{7.65} & \cellcolor[HTML]{fcbda4}{12.14} \\
 & GPT4 & \cellcolor[HTML]{c9eac2}{24.4} & \cellcolor[HTML]{d0edca}{21.4} & \cellcolor[HTML]{fee6da}{4.62} & \cellcolor[HTML]{fee8dd}{4.00} & \cellcolor[HTML]{fdcbb6}{9.78} & \cellcolor[HTML]{fedecf}{6.72} \\
 & GPT4o & \cellcolor[HTML]{e4f5df}{12.9} & \cellcolor[HTML]{ebf7e7}{8.7} & \cellcolor[HTML]{fc9070}{19.01} & \cellcolor[HTML]{f96245}{26.07} & \cellcolor[HTML]{fc8d6d}{19.69} & \cellcolor[HTML]{f96245}{26.14} \\
\bottomrule
\end{tabular}
\end{table*}

\begin{figure*}[htbp]
    \centering
    \begin{subfigure}[b]{0.32\textwidth}
        \centering
        \includegraphics[width=\textwidth]{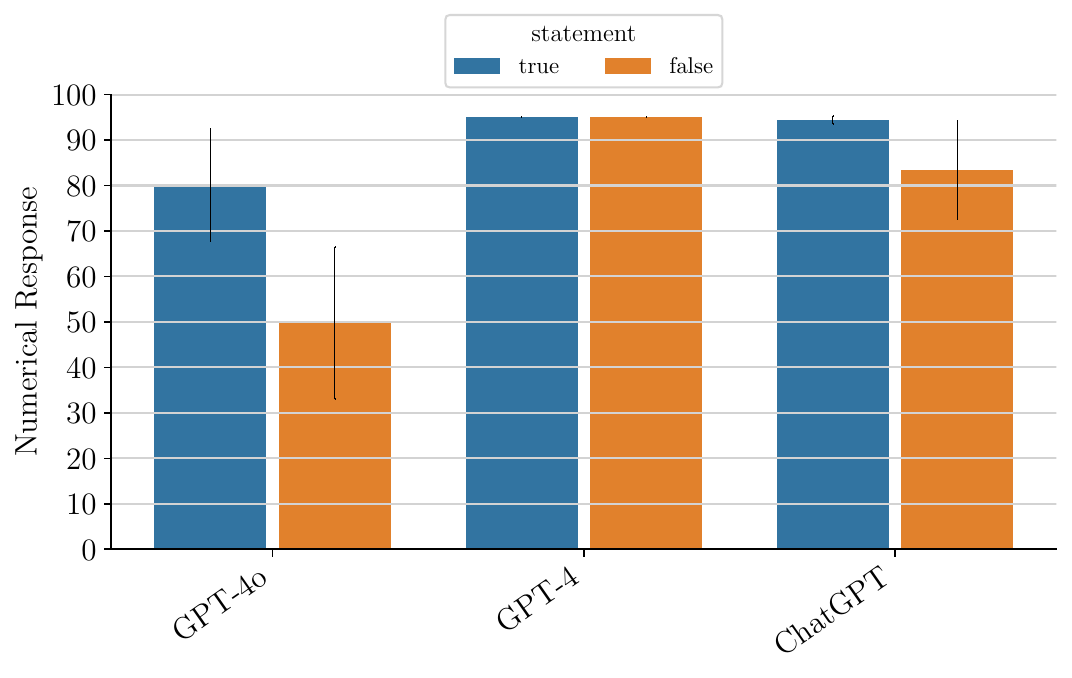}
        \caption{almost certain}
        \label{sfig:mean-rated-prob:almost-certain:prob-decoding}
    \end{subfigure}
    \hfill
    \begin{subfigure}[b]{0.32\textwidth}
        \centering
        \includegraphics[width=\textwidth]{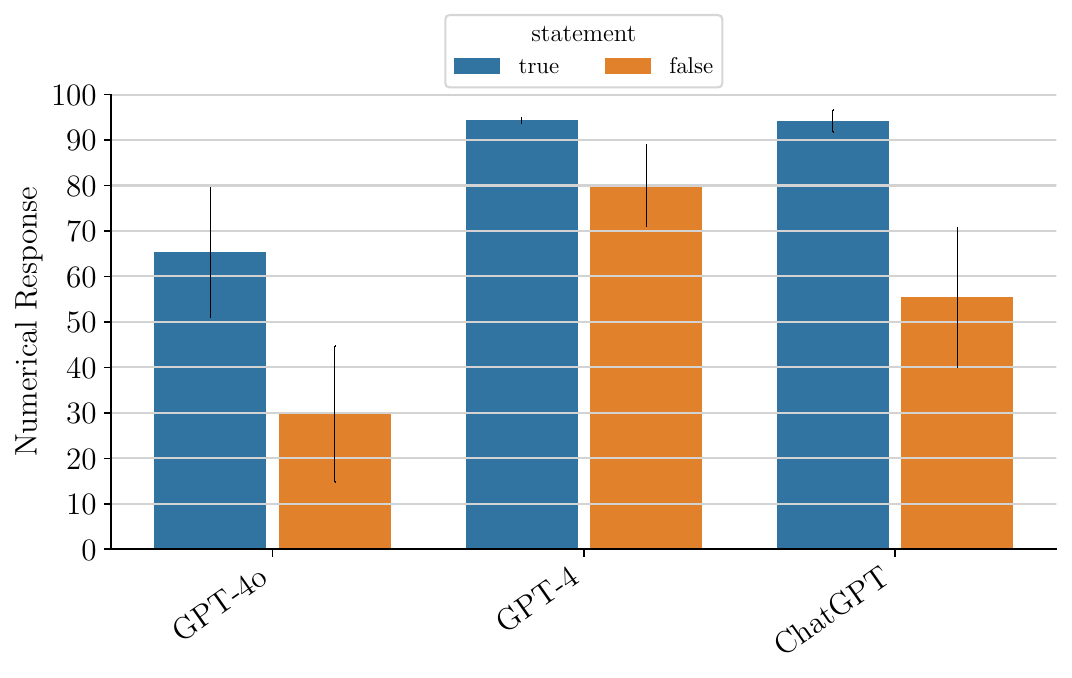}
        \caption{``highly likely''}
        \label{sfig:mean-rated-prob:highly-likely:prob-decoding}
    \end{subfigure}
    \hfill
    \begin{subfigure}[b]{0.32\textwidth}
        \centering
        \includegraphics[width=\textwidth]{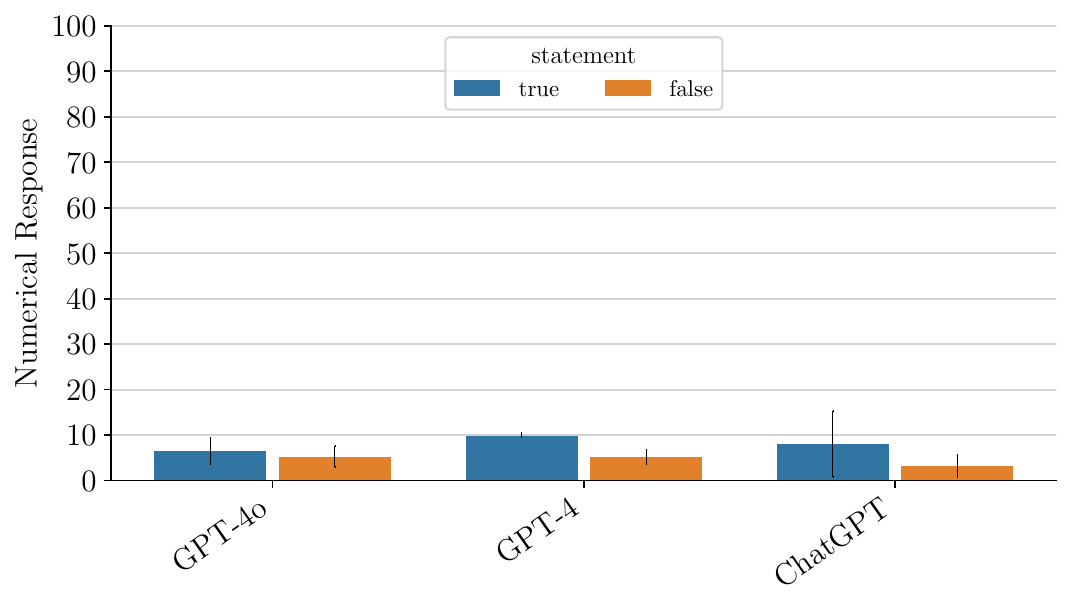}
        \caption{``highly unlikely''}
        \label{sfig:mean-rated-prob:highly-unlikely:prob-decoding}
    \end{subfigure}
    \begin{subfigure}[b]{0.32\textwidth}
        \centering
        \includegraphics[width=\textwidth]{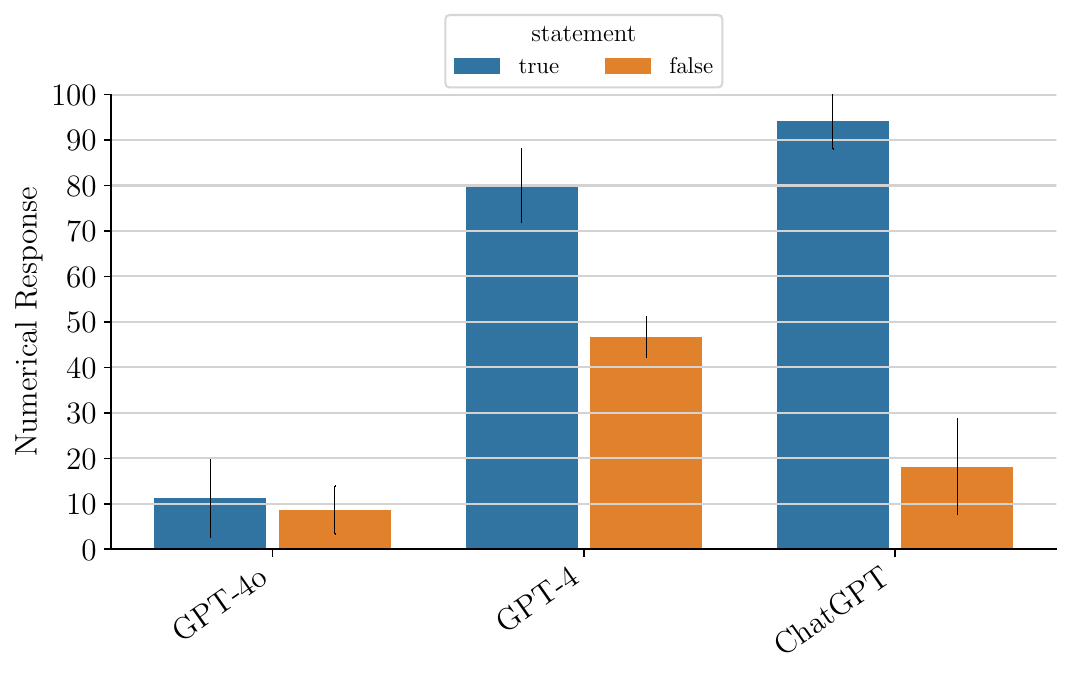}
        \caption{``possible''}
        \label{sfig:mean-rated-prob:possible:prob-decoding}
    \end{subfigure}
    \hfill
    \begin{subfigure}[b]{0.32\textwidth}
        \centering
        \includegraphics[width=\textwidth]{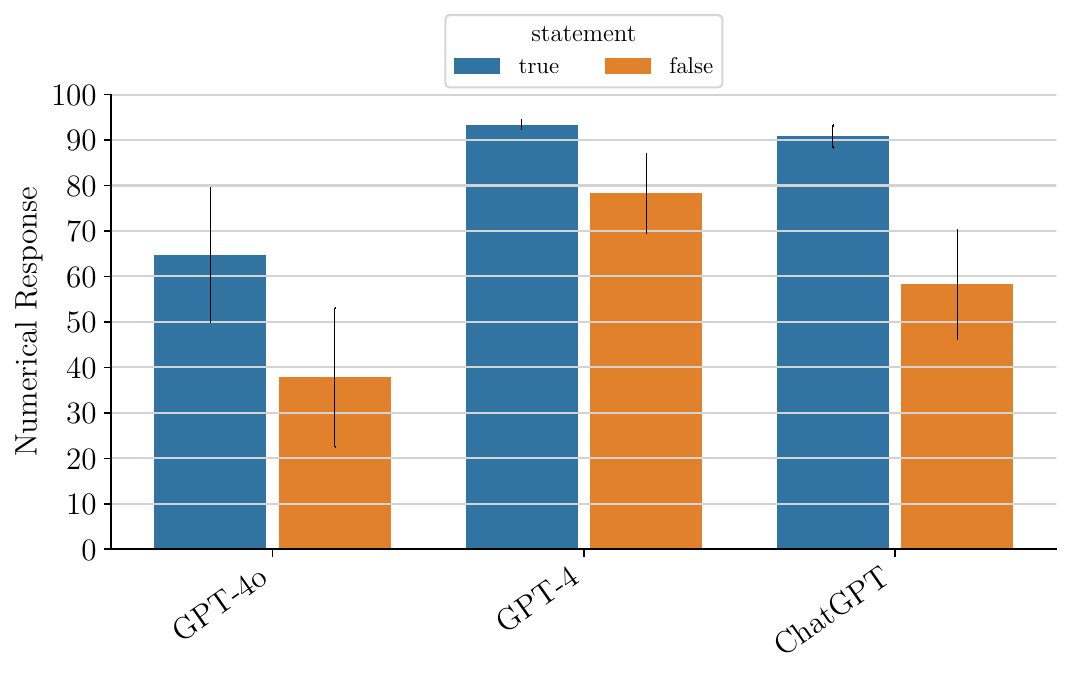}
        \caption{``very likely''}
        \label{sfig:mean-rated-prob:very-likely:prob-decoding}
    \end{subfigure}
    \hfill
    \begin{subfigure}[b]{0.32\textwidth}
        \centering
        \includegraphics[width=\textwidth]{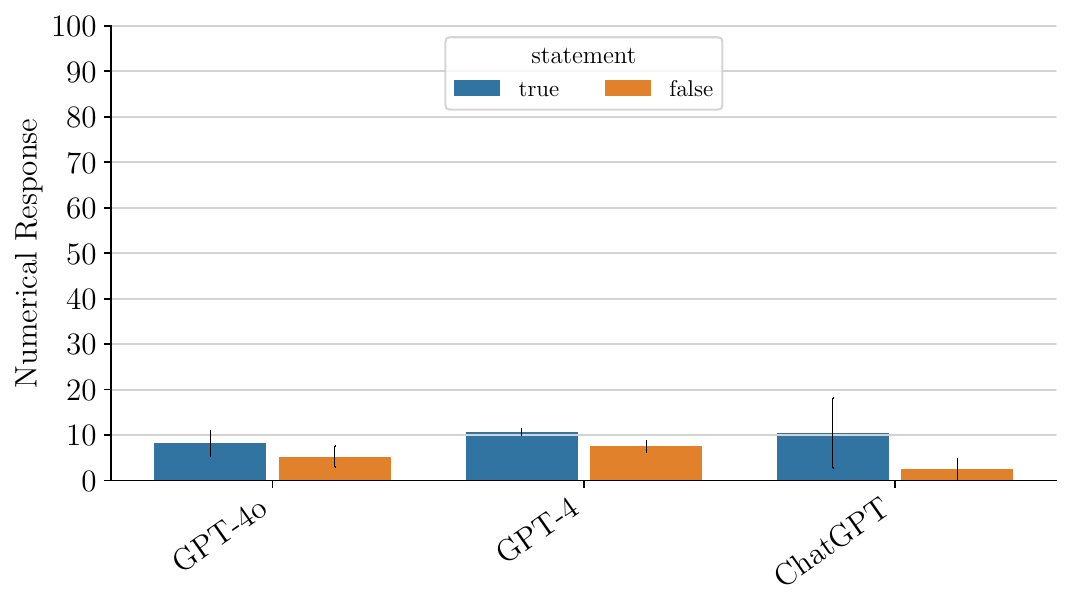}
        \caption{``very unlikely''}
        \label{sfig:mean-rated-prob:very-unlikely:prob-decoding}
    \end{subfigure}
    \begin{subfigure}[b]{0.32\textwidth}
        \centering
        \includegraphics[width=\textwidth]{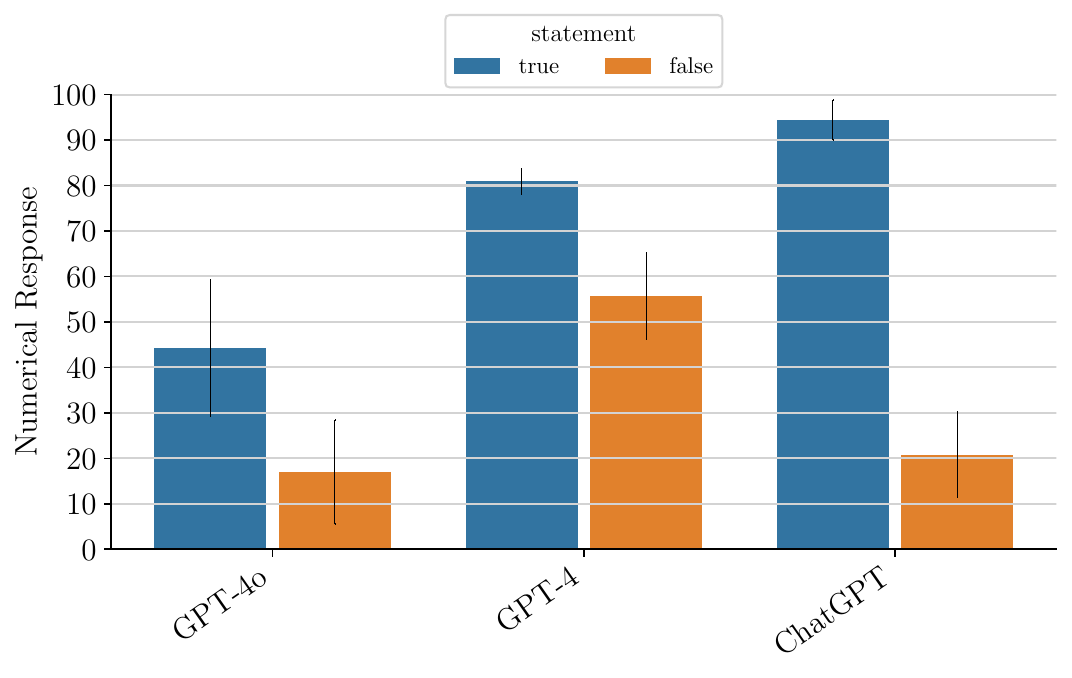}
        \caption{``probable''}
        \label{sfig:mean-rated-prob:probable:prob-decoding}
    \end{subfigure}
    \begin{subfigure}[b]{0.32\textwidth}
        \centering
        \includegraphics[width=\textwidth]{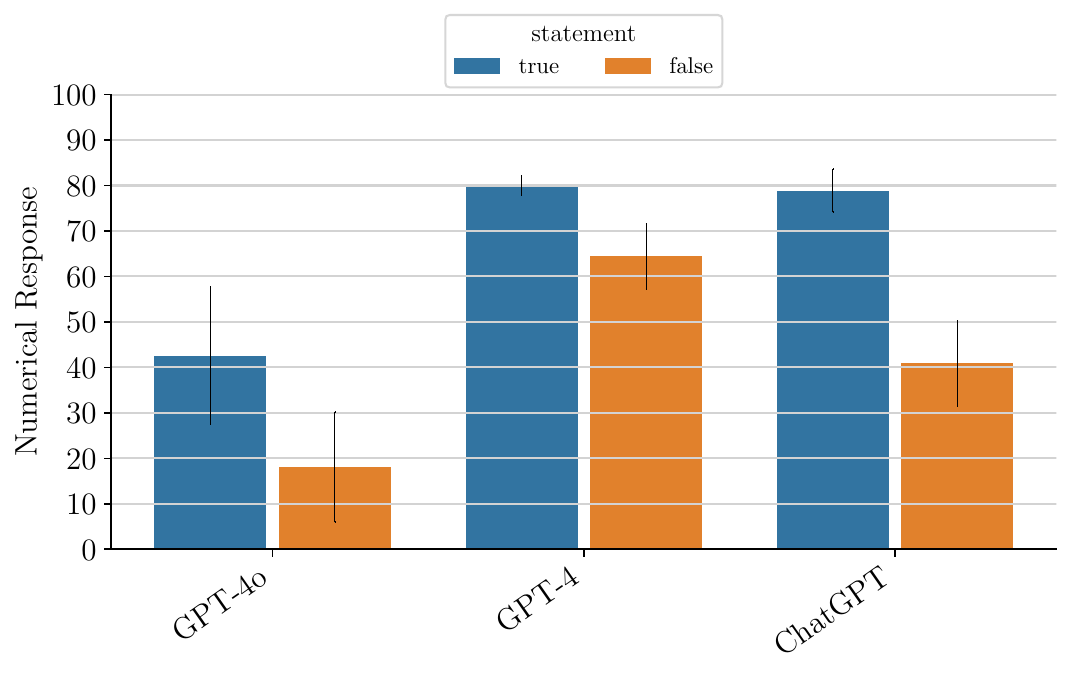}
        \caption{``likely''}
        \label{sfig:mean-rated-prob:likely:prob-decoding}
    \end{subfigure}
    \hfill
    \begin{subfigure}[b]{0.32\textwidth}
        \centering
        \includegraphics[width=\textwidth]{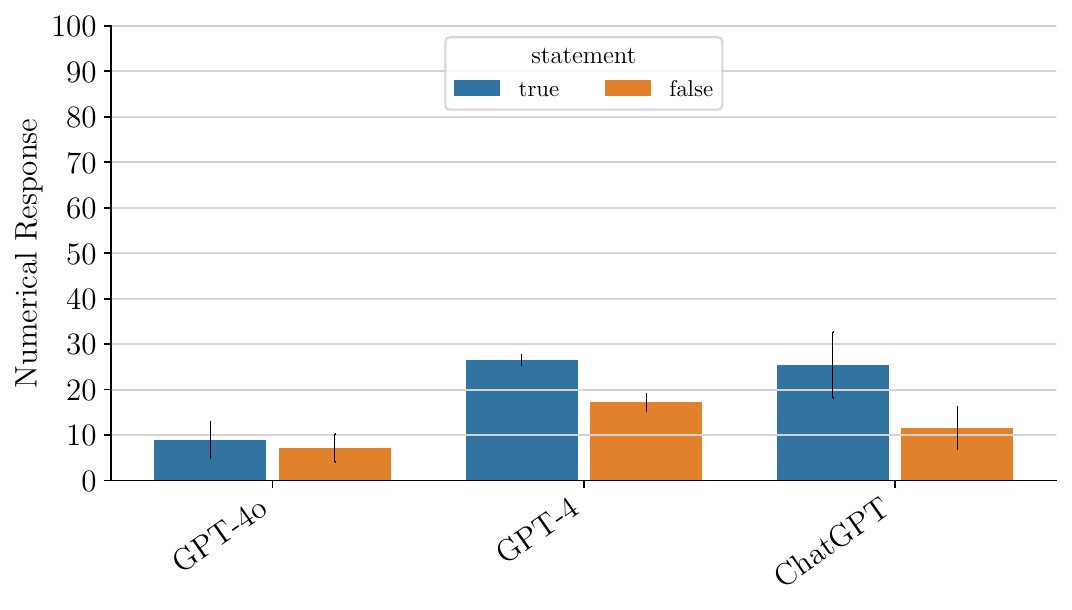}
        \caption{``not likely''}
        \label{sfig:mean-rated-prob:not-likely:prob-decoding}
    \end{subfigure}
    \hfill
    \begin{subfigure}[b]{0.32\textwidth}
        \centering
        \includegraphics[width=\textwidth]{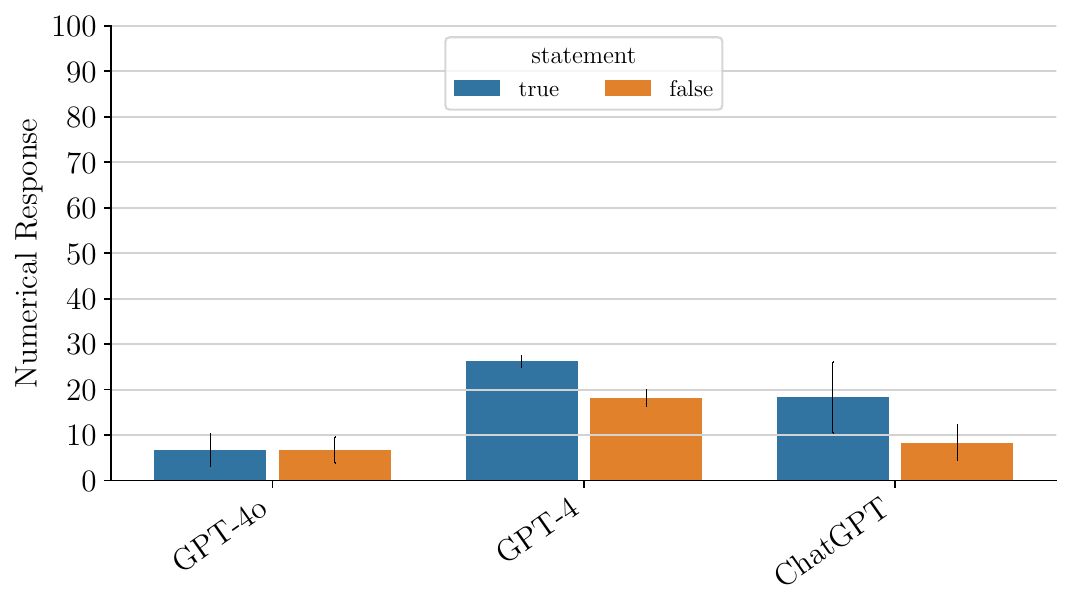}
        \caption{``unlikely''}
        \label{sfig:mean-rated-prob:unlikely:prob-decoding}
    \end{subfigure}
    \begin{subfigure}[b]{0.32\textwidth}
        \centering
        \includegraphics[width=\textwidth]{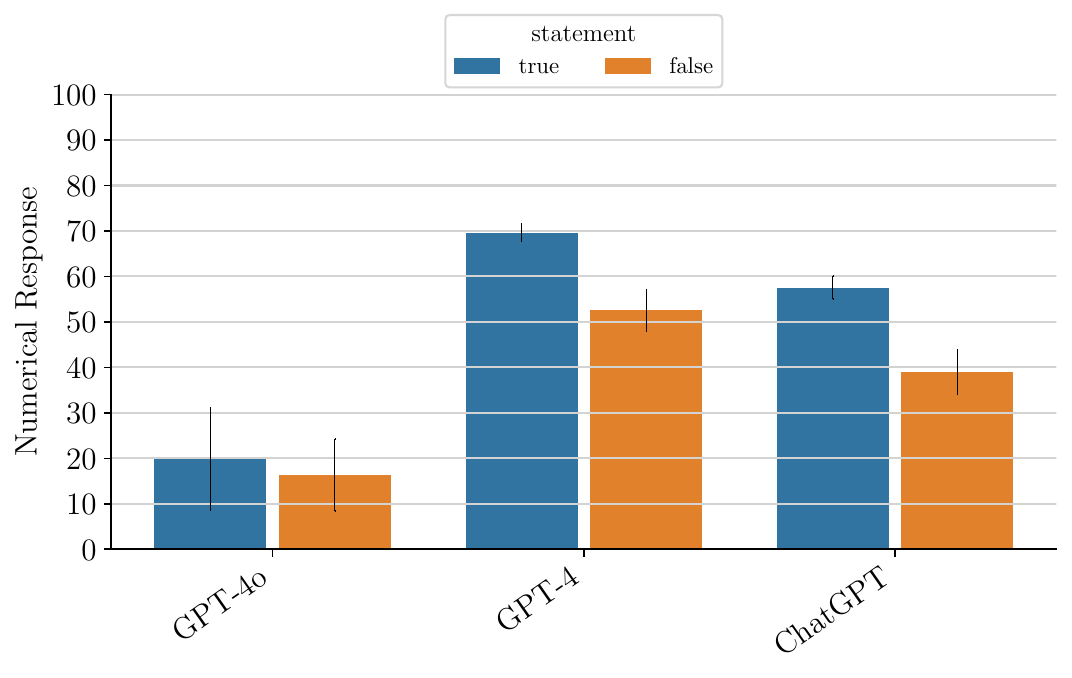}
        \caption{``somewhat likely''}
        \label{sfig:mean-rated-prob:somewhat-likely:prob-decoding}
    \end{subfigure}
    \hfill
    \begin{subfigure}[b]{0.32\textwidth}
        \centering
        \includegraphics[width=\textwidth]{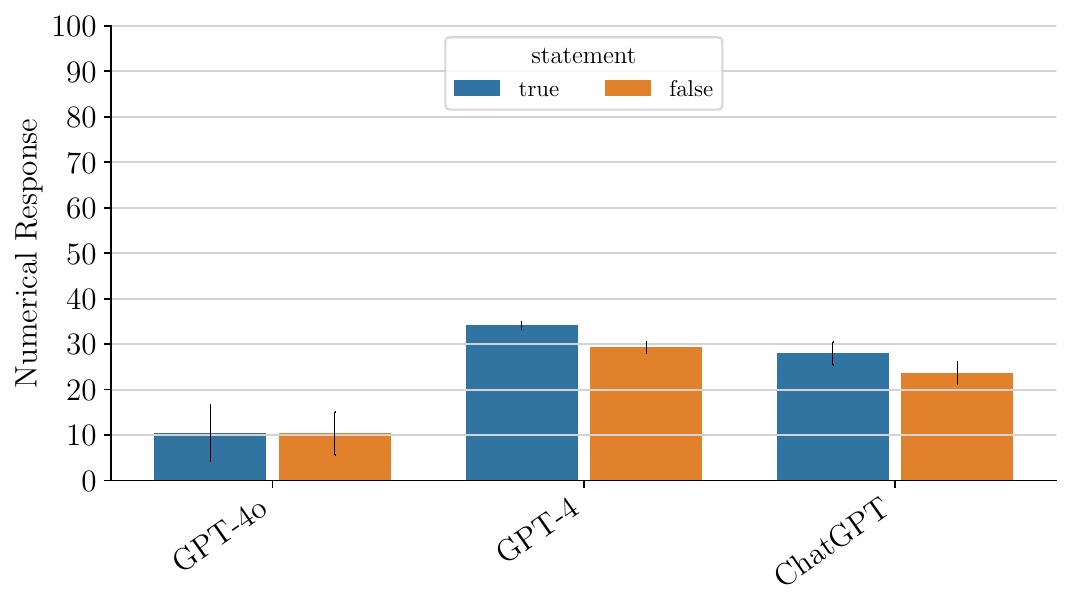}
        \caption{``somewhat unlikely''}
        \label{sfig:mean-rated-prob:somewhat-unlikely:prob-decoding}
    \end{subfigure}
    \begin{subfigure}[b]{0.32\textwidth}
        \centering
        \includegraphics[width=\textwidth]{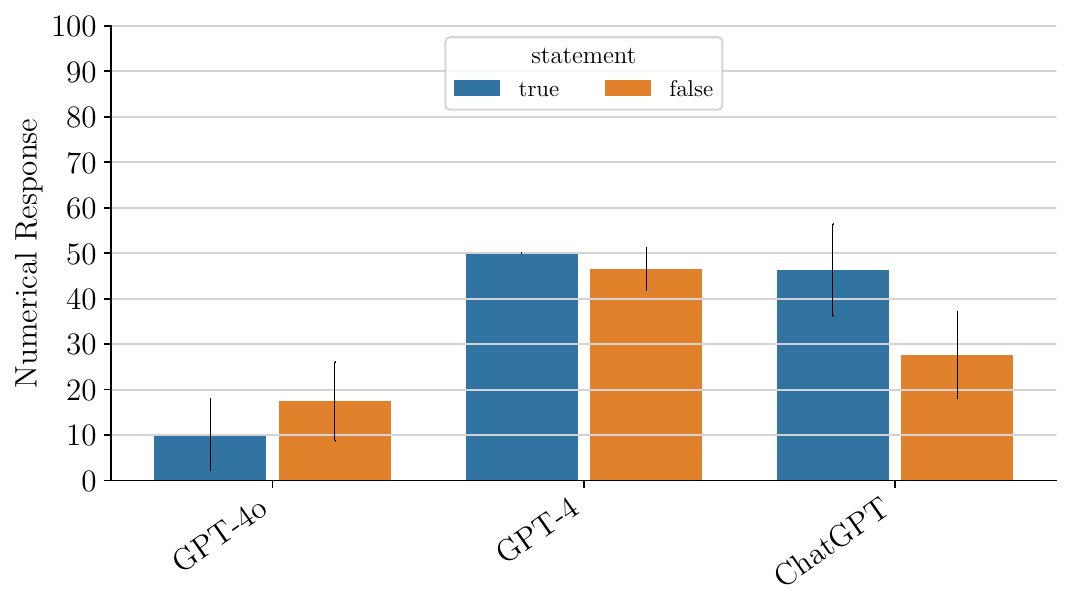}
        \caption{``uncertain''}
        \label{sfig:mean-rated-prob:uncertain:prob-decoding}
    \end{subfigure}
    \begin{subfigure}[b]{0.32\textwidth}
        \centering
        \includegraphics[width=\textwidth]{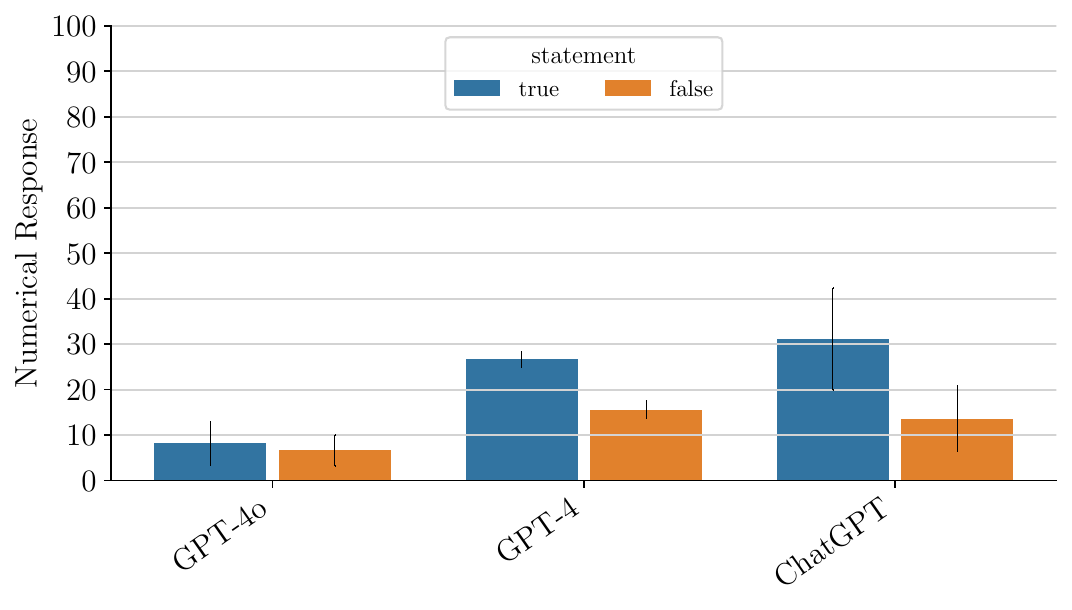}
        \caption{``doubtful''}
        \label{sfig:mean-rated-prob:doubtful:prob-decoding}
    \end{subfigure}
    \caption{\textbf{Mean numerical response discriminated by truthfulness of verifiable statements, estimated using probabilistic decoding (\texttt{temperature=1}).} Overall, we still observe statistically meaningful gaps depending on the truthfulness of the statement.}
    \label{fig:fractional:mean-rated-prob}
\end{figure*}

\section{Ablation: 0-shot vs 2-shot}
\label{app:sec:zero-shot-vs-2-shot}

Table \ref{tab:app:pa-breakdown-expr:nv-0shot} reports the proportional agreement (PA) metric discriminated by uncertainty expression in the non-verifiable and verifiable settings, respectively.
The values are reported with respect to the population-level human distribution described in Section \ref{sec:initalexp}, denoted \texttt{Human+NV}. 

\begin{table*}
\tiny
\centering
\caption{\textbf{Proportional Agreement (PA) score per uncertainty expression in the non-verifiable setting}. The scores are with respect to the population-level human reference distribution (\texttt{Human+NV}).}
\label{tab:app:pa-breakdown-expr:nv-0shot}
\begin{tabular}{lcccccccccc}
\toprule
 & \olmo  & \gemma & \llama  & \llamasmall & \chatgpt & \gptf & \gptfo & \mixtralmoelg & \mixtralmoe \\
Methodology & full & full & full & full & top-k & top-k & top-k & sampling & sampling \\
\midrule
Average & {9.3} & {2.4} & {18.3} & {14.4} & {17.1} & {25.9} & {22.3} & {20.2} & {18.8} \\
Standard Deviation & {9.2} & {4.0} & {6.4} & {5.2} & {8.3} & {12.2} & {13.1} & {14.5} & {13.1} \\
\hdashline
almost certain & \cellcolor[HTML]{a8dca2}{35.3} & \cellcolor[HTML]{f7fcf5}{0.2} & \cellcolor[HTML]{bee5b8}{27.8} & \cellcolor[HTML]{def2d9}{15.5} & \cellcolor[HTML]{afdfa8}{33.0} & \cellcolor[HTML]{53b466}{58.0} & \cellcolor[HTML]{56b567}{57.4} & \cellcolor[HTML]{48ae60}{60.6} & \cellcolor[HTML]{4eb264}{59.0} \\
doubtful & \cellcolor[HTML]{ebf7e7}{8.4} & \cellcolor[HTML]{f7fcf5}{0.0} & \cellcolor[HTML]{dbf1d6}{16.5} & \cellcolor[HTML]{dbf1d6}{16.5} & \cellcolor[HTML]{ebf7e7}{8.7} & \cellcolor[HTML]{d8f0d2}{18.0} & \cellcolor[HTML]{e5f5e0}{12.7} & \cellcolor[HTML]{ebf7e7}{8.9} & \cellcolor[HTML]{e8f6e3}{10.7} \\
highly likely & \cellcolor[HTML]{d0edca}{21.2} & \cellcolor[HTML]{f7fcf5}{0.4} & \cellcolor[HTML]{ebf7e7}{8.7} & \cellcolor[HTML]{f0f9ed}{5.1} & \cellcolor[HTML]{e6f5e1}{12.0} & \cellcolor[HTML]{c1e6ba}{27.2} & \cellcolor[HTML]{d5efcf}{19.4} & \cellcolor[HTML]{c6e8bf}{25.6} & \cellcolor[HTML]{dcf2d7}{16.3} \\
highly unlikely & \cellcolor[HTML]{f1faee}{4.5} & \cellcolor[HTML]{f6fcf4}{0.5} & \cellcolor[HTML]{dcf2d7}{16.0} & \cellcolor[HTML]{e1f3dc}{14.1} & \cellcolor[HTML]{cdecc7}{22.6} & \cellcolor[HTML]{a9dca3}{35.1} & \cellcolor[HTML]{afdfa8}{33.0} & \cellcolor[HTML]{bce4b5}{28.7} & \cellcolor[HTML]{dbf1d6}{16.7} \\
not likely & \cellcolor[HTML]{f1faee}{4.3} & \cellcolor[HTML]{f6fcf4}{0.5} & \cellcolor[HTML]{daf0d4}{17.6} & \cellcolor[HTML]{daf0d4}{17.6} & \cellcolor[HTML]{e2f4dd}{13.8} & \cellcolor[HTML]{d9f0d3}{17.9} & \cellcolor[HTML]{e7f6e2}{11.5} & \cellcolor[HTML]{e9f7e5}{9.9} & \cellcolor[HTML]{e8f6e3}{10.8} \\
possible & \cellcolor[HTML]{f3faf0}{3.1} & \cellcolor[HTML]{e5f5e0}{12.7} & \cellcolor[HTML]{dff3da}{15.2} & \cellcolor[HTML]{e5f5e0}{12.8} & \cellcolor[HTML]{e0f3db}{14.5} & \cellcolor[HTML]{def2d9}{15.4} & \cellcolor[HTML]{e8f6e3}{10.7} & \cellcolor[HTML]{e9f7e5}{9.5} & \cellcolor[HTML]{dff3da}{14.9} \\
probable & \cellcolor[HTML]{edf8ea}{6.9} & \cellcolor[HTML]{f0f9ed}{4.9} & \cellcolor[HTML]{ddf2d8}{15.9} & \cellcolor[HTML]{ebf7e7}{8.7} & \cellcolor[HTML]{e8f6e4}{10.2} & \cellcolor[HTML]{e0f3db}{14.7} & \cellcolor[HTML]{def2d9}{15.4} & \cellcolor[HTML]{ebf7e7}{9.0} & \cellcolor[HTML]{e3f4de}{13.4} \\
somewhat likely & \cellcolor[HTML]{f4fbf2}{2.2} & \cellcolor[HTML]{e9f7e5}{9.7} & \cellcolor[HTML]{dcf2d7}{16.0} & \cellcolor[HTML]{e8f6e3}{10.9} & \cellcolor[HTML]{ecf8e8}{8.1} & \cellcolor[HTML]{dbf1d6}{16.6} & \cellcolor[HTML]{e7f6e3}{11.0} & \cellcolor[HTML]{dbf1d6}{16.5} & \cellcolor[HTML]{eff9eb}{6.1} \\
somewhat unlikely & \cellcolor[HTML]{f2faef}{3.5} & \cellcolor[HTML]{f6fcf4}{0.6} & \cellcolor[HTML]{ceecc8}{22.2} & \cellcolor[HTML]{d0edca}{21.2} & \cellcolor[HTML]{d7efd1}{18.4} & \cellcolor[HTML]{cdecc7}{22.3} & \cellcolor[HTML]{d5efcf}{19.5} & \cellcolor[HTML]{ecf8e8}{7.5} & \cellcolor[HTML]{e8f6e3}{10.8} \\
uncertain & \cellcolor[HTML]{f3faf0}{3.0} & \cellcolor[HTML]{f6fcf4}{0.6} & \cellcolor[HTML]{a9dca3}{35.1} & \cellcolor[HTML]{c6e8bf}{25.5} & \cellcolor[HTML]{aadda4}{34.5} & \cellcolor[HTML]{a9dca3}{35.1} & \cellcolor[HTML]{abdda5}{34.1} & \cellcolor[HTML]{aadda4}{34.5} & \cellcolor[HTML]{b0dfaa}{32.6} \\
unlikely & \cellcolor[HTML]{eff9eb}{6.0} & \cellcolor[HTML]{f7fcf5}{0.0} & \cellcolor[HTML]{dbf1d5}{17.0} & \cellcolor[HTML]{dbf1d5}{16.8} & \cellcolor[HTML]{e3f4de}{13.6} & \cellcolor[HTML]{dbf1d6}{16.7} & \cellcolor[HTML]{e7f6e3}{11.0} & \cellcolor[HTML]{eaf7e6}{9.2} & \cellcolor[HTML]{dff3da}{15.2} \\
very likely & \cellcolor[HTML]{dbf1d6}{16.6} & \cellcolor[HTML]{f6fcf4}{0.9} & \cellcolor[HTML]{e0f3db}{14.5} & \cellcolor[HTML]{ebf7e7}{8.9} & \cellcolor[HTML]{e7f6e3}{11.3} & \cellcolor[HTML]{d0edca}{21.1} & \cellcolor[HTML]{d0edca}{21.4} & \cellcolor[HTML]{d1edcb}{20.7} & \cellcolor[HTML]{d7efd1}{18.5} \\
very unlikely & \cellcolor[HTML]{eff9eb}{6.1} & \cellcolor[HTML]{f7fcf5}{0.0} & \cellcolor[HTML]{def2d9}{15.6} & \cellcolor[HTML]{e1f3dc}{14.2} & \cellcolor[HTML]{d0edca}{21.2} & \cellcolor[HTML]{9cd797}{38.8} & \cellcolor[HTML]{b1e0ab}{32.1} & \cellcolor[HTML]{cdecc7}{22.3} & \cellcolor[HTML]{d5efcf}{19.2} \\
\bottomrule
\end{tabular}
\end{table*}

\end{document}